%% file: manuscript.tex
\documentclass{article}

\usepackage{arxiv}

\usepackage{color,soul}
\usepackage{comment}
\usepackage{amsmath}
\usepackage{natbib}
\usepackage{dblfloatfix}  
\setcitestyle{square,comma,numbers,sort}
\usepackage{float}
\usepackage[font={footnotesize},labelfont={footnotesize}]{caption}
\usepackage[font={footnotesize},labelfont={footnotesize}]{subcaption}
\usepackage{tikz,pgfplots,adjustbox}
\tikzset{>=latex}
\pgfplotsset{compat=1.3}
\usetikzlibrary{patterns}
\usetikzlibrary{pgfplots.groupplots}
\usepackage{bbold}
\usepgfplotslibrary{fillbetween}

\usepackage[utf8]{inputenc} 
\usepackage[T1]{fontenc}    
\usepackage{hyperref}       
\usepackage{url}            
\usepackage{booktabs}       
\usepackage{amsfonts}       
\usepackage{nicefrac}       
\usepackage{microtype}      
\usepackage{lipsum}
\usepackage{graphicx}
\graphicspath{ {./images/} }

\title{Geometric Affinity Propagation for Clustering with Network Knowledge}

\author{
 Omar Maddouri \\
  Department of Electrical and Computer Engineering\\
  Texas A\&M University\\
  College Station, TX 77843, USA \\
  \texttt{omar.maddouri@tamu.edu} \\
   \And
 Xiaoning Qian \\
  Department of Electrical and Computer Engineering\\
  Texas A\&M University\\
  College Station, TX 77843, USA \\
  \texttt{xqian@ece.tamu.edu } \\
  \And
 Byung-Jun Yoon \\
  Department of Electrical and Computer Engineering\\
  Texas A\&M University\\
  College Station, TX 77843, USA \\
  \texttt{bjyoon@ece.tamu.edu} \\
}

\begin{document}
\maketitle
\begin{abstract}

Clustering data into meaningful subsets is a major task in scientific data analysis. To date, various strategies ranging from model-based approaches to data-driven schemes, have been devised for efficient and accurate clustering. One important class of clustering methods that is of a particular interest is the class of exemplar-based approaches. This interest primarily stems from the amount of compressed information encoded in these exemplars that effectively reflect the major characteristics of the respective clusters. Affinity propagation (AP) has proven to be a powerful exemplar-based approach that refines the set of optimal exemplars by iterative pairwise message updates.  However, a critical limitation is its inability to capitalize on known networked relations between data points often available for various scientific datasets. To mitigate this shortcoming, we propose geometric-AP, a novel clustering algorithm that effectively extends AP to take advantage of the network topology. Geometric-AP obeys network constraints and uses max-sum belief propagation to leverage the available network topology for generating smooth clusters over the network. Extensive performance assessment reveals a significant enhancement in the quality of the clustering results when compared to benchmark clustering schemes. Especially, we demonstrate that geometric-AP performs extremely well even in cases where the original AP fails drastically.
\footnote{This work has been submitted to the IEEE for possible publication. Copyright may be transferred without notice, after which this version may no longer be accessible.}
\end{abstract}


\section{Introduction}
\label{sec:introduction}
Clustering refers to the process of partitioning of data into groups of points that share specific characteristics. Similar instances therefore must be assigned to the same cluster. The definition of similarity here is often subjective and greatly depends on the ultimate goal expected from the analysis~\citep{Jain1999}. This makes clustering a difficult combinatorial problem where researchers continuously strive to develop innovative computational methods to address the increasing complexity associated with new large-scale datasets. The majority of current clustering methods are generally fed either with a vector of observations in the feature space or with measures of proximity between data points~\citep{Danial2009}. Their mission, consequently, is to identify expressive clusters that dissect the dynamics present in the data. Towards this goal, two broad classes of approaches have been proposed. The first set of methods includes models such as kmeans~\citep{Hartigan1979} and kmedoids~\citep{Kaufman1990} and directly operates on the original feature space to group the data points based on raw pairwise similarities. The other class of approaches maps the manifold structures of the observed feature space into a different latent space where the data might be more separable. Spectral clustering for instance tracks clusters of irregular shapes by leveraging the eigenvalues of the similarity matrix to embed the data into a lower dimensional space~\citep{Ng2001}. More recent approaches employ deep neural architectures to train non-linear embeddings that better capture the hidden interactions underlying complex systems~\citep{Perozzi2014,Cao2016,Kipf2017,Zhang2019}.

Despite the enhanced performance achieved by representation-learning-based approaches, methods that directly operate on the primitive data remain highly sought after in the research community. Exemplar-based approaches present an epitome of such methods that have been extensively applied in diverse fields. For example, exemplars have been widely used in management sciences to find optimal facility locations~\citep{Maranzana1964}. In multi-controller placement problem, exemplars were utilized to locate the best controller sites for software-defined networks to minimize the propagation latency with the switches~\citep{Heller2012}. Affinity propagation (AP)~\citep{Frey2007} is one of the most appealing exemplar-based clustering methods that have been proposed in recent years. AP iteratively refines the set of candidate points that best exemplify the entire dataset by exchanging messages between all pairs of data points until a set of representatives emerges. Many other implicit exemplar-based methods, including kmeans, consider for ``virtual'' centroids that may not belong to the original set of data points. For example, each exemplar in kmeans clustering is determined by the average features of the corresponding cluster members. The fact that AP explicitly selects exemplars from the original dataset gives it a significant advantage over other implicit methods--especially, in terms of interpretation and utilization, since the identified exemplars seamlessly relate to many real-world applications.
While finding the optimal set of exemplars is an NP-hard problem, AP has proven to be very efficient in realistic settings, being capable of rapidly handling thousands of high dimensional instances. This is enabled by an efficient belief propagation scheme that can take advantage of parallel implementation, which makes AP operate with only $\mathcal{O}(N^2)$ messages, where $N$ represents the total number of data points~\citep{Frey2007}. In addition to computational efficiency, AP has been shown to yield accurate clustering results, which are relatively insensitive to initialization.

However, one notable shortcoming of the original AP is its limited ability to integrate different levels of information to perform clustering, since it solely relies on pairwise affinities between data points for partitioning the dataset. In various applications such as the discovery of communities in social networks~\citep{Parthasarathy2011}, finding functional modules in biological networks~\citep{Mitra2013, Wang2020}, or optimizing the usage of communication channels in transportation networks~\citep{Zhang2006}, the systems are often described using node features as well as network information. AP is unable to leverage such network knowledge to enhance the clustering accuracy. Furthermore, the original AP often faces challenges for datasets with irregularly shaped clusters, sparse datasets, and multi-subclass systems.

\subsection{Previous Work}

To alleviate the aforementioned limitations, efforts have been made to adapt AP to specific applications, which we briefly review in this section.
In the original formulation of AP, hard consistency constraints have been placed on the elected exemplars that do not refer to themselves~\citep{Frey2007}, which led to efficient identification of convex clusters where data points are well represented by their associated exemplars. To extend AP beyond regularly shaped clusters, a soft-constraint AP (SCAP) method has been proposed, in which the hard constraints have been relaxed. Evaluation on clustering microarray data has shown that SCAP is more efficient than AP in analyzing noisy and irregularly organized datasets~\citep{Leone2007}. For sparse datasets such as sparse graphs, a fast implementation of AP sets the similarity between unconnected nodes to very small values. This confines the exemplars within direct adjacency of the data points, leading to finely fragmented clusters. To mitigate this shattering pattern, a greedy hierarchical AP (GHAP) algorithm has been proposed~\citep{Xiao2007}. GHAP repeatedly clusters the set of exemplars that emerge from the previous iterations and updates the exemplars labels until a satisfactory coarse clustering is obtained~\citep{Xiao2007}. An evolved theoretical approach for hierarchical clustering by affinity propagation, called Hierarchical AP (HAP), adopts an inference algorithm that disseminates information up and down the hierarchy~\citep{Givoni2011}. HAP outperforms GHAP that clusters only one layer at a time. A semi-supervised AP was proposed in~\citep{Givoni2009},  which considers clustering when  prior knowledge exists for some pairs of data points indicating their  similarity (must-link (ML)) or dissimilarity (cannot-link (CL)).
Building on~\citep{Leone2007}, a soft instance-level constraint version has been presented for the semi-supervised AP~\citep{Arzeno2014}. Furthermore, AP has been utilized to analyze data streaming dynamics. Instead of operating on high-throughput data, Streaming-AP puts in cascade a weighted clustering step to extract subsets from the data and then performs hierarchical clustering followed by an additional weighted clustering procedure~\citep{Zhang2008}. Another attractive advantage of AP is that it automatically identifies the number of clusters in the data, but its downside is the lack of control over the desired size of the identified clusters. To remedy this limitation, AP has been extended to make the cluster size more manageable. For example, various priors such as the Dirichlet process priors have been integrated into the clustering process for this purpose~\citep{Tarlow2008}. Notably, hierarchical clustering principles have been widely utilized to extend the original AP. The main reason is that the single-exemplar design of AP becomes inadequate when applied to model multi-subclass systems. In this regard, a more explicit approach called multi-exemplar affinity propagation (MEAP) has been proposed to address the limitations of AP for multi-subclass problems. In MEAP, two types of exemplars are being identified: A set of sub-exemplars are associated with super-exemplars to approximate the subclasses in the category~\citep{Wang2013}. MEAP has shown consistent performance in handling problems like scene analysis and character recognition.

As for network information, it has been less considered in exemplar-based clustering literature. Fundamentally, it is more difficult to combine pairwise similarity measures obtained from two different observations: node features and network topology. Additionally, a unified criterion for identifying exemplars and cluster membership based on a compound affinity needs to be determined. For AP, an early attempt employed diffusion kernel similarities obtained using the Laplacian matrix of the network to perform a community detection task~\citep{Liu2011}. A more recent approach has addressed the same problem by adaptively updating the similarity matrix during the message updates using the degree centrality of potential exemplars~\citep{Taheri2020}.
Although tailored similarity measures can slightly improve the efficiency of AP as discussed in~\citep{Frey2007}, they are known to be insufficient and very limited in handling problems with complex underlying structures~\citep{Tarlow2008}.

\subsection{Extension of Affinity Propagation to geometric-AP}

Motivated by the increasing availability of network information in many structured datasets, this paper extends the feature-based affinity propagation (AP) algorithm to a geometric model, which we call Geometric Affinity Propagation (geometric-AP), where the original energy function is being minimized under additional topological constraints. Indeed, our work builds on top of the latest advances in graph clustering research and endorses two universally accepted properties of connectivity and density for any desired graph cluster. That being said, a good graph cluster should intuitively be connected. Also, its internal density should be significantly higher than the density of the full graph~\citep{Schaeffer2007}. In the context of our work, we adopt a more lenient definition of connectivity and density as we are not strictly performing graph clustering. Instead, we require that members of each desired cluster should lie within the same region in the network. Additionally, we promote higher internal density of identified clusters by assuming that highly interacting nodes should belong to the same cluster.

To implement the above requirements for AP, we jointly modify the exemplar identification mechanism and the membership assignment procedure to incorporate the connectivity and  density properties, respectively.
First, we require that a potential exemplar should lie within the local neighborhood of referring nodes with respect to the network. Second, highly interacting nodes must share the same cluster membership. The first connectivity constraint is ensured through an additional penalty term in the optimized net similarity of AP. The second density requirement is secured using a new assignment policy that promotes membership selection among neighbor exemplars. Afterwards, a label smoothing operation is applied to reduce the misassignments and enable better generalization~\citep{Schindler2012,Muller2019}.

Compared to the original feature-based AP, geometric-AP has the following advantages.
\begin{itemize}
    \item It can seamlessly integrate the network information into the clustering setup and notably improve the performance without increasing the model complexity. 
    \item Unlike other approaches, geometric-AP does not use the network information to tailor the feature-based similarity but instead it jointly employs the node features along with the network information to conduct efficient clustering.
\end{itemize}

The remainder of this paper is organized as follows:
In section~\ref{sec:Affinity_Propagation}, we briefly provide a general description of AP. In
Section~\ref{sec:Geometric_Affinity_Propagation}, we introduce the new geometric-AP algorithm. The new model is first described and its underlying rationale is discussed. Then the new message updates are derived using a max-sum belief propagation algorithm to optimize the redesigned net similarity. A comparative study between AP and geometric-AP is conducted to show that the geometric-AP algorithm can
be viewed as a special case of AP that penalizes some clustering configurations under topological constraints. Sections~\ref{sec:Unsupervised_document_Clustering} and \ref{sec:Clustering_of_social_networks}
report the experimental results on two citation networks and one social network, respectively.
In section~\ref{sec:Geometric_AP_with_random_networks} we establish the statistical significance of the improvement claimed by the used network information. An ablation study is performed with 100 randomly permuted networks and the average clustering performance is reported.
In section~\ref{sec:Conclusion} we provide concluding remarks for this paper.

\section{Brief Review of Affinity Propagation}
\label{sec:Affinity_Propagation}
Affinity Propagation (AP) is a message passing algorithm that takes as input user-defined similarity measures for all data point pairs. Real-valued messages called \emph{responsibility} and \emph{availability} are iteratively exchanged between data points until a set of high-quality clusters gradually emerge around representative data points referred to as exemplars~\citep{Frey2007}.
Instead of pre-specifying the number of the desired clusters, AP can automatically promote some data points to be selected as exemplars by assigning large values to them in the diagonal of the similarity matrix. Thus, we call preferences the values $s\left ( k,k \right )$ for all data points $k$ that could be set in a way to foster some points to be exemplars. In the absence of any prior knowledge about the potential exemplars, all data points are initially considered equally likely to be selected as exemplars.
Based on the provided input $s(i,j)$, AP subsequently exchanges the two types of messages between data points to decide which instance would serve as a good exemplar.
The first message, called \emph{responsibility} and denoted by $r\left ( i,j \right )$, designates the message sent from point $i$ to candidate exemplar point $j$. By sending this message, point $i$ tells point $j$ about the accumulated evidence that point $j$ would be a good exemplar for point $i$ after assessing the potential of all other candidates.
The second communicated message is called \emph{availability} and is denoted by $a\left ( i,j \right )$. This message could be interpreted as the feedback from point $j$ to point $i$ delivering information about the accumulated evidence for how well suited it would be for point $i$ to select point $j$ as its exemplar by taking the information collected from other points into consideration.
The exchanged messages are defined and updated as follows:
\begin{equation}\label{eq1}
  r\left ( i,j \right )\leftarrow s\left ( i,j \right )-\max_{j^{'} s.t. j^{'} \neq  j}\left \{ a(i,j^{'}) + s(i,j^{'})\right \}.
\end{equation}
\begin{equation}\label{eq2}
  a\left ( i,j \right )\leftarrow \min\left \{ 0, r(j,j) + \sum_{i^{'} s.t. i^{'} \notin \left \{ i,j \right \}}^{ } \max \left \{ 0, r(i^{'},j) \right \}\right \}.
\end{equation}
Initially, all the availability messages are set to 0, except for the self availability, which is computed as follows:
\begin{equation}\label{eq3}
  a\left ( j,j \right )\leftarrow \sum_{i^{'} s.t. i^{'} \neq j}^{ } \max \left \{ 0, r(i^{'},j) \right \}.
\end{equation}
This ensures that the self availability of a given point is not inflated by higher responsibilities received from other points. In order to avoid numerical instabilities that may result from oscillating updates, an exchanged message $\mathrm{m}$ is damped as follows:
\begin{equation}\label{eq4}
  \mathrm{m}^{(t)} \leftarrow  \lambda ~\mathrm{m}^{(t-1)} + (1-\lambda)~\mathrm{m}^{(t)},
\end{equation}
where $\lambda$ is the damping factor.
Finally, at each iteration, we can determine the exemplar associated with each point by evaluating the following equation:
\begin{equation}\label{eq5}
  \mathtt{exemplar}(i) = \arg\max_{j}\left \{ a(i,j)+r(i,j) \right \}.
\end{equation}
AP converges when the clustering configuration remains steady for a predefined number of iterations.
\section{Geometric Affinity Propagation}
\label{sec:Geometric_Affinity_Propagation}
Geometric Affinity Propagation (geometric-AP) stems from the original formulation of AP where the clustering task has been viewed as a search over a wide set of valid configurations of the class labels $\mathbf{c}= \left ( c_{1}, c_{2}, ..., c_{N}\right )$ for $N$ data points~\citep{Frey2007}. Given a user-defined similarity matrix $\left [  s_{ij}\right ]_{N\times N}$, the search task turns out to be an optimization problem that aims at minimizing an energy function:
\begin{equation}\label{eq6}
  E\left ( \mathbf{c} \right ) = -\sum_{i=1}^{N}s\left ( i,c_{i} \right ),
\end{equation}
where $s\left ( i,c_{i} \right )$ is the similarity measure between data point $i$ and its corresponding exemplar $c_{i}$.

Under valid configuration constraints, the optimization problem can be reformulated as the maximization of a net similarity $\mathcal{S}$, defined as:
\begin{align}\label{eq7}
  \nonumber   \mathcal{S}\left ( \mathbf{c} \right ) & =-E\left ( \mathbf{c} \right )+\sum_{k=1}^{N}\delta _{k}\left ( \mathbf{c} \right ) \\
     & =\sum_{i=1}^{N}s\left ( i,c_{i} \right )+\sum_{k=1}^{N}\delta _{k}\left ( \mathbf{c} \right ),
\end{align}
where $\delta _{k}\left ( \mathbf{c} \right )$ is a penalty term expressed as:
\begin{align}\label{eq8}
  \delta _{k}\left ( \mathbf{c} \right )=
 \begin{cases}
-\infty,& \mbox{if }c_{k} \neq k \mbox{ but }\exists\ i:\ c_{i}=k, \\
 0,& \mbox{otherwise}.
\end{cases}
\end{align}
In geometric-AP, for valid association between any pair of data points $\left (i,k \right )$ that satisfies $c_{i}=k$, we further require that $k \in N_{\mathcal{G}}^{\tau}\left ( i \right )$ where $N_{\mathcal{G}}^{\tau}\left ( i \right )$ is the topological neighborhood of diameter $\tau$ with respect to graph $\mathcal{G}$ for the point $i$ defined as:
\begin{equation}\label{eq9}
N_{\mathcal{G}}^{\tau}\left ( i  \right ) = \left \{ x: \mbox{distance}_{\mathcal{G}}\left( x, i\right) \leq \tau\right \},
\end{equation}
where $\mbox{distance}_{\mathcal{G}}$ represents a topological distance with respect to graph $\mathcal{G}$.

\begin{figure}[ht!]
    \centering
    \includegraphics[scale=0.5]{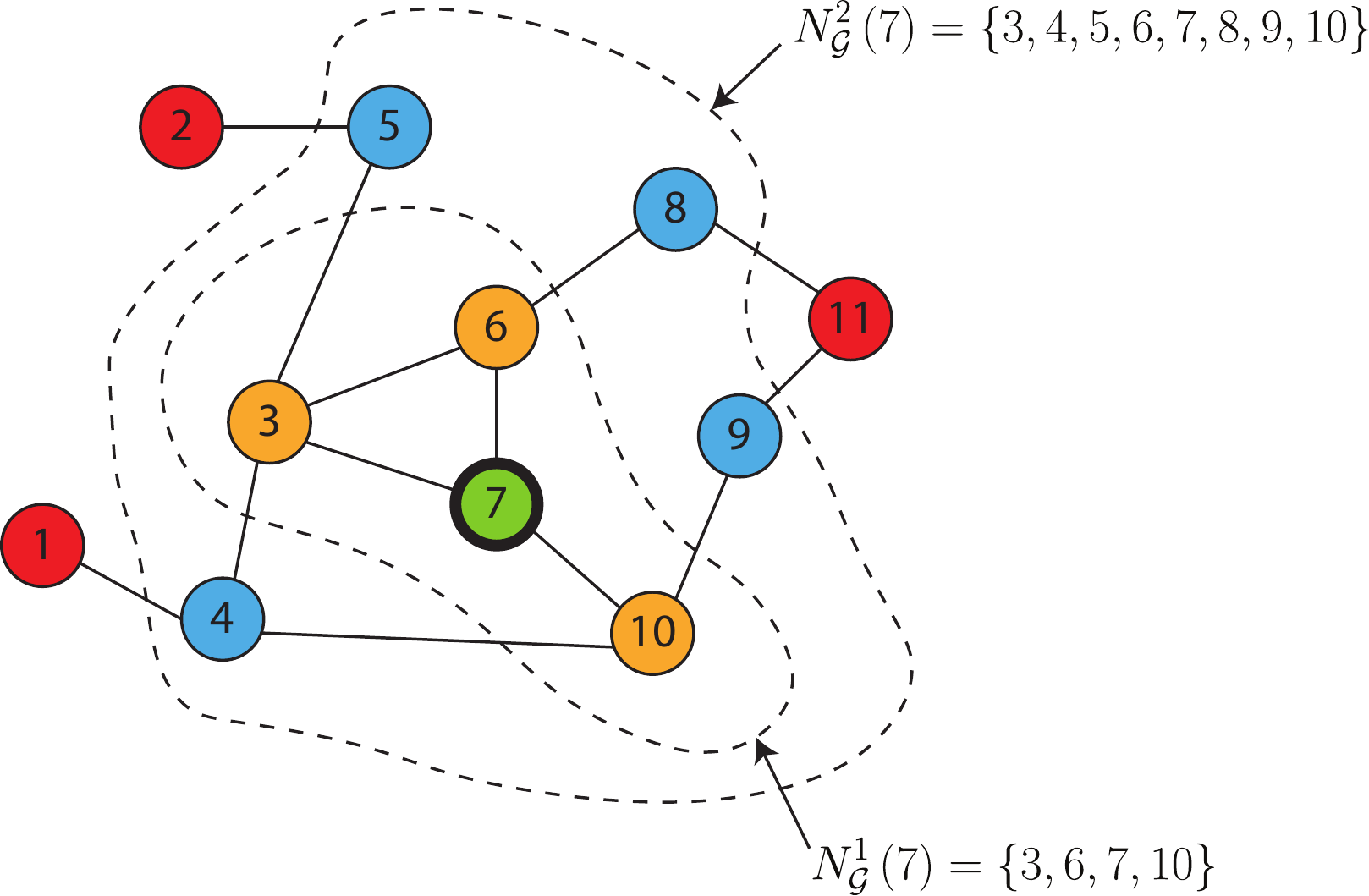}
    \caption{Node neighborhood using $\mathtt{shortest\_path}$ distance. Neighbors around node 7 are color-coded based on their corresponding $\mathtt{shortest\_path}$ distance layer. Orange color represents directly adjacent nodes, blue color highlights nodes at 2 hops from node 7, and red color labels the remaining nodes in the graph.}\label{node_neighborhood}
\end{figure}

Fig.~\ref{node_neighborhood} illustrates the node neighborhood using a $\mathtt{shortest\_path}$ distance.

\subsection{The Geometric Model}
By implementing the network constraints, we aim at avoiding the configurations where a data point $i$ chooses $k$ as its exemplar (i.e., $c_{i}=k$) while $k \notin N_{\mathcal{G}}^{\tau}\left ( i  \right )$.
Towards this end, we amend the penalty term $\delta _{k}\left ( \mathbf{c} \right )$ in (\ref{eq8}) to a new penalty term $\gamma _{k}\left ( \mathbf{c} \right )$ that takes the form:
\begin{align}\label{eq10}
  \gamma _{k}\left ( \mathbf{c} \right )=\begin{cases}
-\infty,& \mbox{if } c_{k} \neq k \mbox{ but }\exists\ i:\ c_{i}=k, \\
-\infty,& \mbox{if } \exists\ i:\ c_{i}=k \mbox{ but } k \notin N_{\mathcal{G}}^{\tau}\left ( i  \right ),\\
 0,& \mbox{otherwise}.
\end{cases}
\end{align}
The net similarity $\mathcal{S}$ in (\ref{eq7}) becomes:
\begin{equation}\label{eq11}
    \mathcal{S}\left ( \mathbf{c} \right ) =\sum_{i=1}^{N}s\left ( i,c_{i} \right )+\sum_{k=1}^{N}\gamma _{k}\left ( \mathbf{c} \right ).
\end{equation}
In addition to maximizing the within-cluster feature-based similarity, this new formulation intuitively maximizes the within-cluster topological similarity. As a result, the optimization task jointly searches for valid configurations that account for both feature-based and network-based similarities. From this perspective, geometric-AP can be viewed as a more constrained special case of AP that is NP-hard. As the max-sum belief propagation algorithm is used, the obtained solution is guaranteed to be the neighborhood maximum~\citep{Weiss2001} simultaneously congruent with both topological and feature spaces. Thus, assigning data points to close exemplars based only on node features becomes inadequate. To solve this dilemma, we straiten the search space of proximal exemplars for a given point $i$ to the local neighborhood $N_{\mathcal{G}}^{\tau}\left ( i  \right )$. When no exemplar exists in $N_{\mathcal{G}}^{\tau}\left ( i  \right )$, the point $i$ is allowed to select among the full list of emerged exemplars. Clearly, this assignment policy may raise some misassignments when the exemplar selection occurs outside the local neighborhood. To remedy this deficiency, we smooth the labels throughout the network using adjacency majority voting. The underlying rationale for label smoothing is that we are, in principle, more confident about the predicted labels of the adjacent neighbors of any given node than the estimated label of the single node itself. Also, the internal density characteristic of graph clusters stipulates that highly interacting nodes are more likely to share the same label. For labeling purposes, the network adjacency of any given point $i$ with respect to a graph $\mathcal{G}$, denoted by $\mathcal{A}_{\mathcal{G}}\left ( i\right)$, can be viewed as an $\alpha$-cover of the reduced graph formed by $i$ and $\mathcal{A}_{\mathcal{G}}\left ( i\right)$. For instance, in~\citep{Guillory2009} the authors provided a label selection strategy using graph coverings and have derived an upper-bound expression for the error committed by majority voting in binary labeled graphs.
\begin{figure}[!ht]
\centering
\begin{subfigure}{0.35\textwidth}
  \includegraphics[width=\linewidth]{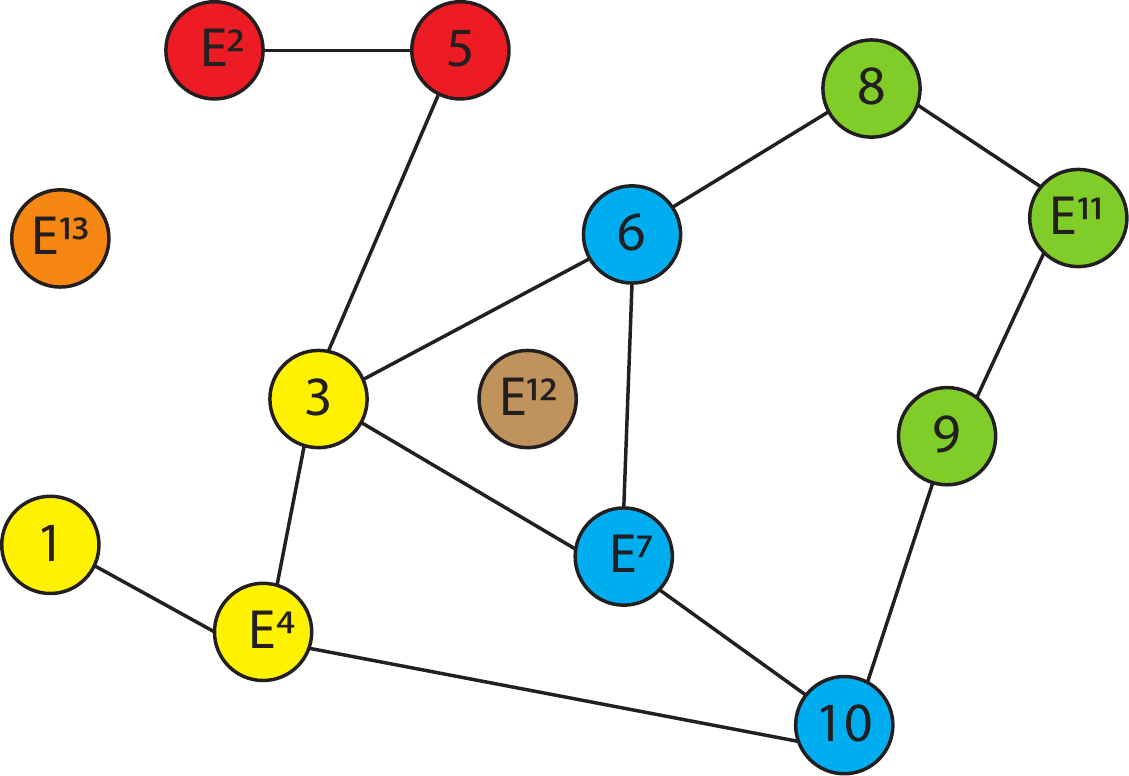}
  \caption{Geometric-AP}
  \label{fig:synthetic_gap}
\end{subfigure}
\quad \quad \quad \quad \quad
\begin{subfigure}{0.35\textwidth}
  \includegraphics[width=\linewidth]{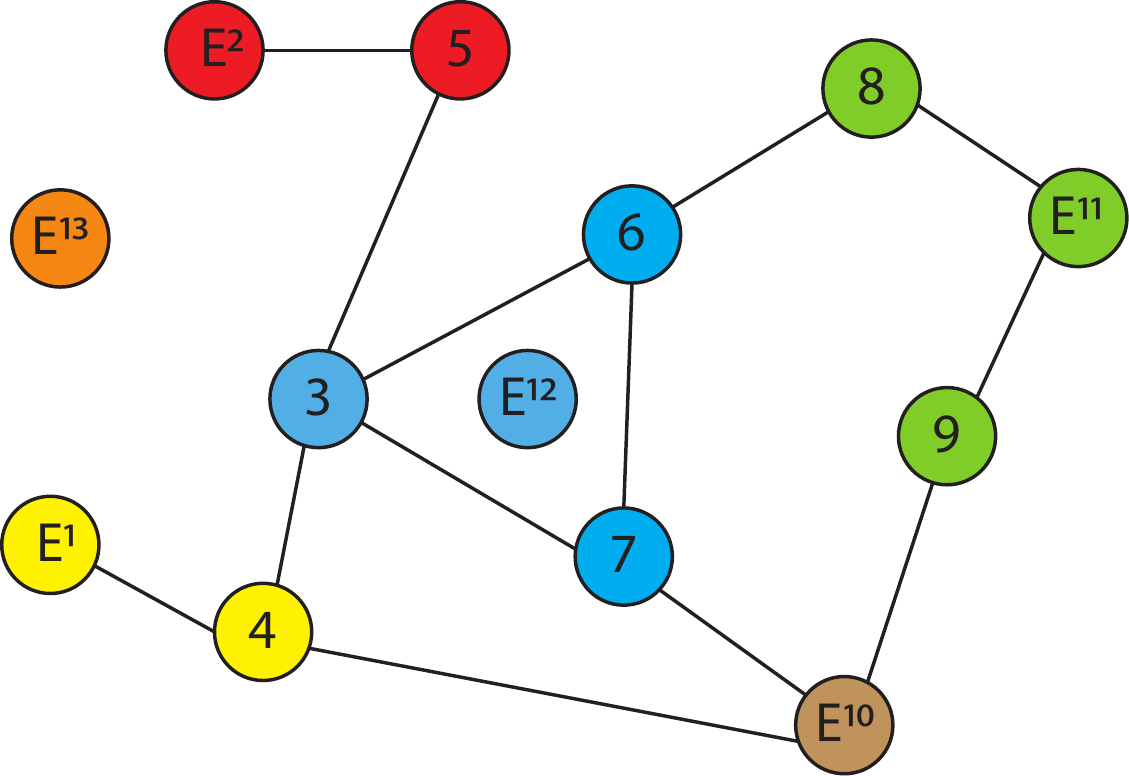}
  \caption{AP}
  \label{fig:synthetic_ap}
\end{subfigure}
\caption{Geometric-AP versus AP. Identified clusters are color-coded. The shared preferences for AP and geometric-AP are preselected to get the same number of clusters. The node features are the plane coordinates. Geometric-AP has been launched with the $\mathtt{shortest\_path}$ distance and a neighborhood threshold of 2 ($\tau=2$). Cluster exemplars are labeled by $E^{ID}$.}
\label{fig:synthetic}
\end{figure}

Fig.~\ref{fig:synthetic} illustrates the clustering properties of geometric-AP as compared to AP when applied to one synthetic dataset. Obviously, geometric-AP is more robust against outliers and generates better connected modules in the network. In contrast, AP is hypersensitive to an unconnected vertex (node 12) as it only relies on feature similarities. Additionally, the exemplars identified by geometric-AP occupy more centric locations in the network when compared to the ones selected by AP.

\subsection{Topological Neighborhood}
geometric-AP greatly depends on the neighborhood function $N_{\mathcal{G}}^{\tau}$ defined in (\ref{eq9}). In order to probe the effect of the topological distance ``$\mbox{distance}_{\mathcal{G}}$" on the performance of geometric-AP, we consider throughout this paper three widely used topological distance metrics. We evaluate the performance of our geometric model using the Jaccard, cosine, and shortest path distances. Unless stated otherwise, we consider that the available network knowledge takes the form of an undirected graph $\mathcal{G}=\left( \mathcal{V},\mathcal{E}\right)$ where $\mathcal{V}$ is the set of vertices of size $N$ mapped to the observed data points in the feature space (i.e. every data point is mapped to a unique vertex in $\mathcal{G}$). $\mathcal{E}$ represents the set of undirected edges in the graph $\mathcal{G}$.

\subsubsection{Jaccard Distance}
The Jaccard distance is derived from the Jaccard index defined for two sets A and B as:
\begin{equation}\label{eq12}
    \rho\left ( A,B \right )=\frac{\left | A \right |\cap \left | B \right |}{\left | A \right |\cup  \left | B \right |}.
\end{equation}
From a topological viewpoint, we characterize every vertex $V\in\mathcal{V}$ by an $M$-dimensional binary vector $V=\left(v_{1}, v_{2}, ..., v_{M} \right)$ such that $v_{i_{\left (i=1..M \right )}}=1$ if vertex $V$ and $V_{i}$ are connected in $\mathcal{G}$ and 0 otherwise.
The Jaccard distance between two vertices $V_{1}=\left(v_{11}, v_{12}, ..., v_{1M} \right)$ and $V_{2}=\left(v_{21}, v_{22}, ..., v_{2M} \right)$ $\in\mathcal{V}$ is then defined as:
\begin{align}\label{eq13}
       \nonumber \mbox{distance}_{\mathcal{G}}^{Jaccard}\left ( V_{1},V_{2} \right ) &= 1-\rho\left ( V_{1},V_{2} \right )\\
        &= \frac{C_{1,0}+C_{0,1}}{C_{1,0}+C_{0,1}+C_{1,1}},
\end{align}
where $C_{i,j}$ is the number of positions $k\in\left[ 1..M\right]$ in which $v_{1k}=i$ and $v_{2k}=j$.
\subsubsection{Cosine Distance}
The cosine distance between two vertices $V_{1}=\left(v_{11}, v_{12}, ..., v_{1M} \right)$ and $V_{2}=\left(v_{21}, v_{22}, ..., v_{2M} \right)$ $\in\mathcal{V}$ is given by:
\begin{align}\label{eq14}
    \mbox{distance}_{\mathcal{G}}^{cosine}\left ( V_{1},V_{2} \right )=1-\frac{V_{1}.V_{2}}{\sqrt{\sum_{k=1}^{M}v_{1k}^{2}}.\sqrt{\sum_{k=1}^{M}v_{2k}^{2}}}.
\end{align}
\subsubsection{Shortest Path Distance}
The shortest path distance between two vertices $V_{1}$ and $V_{2}$ $\in\mathcal{G}$ is universally defined as the shortest sequence of edges in $\mathcal{E}$ starting at vertex $V_{1}$ and ending at vertex $V_{2}$. Axiomatically, the distance from a vertex to itself is zero and the path from a vertex to itself is an empty edge sequence~\citep{Schaeffer2007}. Although the shortest path problem is P-complete (i.e. solvable in a polynomial time), the computational complexity of geometric-AP may significantly deteriorate as we repeatedly compute pairwise distances between all graph vertices. However, to fully determine the neighborhood function $N_{\mathcal{G}}^{\tau}$ we need only to know about the existence of shortest paths with specified lengths but not the full sequence of edges. This observation leads to an efficient implementation of the neighborhood function $N_{\mathcal{G}}^{\tau}$ based on the shortest path distance. Indeed, for any vertex $i$, the determination of $N_{\mathcal{G}}^{\tau}\left ( i  \right )$ is straightforward if we observe that the $\nu^{th}$ power of a graph $\mathcal{G}$ is also a graph with the same set of vertices as $\mathcal{G}$ and an edge between two vertices if and only if there is a path of length at most $\nu$ between them. If we denote by $\mathcal{A(G)}$ the adjacency matrix of graph $\mathcal{G}$ and by $\mathcal{A(G}^{\nu})$ the adjacency matrix of graph $\mathcal{G}^{\nu}$, the entries of $\mathcal{A(G}^{\nu})$ are derived using an indicator function as follows:
\begin{equation}\label{eq15}
  \mathcal{A(G}^{\nu}) = \mathbb{1}_{
  \begin{bmatrix}
    \sum_{i=1}^{\nu}\mathcal{A(G)}^{i}
  \end{bmatrix}},
\end{equation}
where the indicator function $\mathbb{1}$ of matrix $X$ with entries $x_{ij}$, is a matrix of the same dimension as $X$ and entries defined by: $\mathbb{1}_{X}\left [ i,j \right ]=1$ if $x_{ij}\neq 0$ and $\mathbb{1}_{X}\left [ i,j \right ]=0$ otherwise.
\subsection{Optimization}
\begin{figure}[!ht]
    \centering
    \includegraphics[scale=0.9]{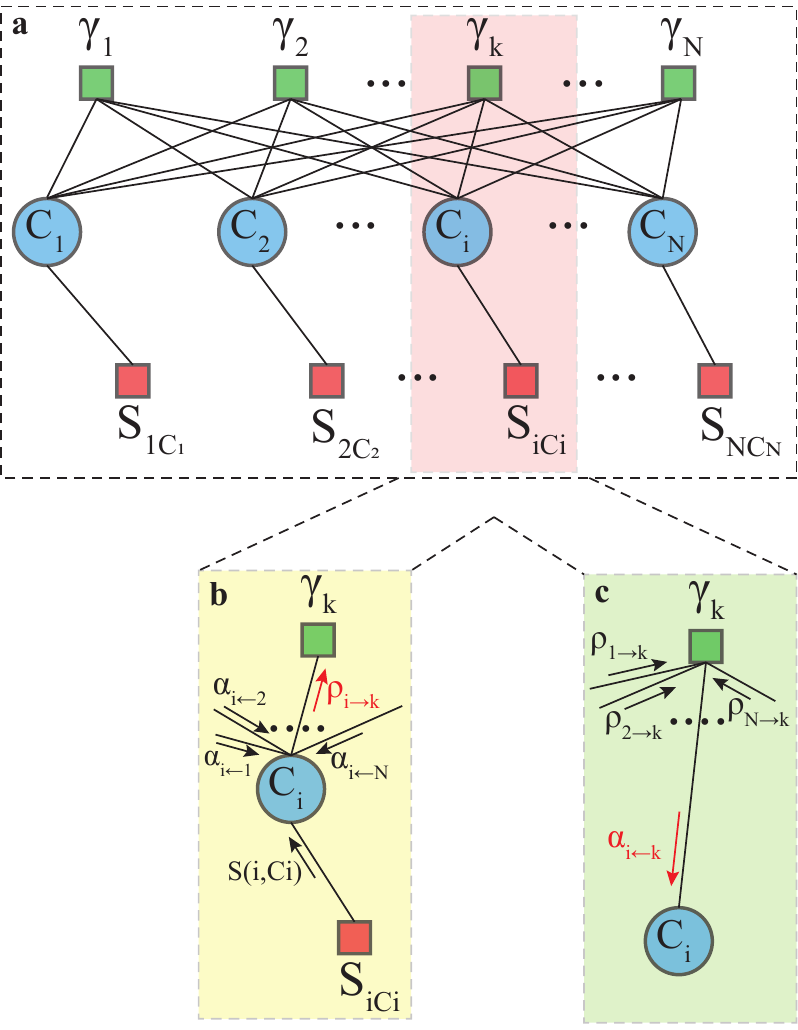}
    \caption{Factor graph for geometric-AP}
    \label{fig:factor}
\end{figure}

The optimization of the objective function introduced in geometric-AP is NP-hard. This follows from the fact that geometric-AP is a network-constrained version of AP, which is known to be NP-hard. In order to estimate the optimal label configuration we follow similar derivation as the one introduced in~\citep{Frey2007}. Thus, we solve the optimization problem using max-sum belief propagation over the factor graph depicted in Fig.~\ref{fig:factor}. 
We note that the function node in Fig.~\ref{fig:factor} is different from the one presented in~\citep{Frey2007} as it leverages the network information to account for the topological similarity. 

In the max-sum belief propagation algorithm, a bipartite message communication between two types of nodes is conducted with an alternation between summation and maximization steps as illustrated in Fig.~\ref{fig:factor}.a. The first type of nodes is called variable node and it sums up the received messages from all second type nodes, called function nodes, other than the one receiving the message (Fig.~\ref{fig:factor}.b). Likewise, every function node maximizes its value over all the variables except the variable the message is being sent to (Fig.~\ref{fig:factor}.c). Besides its computational merits, the max-sum algorithm has set the record in solving highly constrained search problems. We next provide the set of derived message updates that govern geometric-AP.
\subsubsection{Message Updates}
With analogy to AP, the message sent from variable node $c_{i}$ to function node $\gamma_{k}$ sums together all the messages received from the remaining function nodes. As shown in Fig.~\ref{fig:factor}.b, this message is denoted by $\rho_{i\rightarrow k}$ and takes the form:
\begin{align}\label{eq16}
    \rho_{i\rightarrow k}\left ( c_{i} \right )=s\left ( i, c_{i}\right )+\sum_{k':k'\neq k}\alpha_{i\leftarrow k'}\left ( c_{i} \right ).
\end{align}

Similarly, the message sent from function node $\gamma_{k}$ to variable node $c_{i}$ computes the maximum over all variable nodes except $c_{i}$ (Fig.~\ref{fig:factor}.c) and can be given by:
\begin{align}\label{eq17}
    \alpha_{i\leftarrow k}\left ( c_{i} \right )=\max_{\left ( c_{1},c_{2},...,c_{i-1},c_{i+1},...,c_{N} \right )}
    \left [ \gamma_{k}\left (  c_{1},c_{2},...,c_{i-1},\mathbf{c_{i}},c_{i+1},...,c_{N}\right ) +\sum_{i':i'\neq i}\rho_{i'\rightarrow k}\left ( c_{i'} \right )\right ].
\end{align}

Using a set of mathematical simplifications utilized in~\citep{Frey2007} we derive two message updates called also \emph{responsibility} and \emph{availability}. Thus, the responsibility message, denoted by $r\left(i,k \right)$, replaces the message $\rho_{i\rightarrow k}$ and the availability message designated by $a\left(i,k \right)$ substitutes the message $\alpha_{i\leftarrow k}$ as shown in Fig.~\ref{fig:factor}.
Ultimately, the simplified messages are given by (\ref{eq18}) and (\ref{Eq19}).
\begin{align}\label{eq18}
     r\left ( i,k \right ) = s\left ( i,k \right )-\max_{j:j \neq k}\left [ s\left ( i,j \right )+a\left ( i,j \right ) \right ].
\end{align}
\begin{align}\label{Eq19}
 a\left ( i,k \right ) =\tilde{\alpha}_{i \rightarrow k}\left ( c_{i}=k \right ) = \begin{cases}       
\sum_{i':i' \neq k}\max\left ( 0,r\left ( i',k \right ) \right ), & \mbox{if}\ k=i ,
\\
\min\left ( 0,r\left ( k,k \right )+\sum_{i':i' \notin \left \{ i,k \right \}}\max\left ( 0,r\left ( i',k \right ) \right ) \right ), & \mbox{if}\ k\neq i\ \&\ k \in N_{\mathcal{G}}^{\tau}\left ( i  \right ),
 \\
-\max\left ( 0,r\left ( k,k \right )+\sum_{i':i' \notin \left \{ i,k \right \}}\max\left ( 0,r\left ( i',k \right ) \right ) \right ), & \mbox{if}\ k\neq i\ \&\ k \notin N_{\mathcal{G}}^{\tau}\left ( i  \right ).
\end{cases}
\end{align}
The detailed derivation of message updates for geometric-AP is provided in appendix A.
Obviously, the responsibility message remained unchanged as compared to AP. However, the availability message has accommodated the network constraints and contained a lower level update that takes into account the network neighborhood between data points and potential exemplars. This new message delivers additional information about the accumulated evidence for how well suited it would be for a point to select an exemplar by taking into consideration both, the information collected from other points and the topological gap between that point and the candidate exemplar. Thereby, every exemplar encourages data points to select candidates among their proximal neighborhood with respect to the network. These observations are reflected in the message update in (\ref{Eq19}) since the availability message remains unchanged when the candidate exemplar falls within the topological neighborhood of the communicating data point but takes a new expression whenever the potential exemplar is located outside the network proximity of the data point of interest.
This new formulation provides a simple, yet effective, theoretical approach that jointly combines two different sources of similarities to conduct a more rigorous search over the domain of valid configurations. Furthermore, we will discuss an updated assignment policy that consolidates the underlying concepts of the derived messages.
\subsubsection{Assignment of Clusters}
At any given iteration of AP, the value of a variable node $c_{i}$ can be estimated by summing together all messages that $c_{i}$ receives. Subsequently, the argument that maximizes these incoming messages, denoted by $\hat{c}_{i}$,  will be a good estimate for $c_{i}$~\citep{Frey2007}.
\begin{align}\label{eq20}
    \hat{c}_{i}= \arg\max_{j}\left [ a\left (i,j \right )+s\left ( i,j \right ) \right ].
\end{align}
In geometric-AP, this rule is amended to be consistent with the joint similarity criteria respected during the derivation of the message updates. Indeed, geometric-AP prioritizes the assignment of data points to the closest exemplar that lies withing the local neighborhood of each data point. As the number of emerging exemplars is automatically determined by the algorithm, some proximal exemplars may breach the topological constraint and mislead the membership assignment as illustrated in the example with the synthetic dataset in Fig.~\ref{fig:synthetic}. To remedy this deficiency, we prioritize at a first stage the selection among close exemplars that fall within the topological sphere determined by $N_{\mathcal{G}}^{\tau}$. However, geometric-AP does not guarantee that at least one exemplar emerges in the network neighborhood of every data point. Thus, we allow at a second stage a more lenient selection among all available exemplars. This selection protrudes as a best decision policy.
The new updated rule takes then the form: 
\begin{align}\label{eq21}
    \hat{c}_{i}= \operatorname*{arg~max}_{
  \scriptsize{  \begin{cases}
\left \{j:~j\in N_{\mathcal{G}}^{\tau} \left ( i \right )~\&~c_{j}=j  \right \}
, &\mbox{if~} \left \{ \exists~k \in N_{\mathcal{G}}^{\tau}\left ( i \right )
:c_{k}=k \right \}
,\\
 j, &\mbox{otherwise}.
\end{cases}}\normalsize
}\left [ a\left (i,j \right )+s\left ( i,j \right ) \right ].
\end{align}\normalsize

\subsubsection{Label Smoothing}
In belief propagation, the algorithm has been proven to converge to the global optimum for trees but to the maximum neighborhood for arbitrary graphs~\citep{Weiss2001}. Thus, search errors are naturally expected to occur when dealing with arbitrary graphs. This fact together with the misassignments that may happen in geometric-AP, particularly when the exemplar selection takes place outside the local neighborhood, reduces the performance of the carried out clustering. To mitigate these issues we adopt a label smoothing strategy that has been widely used in label selection on graphs~\citep{Guillory2009}. Our choice has been motivated also by the successful use of label smoothing in many fields such as image segmentation and deep learning classification~\citep{Schindler2012,Muller2019}. In label selection on graphs, one successful application of label smoothing involves the notion of graph covering and uses $\alpha$-cover sets to label the remaining nodes in the graph by majority vote~\citep{Guillory2009}. By definition, we say that a set $S$ $\alpha$-covers a graph $\mathcal{G}=\left( \mathcal{V},\mathcal{E}\right)$ if $\forall i \in \mathcal{V} \mbox{~either~} i\in S \mbox{~or~}\sum_{j\in S}W_{ij}>\alpha$ where $W_{ij}$ denotes the weight on the edge between vertex $i$ and $j$. For unweighted graphs $W_{ij}$ takes a binary value. Realistically, a vertex $V$ in $\mathcal{G}$ can be labeled efficiently by majority vote if some voting nodes are adjacent to $V$ w.r.t $\mathcal{G}$.
From this perspective, we target all adjacent voters and we consider the network adjacency of any given point $i$ w.r.t $\mathcal{G}$, denoted by $\mathcal{A}_{\mathcal{G}}\left ( i\right)$, to form a reduced graph formed by $i$ and $\mathcal{A}_{\mathcal{G}}\left ( i\right)$. In this reduced graph, for $\alpha=1$, $\mathcal{A}_{\mathcal{G}}\left ( i\right)$ can be viewed as an $\alpha$-cover for $\mathcal{A}_{\mathcal{G}}\left ( i\right) \cup i$. Subsequently, we take advantage of the concept of graph covering  and we perform an inclusive label smoothing throughout the full network in a way to prevent geometric-AP from being overconfident. We propose the following label smoothing policy:
\begin{equation}\label{eq22}
    \hat{c}_{i}=\arg\max_{k}\left [ \sum_{j\in \mathcal{A}_{\mathcal{G}}\left ( i \right )}\delta \left ( c_{j}=c_{k} \right ) \right ],
\end{equation}
where $\delta \left(.\right)$ is the Dirac delta function, such that the sum in (\ref{eq22}) counts the number of vertices in $\mathcal{A}_{\mathcal{G}}\left ( i \right )$ that have class $c_{k}$. In order to comprehensively handle all vertices in $\mathcal{G}$, including disconnected nodes, geometric-AP counts the vote of the vertex $i$ as well.
\subsection{Comparison to Affinity Propagation}
Geometric-AP leverages the available network information by implementing a more constrained optimization problem as compared to AP. This implementation assumes that significant clusters are jointly compact in two different domains that are the node-feature domain and the network topological domain. Under this assumption, geometric-AP is expected to be successful in boosting the clustering performance if and only if the information carried by the node features and the network topology about the structure of the clusters is consistent. From this standpoint, the network information can be viewed as a chaperone for the clustering task to achieve a more significant partitioning of the data by avoiding some local optimum traps. To further elucidate this notion, we rewrite the message updates in (\ref{Eq19}) using the identity:
\begin{equation}\label{eq23}
    x-\max\left ( 0,x \right )=\min\left ( 0,x \right ),
\end{equation}
which leads to the expression given in (\ref{Eq24}).
\begin{align}\label{Eq24}
a\left ( i,k \right )  =\tilde{\alpha}_{i \rightarrow k}\left ( c_{i}=k \right ) = \begin{cases}
 \sum_{i':i' \neq k}\max\left ( 0,r\left ( i',k \right ) \right ), &\ \mbox{if}\ k=i,
\\
\\
\min\left ( 0,r\left ( k,k \right )+\sum_{i':i' \notin \left \{ i,k \right \}}\max\left ( 0,r\left ( i',k \right ) \right ) \right ), & \ \mbox{if}\ k\neq i\ \&\ k \in N_{\mathcal{G}}^{\tau}\left ( i  \right ),
\\
\\
\min\left ( 0,r\left ( k,k \right )+\sum_{i':i' \notin \left \{ i,k \right \}}\max\left ( 0,r\left ( i',k \right ) \right ) \right )\\
-~\left [ r\left ( k,k \right )+\sum_{i':i' \notin \left \{ i,k \right \}}\max\left ( 0,r\left ( i',k \right ) \right )\right ], &\ \mbox{if}\ k\neq i\ \&\ k \notin N_{\mathcal{G}}^{\tau}\left ( i  \right ).
\end{cases}
\end{align}
This new result outlines the difference between geometric-AP and AP as a penalty term deducted from the availability message sent from the potential exemplar $k$ to the data point $i$ when $k \notin N_{\mathcal{G}}^{\tau}\left ( i  \right )$. The expression of the penalty term is given by:
\begin{equation}\label{eq25}
    r\left ( k,k \right )+\sum_{i':i' \notin \left \{ i,k \right \}}\max\left ( 0,r\left ( i',k \right ) \right ).
\end{equation}
Except the time required to compute the neighborhood function $N_{\mathcal{G}}^{\tau}$, the expression provided in (\ref{Eq24}) sets the computational complexity of geometric-AP to $\mathcal{O}(N^2)$ since the expression of the penalty term is already computed and is reusable at no computational cost. All these advantages make the implementation of the proposed geometric-AP as efficient as that of AP while keeping the benefits carried by the network information.
\section{Unsupervised Document Clustering}
\label{sec:Unsupervised_document_Clustering}
We thoroughly study, in this section, the improvement of geometric-AP with reference to AP in unsupervised document classification on two benchmark citation networks, that are the \emph{\textbf{cora}} dataset~\citep{McCallum2000} and the \emph{\textbf{citeseer}} dataset~\citep{Giles1998}. Additionally, we select and perform a variety of clustering methods that span many state-of-the-art clustering approaches for comparison purposes.
Our findings show that geometric-AP consistently outperforms AP on the studied datasets and exhibits high competitiveness with other long-standing and popular methods.
\subsection{Methods and Settings}
\label{Methods_and_Settings}
The list of clustering methods selected to benchmark geometric-AP along with their tuned hyper-parameters are detailed as follows:
\begin{enumerate}
    \item Exemplar-based clustering methods. The selected competing methods are kmedoids~\citep{Kaufman1990} and AP~\citep{Frey2007}. We denote the convergence parameters of AP and geometric-AP by $\mbox{max}_{\mbox{iter}}$, $\mbox{conv}_{\mbox{iter}}$, and $\lambda$ to designate the maximum number of iterations, the number of iterations for convergence, and the message damping factor, respectively. We choose values reported to guarantee high convergence rates~\citep{Dueck2009}. $\mbox{max}_{\mbox{iter}}=1000$, $\mbox{conv}_{\mbox{iter}}=100$, and $\lambda=0.9$. kmedoids relates to the kmeans~\citep{Hartigan1979} algorithm but it identifies the medoid of each cluster by minimizing the sum of distances between the medoid and data points instead of sum-of-squares. Unlike centroids, medoids are selected from the existent data points.
    \item Centroid-based clustering. The most popular method, that is kmeans~\citep{Hartigan1979}, is performed. kmeans is run 1000 times with random centroid seeds and the best performance is reported.
    \item Structural clustering. The clustering method \emph{spectral-g}~\citep{Luxburg2007}, which takes the network adjacency matrix as the similarity matrix, is selected and compared. In \emph{spectral-g}, the eigenvectors of the graph Laplacian are computed and the kmeans algorithm is used to determine the clusters. The assignment process is repeated 1000 times with random initialization and the best result is recorded.
    \item Hierarchical clustering. To mimic the operating mode of AP, where initially all data points can be exemplars, we select a bottom-up hierarchical clustering method that is the hierarchical agglomerative clustering (HAC)~\citep{Ward1963}. To prioritize compact clusters with small diameters we further consider the complete linkage criterion to merge similar clusters.
    \item Model-based clustering. We also test a Gaussian mixture model (GMM)~\citep{McLachlan1988} method that utilizes the Expectation-Maximization (EM) algorithm to fit a multi-variate Gaussian distribution per cluster. Initially, the probability distributions are centered using kmeans and then EM is used to find local optimal model parameters using full covariances. The mixture model is employed afterwards to assign data points to the class that maximizes the posterior density. 100 random restarts are performed and the best performance is reported.
    \item Variational inference clustering. We perform a Bayesian variational inference clustering by fitting a Gaussian mixture model with an additional regularization from a prior Dirichlet process distribution (DPGMM)~\citep{Blei2006}. Similar to GMM, 100 random restarts with full covariances are performed and the best result is reported.
\end{enumerate}
geometric-AP is implemented in 64-bit python 3.6.8 on a workstation (Windows 64 bit, 2.8 GHz Intel Core i7-7700HQ CPU, 16 GB of RAM).
\subsection{Similarity Metrics}
\label{similarity_metric}
Many methods have been devised in the past few decades to provide vector representations for textual data~\citep{Wong1992,Schaeffer2007}. Most popular representations that have been extensively used in the literature include the binary word vector and the term-frequency inverse-document-frequency (tf-idf)~\citep{Wong1992} representations. Distance measures that have been reported as congruent with these representations include the Euclidean, Manhattan, and cosine distances as reviewed in~\citep{Schaeffer2007}. For unbiased comparison, we independently run the kmedoids algorithm using the aforementioned distances on the \emph{\textbf{cora}} and \emph{\textbf{citeseer}} datasets to predict the ground-truth class labels and we retain the distance measure that gives the best clustering result on each dataset. In an $M$-dimensional space, the considered distances between two data points $p=\left(p_{1}, p_{2},...,p_{M}  \right)$ and $q=\left(q_{1}, q_{2},...,q_{M}  \right)$ are defined as follows:
\begin{itemize}
    \item Euclidean Distance: $dist_{Euc}\left(p,q\right)=\sum_{i=1}^{M}\sqrt{\left(p_{i}-q_{i}\right)^{2}}$.\normalsize
    \item Manhattan Distance: $dist_{Man}\left(p,q\right)=\sum_{i=1}^{M}\left|p_{i}-q_{i}\right|$.\normalsize
    \item Cosine Distance: $dist_{Cos}\left(p,q\right)=\frac{p.q}{\sqrt{\sum_{i=1}^{M}\left( p_{i}\right )^{2}}\sqrt{\sum_{i=1}^{M}\left( q_{i}\right )^{2}}}$.\normalsize
\end{itemize}
Consequently, the off-diagonal elements of the similarity matrix $\left [s_{ij}\right ]_{N\times N}$ between $N$ data points are defined as:
\begin{equation}\label{eq26}
    s_{ij} = - dist_{metric}(i,j),
\end{equation}
where $dist_{metric}$ refers to one of the previously discussed distance measures. The preference values $s_{ii}$ are controlled over a range of values to generate different number of clusters.
\subsection{Clustering Evaluations}
In the absence of a unified criterion universally accepted for assessing clustering performance, many evaluation metrics have been proposed. The list of popular metrics include, but not limited to, average purity, entropy, and mutual information~\citep{Strehl2000}. More recently, mutual information measures become accepted with appreciation by the research community as they provide a plausible evaluation of the information shared between the compared clusterings. We adopt three widely used performance measures as discussed in~\citep{Aggarwal2013} that are: normalized mutual information (NMI), classification rate (CR), and macro F1-score (F1).
\subsubsection{Normalized Mutual Information (NMI)}
Given a dataset $\mathcal{D}$ of size $n$, the estimated clustering labels $\Omega$ of $\theta$ clusters and the true class labels $\hat{\Omega}$ of $\hat{\theta}$ classes, a matching matrix $\mathcal{M}$ is computed. $\mathcal{M}$ has entries $m_{ij}$ that specifies how many points in cluster $i$ have the label $j$. The NMI can be derived from $\mathcal{M}$ as follows:
\begin{equation}\label{eq27}
    NMI=\frac{2\sum_{l=1}^{\theta}\sum_{h=1}^{\hat{\theta}}\frac{m_{lh}}{2}\log \frac{m_{lh}\times n}{\sum_{i=1}^{\theta}m_{ih}\sum_{i=1}^{\hat{\theta}}m_{li}}}{H\left ( \Omega \right )+H\left ( \hat{\Omega} \right )},
\end{equation}
where:
\begin{itemize}
    \item $H\left ( \Omega \right )=-\sum_{i=1}^{\theta}\frac{n_{i}}{n}\log\frac{n_{i}}{n}$
    \item $H\left ( \hat{\Omega} \right )=-\sum_{i=1}^{\hat{\theta}}\frac{n^{(i)}}{n}\log\frac{n^{(i)}}{n}$
\end{itemize}
are the Shannon entropy of cluster labels $\Omega$ and class labels $\hat{\Omega}$, respectively.
In the entropy expressions, $n_{i}$ denotes the number of points in cluster $i$, while $n^{(i)}$ designates the number of points that belong to class $i$. Large NMI values reflect a better match between the clustering and the class labels.
\subsubsection{Classification Rate (CR)}
By analogy to the work performed in~\citep{Wang2013}, we compute CR by associating the members of each predicted cluster with the ground-truth class label that appears the most in the cluster. Then, CR is given by the ratio of correctly classified points to the total number of data points in the dataset: 
\begin{equation}\label{eq28}
    CR = \frac{\mbox{number of correctly classified points}}{\mbox{total number of points in the dataset}}\times 100\%.
\end{equation}
Clearly, a high classification rate implies a good clustering accuracy.
\subsubsection{Macro F1-score (F1)}
Using the same labeling scheme proposed for CR, the macro F1-score is defined as the arithmetic mean of the per-class F1-scores denoted by F1-score(c) for $c \in \left \{ 0,..,\mathcal{C} \right \}$ where $c$ is the class index and $\mathcal{C}+1$ designates the total number of classes. F1-score(c) is given by:
\begin{equation}\label{eq29}
    \mbox{F1-score}\left(c\right)=2\times \frac{\mbox{precision} \times \mbox{recall}}{\mbox{precision} + \mbox{recall}}
\end{equation}
Using the true positives (TP), true negatives (TN), false positives (FP), and false negatives (FN) terminology the precision and recall are expressed as follows:
\begin{itemize}
    \item $\mbox{precision}=\frac{TP}{TP+FP}$,
    \item $\mbox{recall}=\frac{TP}{TP+FN}$.
\end{itemize}
As such, the macro F1-score can be expressed as:
\begin{equation}\label{eq30}
    \mbox{macro F1-score}=\frac{1}{\mathcal{C}}\sum_{c=0}^{\mathcal{C}}\mbox{F1-score}\left ( c \right ).
\end{equation}
A high macro F1-score indicates a good match between the estimated labels and the ground-truth classes.

\subsection{Performance Assessment Results}
\subsubsection{Cora Dataset}
\label{subsec:Cora}
The \emph{\textbf{cora}} dataset~\citep{McCallum2000} contains 2708 machine learning papers from seven classes and 5429 links between them. The links indicate a citation relationship between the papers. Each document is represented by a binary vector of 1433 dimensions marking the presence of the corresponding word. The documents in \emph{\textbf{cora}} are short texts extracted from titles and abstracts where the stop words and all words with document frequency less than 10 are removed~\citep{Yang2015}. Stop words include non-informative words like articles and prepositions and are filtered out to avoid inflating the dimensions. Each document in \emph{\textbf{cora}} has on average 18 words and the network is regarded, in the context of this work, as an undirected graph.

After applying the selection procedure described in section~\ref{similarity_metric}, the most appropriate similarity metric on the \emph{\textbf{cora}} dataset has been identified as the negative Euclidean distance.
Next, we plot in Figs.~\ref{fig:subfigurea1},~\ref{fig:subfigureb1}, and~\ref{fig:subfigurec1} the clustering results of geometric-AP with different neighborhood functions $N_{\mathcal{G}}^{\tau}$ as a function of $\tau$ when the ground-truth classes are being considered. In this experiment we aim at tuning $N_{\mathcal{G}}^{\tau}$ defined in (\ref{eq9}) by identifying the best topological distance ``$\mbox{distance}_{\mathcal{G}}$" and the optimal threshold value $\tau$ that lead to the top clustering result w.r.t NMI, CR, and F1. In case of conflicts or ties, the reference metric for identifying the optimal threshold is always the NMI and the smallest optimal threshold is retained. For each topological distance, we run the algorithm with $\tau$ ranging from 1 to 5 for the shortest path distance and from 0.5 to 0.9 for the Jaccard and cosine metrics when the actual classes are being used, i.e., 7 in the case of \emph{\textbf{cora}}. As illustrated in Fig.~\ref{fig:subfigurea1}, the optimal clustering results are obtained using the $\mathtt{shortest\_path}$ distance with a neighborhood threshold $\tau=3$.
\input{figures/cora}
In the remainder of this discussion about the \emph{\textbf{cora}} dataset the neighborhood function $N_{\mathcal{G}}^{\tau}$ will be defined as follows:
\begin{equation}\label{eq31}
N_{\mathcal{G}}^{\tau}\left ( i  \right ) = \left \{ x: \mathtt{shortest\_path}\left( x, i\right) \leq 3\right \}.
\end{equation}
We show in Figs.~\ref{fig:subfigured1},~\ref{fig:subfiguree1}, and ~\ref{fig:subfiguref1}, respectively, the NMI, CR, and F1 histograms for the different algorithms. Clearly, geometric-AP significantly outperforms all other algorithms including non-exemplar-based methods on the three evaluation metrics. This result suggests that the network information comprises a substantial knowledge about the structure of the ground-truth categories. Additionally, the edge distribution captured by the neighborhood function $N_{\mathcal{G}}^{\tau}$ in (\ref{eq31}) seems to be highly compatible with the identified exemplars as it increases the affinity of cluster members to the most appropriate representatives.

In Figs.~\ref{fig:subfigureg1},~\ref{fig:subfigureh1}, and ~\ref{fig:subfigurei1} we plot NMI, CR, and F1 as functions of $K$, the number of identified clusters. Detecting variable $K$ is possible by calibrating the self-preferences $\left [ s_{ii}\right ]_{N\times N}$ over a range of values in AP and geometric-AP. Remaining clustering methods generate the desired number of clusters by specifying the preferred number of cluster components in initialization. Whenever kmeans is used, the simulations are repeated 1000 times with random restarts and the best performance is plotted. Henceforward, the aforementioned setup is used when reporting the
NMI, CR, and F1 values as functions of number of clusters $K$.
The figures show that geometric-AP consistently outperforms other methods by a significant amount w.r.t NMI, CR, and F1 when the number of identified clusters spans the number of ground-truth categories.
\subsubsection{Citeseer Dataset}
\label{subsec:Citeseer}

\input{figures/citeseer}

The \emph{\textbf{citeseer}} dataset~\citep{Giles1998} contains 3312 labeled publications spread over six classes with 4732 links between them. The links between documents indicate a citation relationship and each paper is represented by a binary word vector of dimension 3703 after stemming and removing stop words. Similar to \emph{\textbf{cora}}, words with document frequency less than 10 are removed. On average, each document in \emph{\textbf{citeseer}} has 32 words~\citep{Yang2015} and we harness the provided network information as an undirected graph.

The metric selection procedure discussed in section~\ref{similarity_metric} yields to the negative cosine distance as the best similarity measure on \emph{\textbf{citeseer}} w.r.t our setup.
Plots in Figs.~\ref{fig:subfigurea2},~\ref{fig:subfigureb2}, and~\ref{fig:subfigurec2} illustrate the clustering results of geometric-AP with different neighborhood functions $N_{\mathcal{G}}^{\tau}$ as a function of $\tau$ when the ground-truth classes are being predicted. This hyper-parameter tuning step is crucial and should be performed for any studied dataset. In \emph{\textbf{citeseer}}, parameter tuning serves to identify the best neighborhood function $N_{\mathcal{G}}^{\tau}$ defined in (\ref{eq9}) by searching for the finest topological distance ``$\mbox{distance}_{\mathcal{G}}$" and the most adequate threshold value $\tau$ that engender the highest clustering result w.r.t NMI, CR, and F1. Deterministically, the optimal threshold value corresponds to the one that gives the largest NMI. For each topological distance, geometric-AP is run with $\tau$ ranging from 1 to 7 for the shortest path distance and from 0.5 to 0.9 for the Jaccard and cosine metrics. The reference task for tuning the model parameters is the prediction of the true class labels, i.e., 6 in the case of \emph{\textbf{citeseer}}. Fig.~\ref{fig:subfigurea2} shows that the $\mathtt{shortest\_path}$ distance should be considered with a neighborhood threshold $\tau=5$.
Thereafter, the neighborhood function $N_{\mathcal{G}}^{\tau}$ is set as follows:
\begin{equation}\label{eq32}
N_{\mathcal{G}}^{\tau}\left ( i  \right ) = \left \{ x: \mathtt{shortest\_path}\left( x, i\right) \leq 5\right \}.
\end{equation}
It should be noticed that, on the \emph{\textbf{citeseer}} dataset the optimal neighborhood function encompasses larger diameter ($\tau=5$) as compared to \emph{\textbf{cora}} dataset where the best $\tau$ found to be 3. The main reason is that the network in \emph{\textbf{cora}} dataset is more dense (network density of 14.812*$10^{-4}$) and has higher average node degree (4) while the \emph{\textbf{citeseer}} dataset has a network density of 85.179 *$10^{-5}$ and an average node degree equal to 2~\citep{Rossi2015}. As expected, increasing sparsity in the network usually compels geometric-AP to delve deeper into the network to locate the best clustering structure.

We plot in Figs.~\ref{fig:subfigured2},~\ref{fig:subfiguree2}, and ~\ref{fig:subfiguref2}, respectively, the NMI, CR, and F1 histograms for the different algorithms. Like on \emph{\textbf{cora}}, geometric-AP significantly outperforms its counterparts AP and kmedoids and still behaves comparably to other top performing methods.

In Figs.~\ref{fig:subfigureg2},~\ref{fig:subfigureh2}, and ~\ref{fig:subfigurei2} we plot NMI, CR, and F1 as functions of $K$.
The simulations confirm that geometric-AP consistently outperforms exemplar-based methods and generates comparable results when confronted with other methods. A comparative analysis of the results obtained on \emph{\textbf{cora}} and \emph{\textbf{citeseer}} stipulates that the increase in clustering performance harvested from the network information diminishes as the true categories become more spread over the network. This observation is consistent with findings previously summarized in~\citep{Schaeffer2007}.
Additionally, the $\mathtt{shortest\_path}$ distance has been identified on both datasets as the best distance measure. This is explained by the high flexibility offered by this metric to the potential exemplars as they are allowed to declare availability to non-neighbor nodes and data points at more than 2 hops in the network. In contrast, Jaccard and cosine distances quantify only the shared neighborhood between two nodes in the network and as such the exemplars are unable to communicate their availability to nodes far more than 2 hops in the graph. Figs.~\ref{fig:subfigurea1} and~\ref{fig:subfigurea2} demonstrate that declaring availability to fewer data points than the optimal number engenders scarcity in the communication required to identify good clusters. Also, advertising the availability of exemplars in a broad manner introduces an additional noise that deteriorates the identification of good exemplars and misleads the search for valid label configurations.
\section{Clustering of Social Networks}
\label{sec:Clustering_of_social_networks}
\subsection{Zachary's Karate Club}
The Zachary's karate club~\citep{Zachary1977} is a social network of a university karate club that was monitored for two years. The network contains 34 members and 78 links between them. The links document the interaction between members outside the club. During the study, a conflict between the instructor and the administrator arose, which resulted in the split of the club into two sets.
Each club member is represented by a binary vector of 34 dimensions indicating the interaction outside the club with other members. In the remaining of this section we evaluate the ability of geometric-AP and other algorithms in retrieving the correct split of the club members. Also, we consider an additional 4-class partition obtained by modularity-based clustering~\citep{Brandes2008} and we assess the clustering performance w.r.t these pseudo ground-truth labels.
Indeed, modularity has been first introduced in~\citep{Newman2004} as a quality measure for graph clustering. Thenceforth, it has attracted considerable research attention and becomes widely accepted as a quality index for graph clustering. By considering the additional modularity-based classes we aim at analyzing the power of geometric-AP in sensing the modularity within graph-structured datasets.
We note that the metric selection procedure presented in section~\ref{similarity_metric} for both ground-truth class labels (club split and modularity-based classes) identifies the optimal similarity measure as the negative cosine distance.

\subsubsection{Club Split Clusters}
Similar to \emph{\textbf{cora}} and \emph{\textbf{citeseer}} datasets, we initially start by searching for the optimal topological distance and its corresponding threshold. Figs.~\ref{fig:subfigurea3}-\ref{fig:subfigurec3} shows that the $\mathtt{Jaccard}$ distance with a threshold value of 0.5 performs the best in retrieving the club split. The neighborhood function is then given by:
\begin{equation}\label{eq33}
N_{\mathcal{G}}^{\tau}\left ( i  \right ) = \left \{ x: \mathtt{Jaccard}\left( x, i\right) \leq 0.5\right \}.
\end{equation}
\input{figures/karate_split}
As illustrated in Figs.~\ref{fig:subfigured3}-\ref{fig:subfiguref3}, geometric-AP consistently outperforms its counterparts of exemplar-based methods and performs comparably to other top performing algorithms such as kmeans and Gaussian mixture models. While the standard AP failed to compete in recovering the actual split and misassigned 7 members, geometric-AP has successfully identified the two sets with only one misclassified member.
Obviously, our geometric model boosts the clustering accuracy with respect to all evaluation metrics with the largest enhancement being recorded for NMI, which has increased by 52\% as compared to the standard AP.
\subsubsection{Modularity-Based Clustering}
Likewise, for the three topological distances being tested, the NMI, CR, and F1 scores plotted in Figs.~\ref{fig:subfigurea4}-\ref{fig:subfigurec4} show that the $\mathtt{Jaccard}$ distance detects the best neighborhood function $N_{\mathcal{G}}^{\tau}$ for a threshold value of 0.8. As such, $N_{\mathcal{G}}^{\tau}$ is defined as follows:
\begin{equation}\label{eq34}
N_{\mathcal{G}}^{\tau}\left ( i  \right ) = \left \{ x: \mathtt{Jaccard}\left( x, i\right) \leq 0.8\right \}.
\end{equation}
As expected, geometric-AP achieves the best NMI value of 76.76\% outperforming all other studied methods while the closest result has been attained by ``Spectral-g'' (76.01\%). Meanwhile, for CR and F1, geometric-AP remains comparable with the state-of-the-art graph clustering method ``Spectral-g''.

Obviously, the obtained results for recovering the various ground-truth class labels on the Zachary's karate club network have proven the consistent performance of our proposed method albeit the network information is redundant and explicitly extracted from the node features.

\input{figures/karate_modularity}

This observation confirms that geometric-AP effectively leverages the topological local neighborhood to better unveil irregularly shaped clusters associated with the analyzed data.

\subsection{Clustering with Node Embeddings}
Like the majority of machine learning algorithms, the performance of geometric-AP greatly depends on data representation. For instance, different node embeddings may entangle or expose more or less the structure of the clusters present in the data~\citep{Bengio2013}. With the fact that exemplar-based clustering methods are more successful with regularly shaped structures, representation learning becomes a key factor in achieving satisfactory clustering results.
To provide a proof of concept that geometric-AP can seamlessly be integrated with state-of-the-art representation learning methods while maintaining its efficiency we replace the node features used in the Zachary's Karate Club network by two different node embeddings obtained by two embedding methods widely used in the literature.
We first consider the Fruchterman-Reingold force-directed algorithm (FRFD) that mimics forces in natural systems to embed undirected graphs in two dimensional spaces~\citep{Fruchterman1991}. Similarly, we use t-SNE (t-distributed Stochastic Neighbor Embedding) method~\citep{Maaten2008} that is well suited for visualization of high-dimensional datasets to project the network into the plane. Both embedding methods are observed as an aggregation of dimensionality reduction and representation learning techniques.
As FRFD and t-SNE are randomly initialized, we generate 1000 sets of node embeddings per method and we analyze the average clustering performance of the different clustering algorithms being tested. For each evaluation metric we also report the standard deviation.
For all sets of node embeddings we consider the negative Euclidean distance as the affinity measure for both geometric-AP and AP. Hereafter, we tune the neighborhood function $N_{\mathcal{G}}^{\tau}$ for an arbitrary selected random seed and we retain the best configuration throughout all other repetitions. In our simulations we have used a random seed of value 13579.
Notice that for each set of node embeddings we produce the desired number of clusters by automatically adjusting the shared preference $s_{ii}$ via dichotomic search. Node embeddings for which geometric-AP or AP diverges are regenerated.

\subsubsection{Fruchterman-Reingold Force-Directed Embeddings}
\label{Fruchterman_Reingold}
We produce 1000 sets of FRFD node embeddings and we run geometric-AP and other benchmark algorithms presented in section~\ref{Methods_and_Settings} for two clustering tasks. At first stage, we use the actual classes resulted from the club split. Then, we consider the classes obtained by modularity-based clustering. We report histograms of average NMI, CR, and F1 score values along with standard deviation bars for the tested algorithms. For methods that do not depend on node features such as Spectral-g or for those that show negligible standard deviations ($<10^{-5}$) we omit the error bars.

\input{figures/karate_FRFD}

Figs.~\ref{fig:subfigurea5}-\ref{fig:subfigurec5} show that a significant improvement has been achieved by geometric-AP in modeling the correct club split. Additionally, geometric-AP manifested the lowest variability on the three evaluation metrics. The reason is that the network information makes the algorithm less sensitive to the random noise associated with node features.
Likewise, Figs.~\ref{fig:subfigured5}-\ref{fig:subfiguref5} illustrate the consistent performance of geometric-AP in retrieving the correct modularity-based classes. Overall, geometric-AP remained the most robust method against the randomness associated with the used embeddings.

\subsubsection{t-SNE Embeddings}
Similarly, we employ the t-SNE algorithm to generate 1000 sets of node embeddings and we use the same setting discussed in section~\ref{Fruchterman_Reingold} to run the simulations. We plot in Figs.~\ref{fig:subfigurea6}-\ref{fig:subfigurec6} and \ref{fig:subfigured6}-\ref{fig:subfiguref6} the clustering results when using the club split classes and the modularity-based labels, respectively.
Obviously, geometric-AP outperforms exemplar-based methods by significant margins in all evaluation metrics. Also, it surpasses all feature-dependent methods including the state-of-the-art kmeans. On the other hand, geometric-AP performs either comparably or proximally to the popular graph clustering method Spectral-g even though geometric-AP is not principally designed for graph clustering.

\input{figures/karate_tsne}

By considering the clustering performance of geometric-AP using node embeddings, the comparative results establish the steady efficiency of geometric-AP in leveraging the available network information to boost the clustering accuracy.
Additionally, geometric-AP demonstrates high compatibility with representation-learning-based methods and shows promising potentials if combined with data-driven methods.

\subsubsection{Visualization Example}
\begin{figure*}[ht]
\begin{subfigure}{.33\textwidth}
  \centering
  \includegraphics[width=1\linewidth]{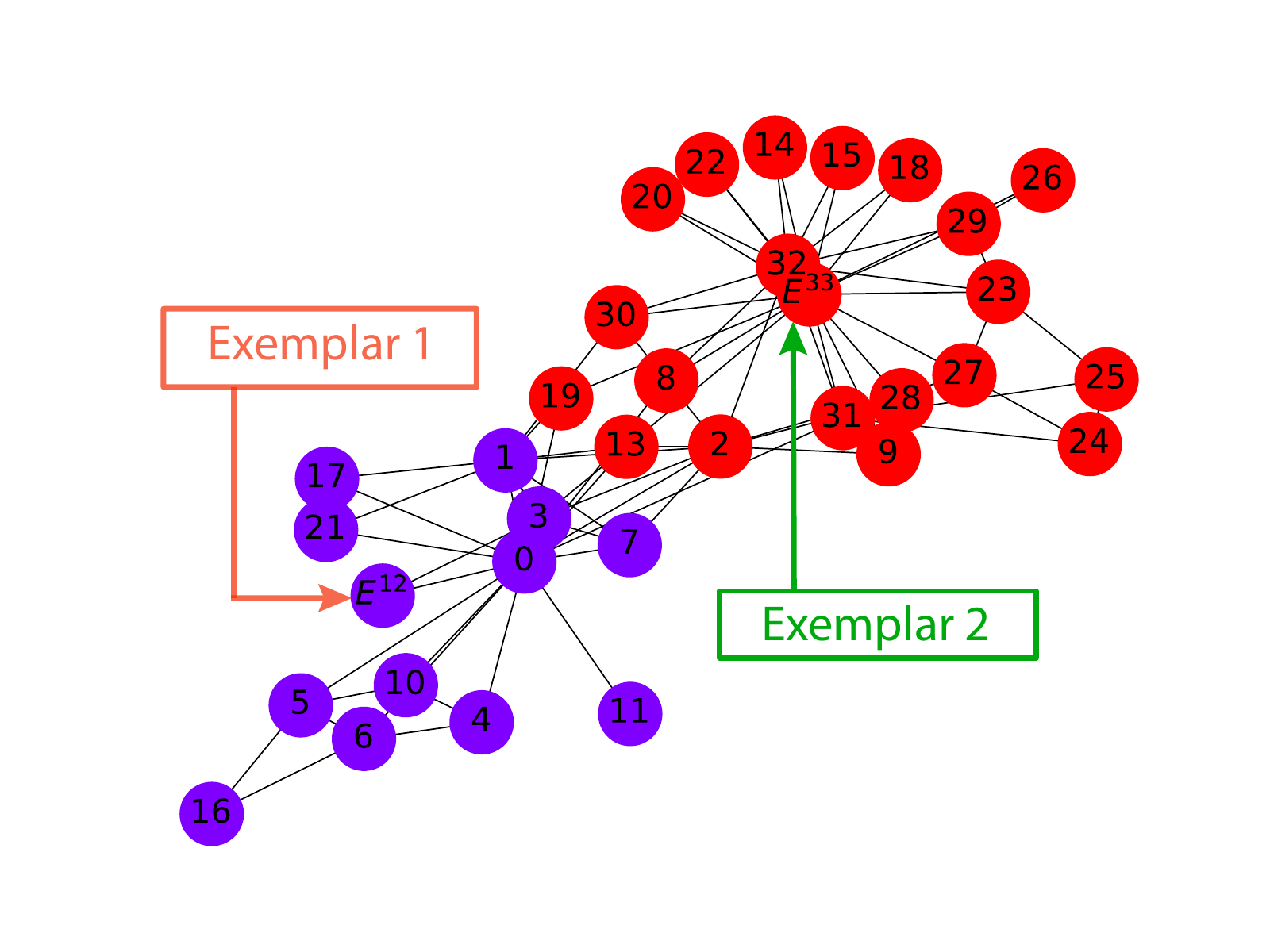}  
  \caption{AP (NMI: 57.78\% , CR: 88.24\%, F1: 88.08\%).}
  \label{fig:karete_split_ap}
\end{subfigure}
\begin{subfigure}{.33\textwidth}
  \centering
  \includegraphics[width=1\linewidth]{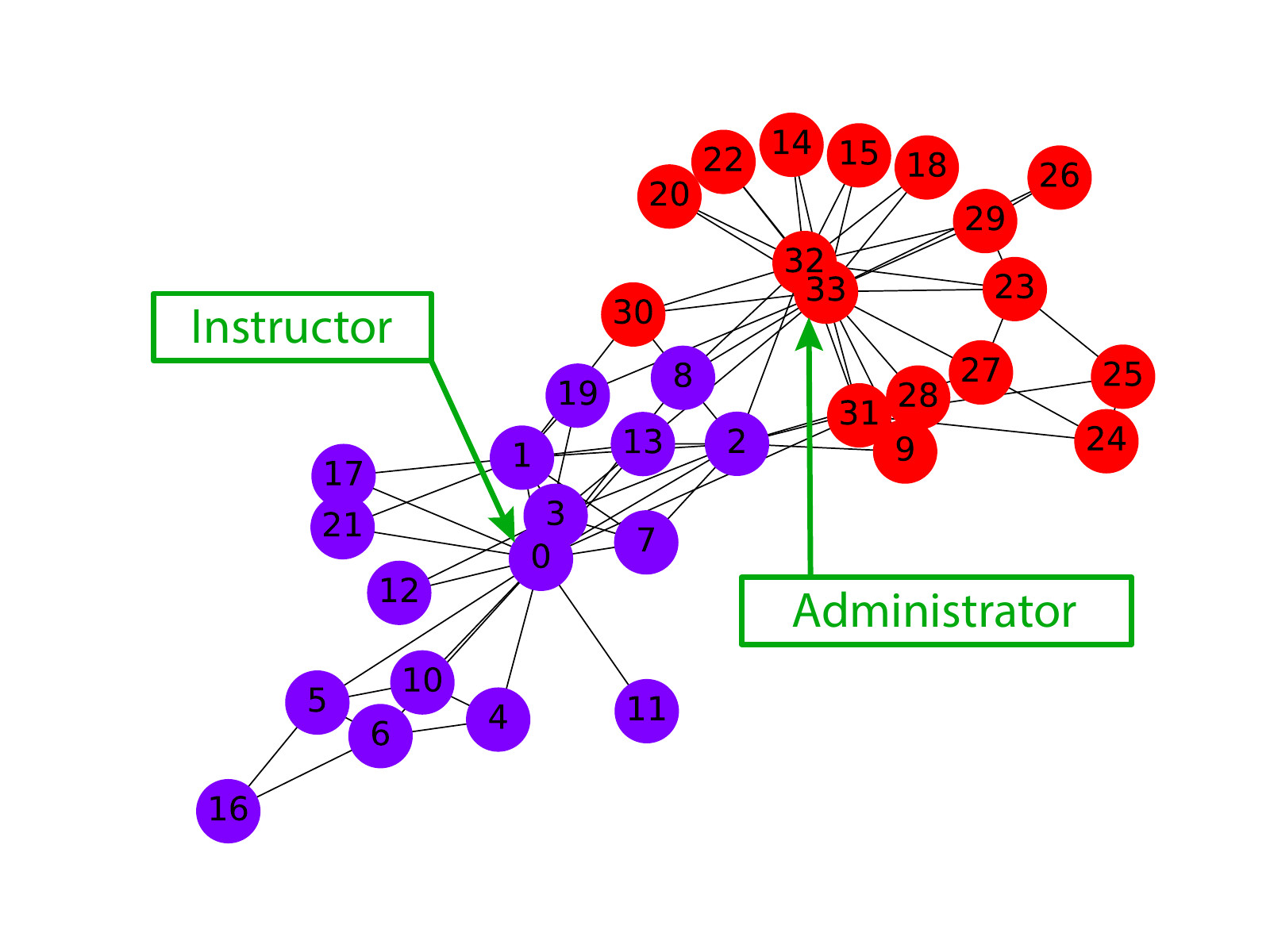}  
  \caption{Ground-truth labels}
  \label{fig:karete_split}
\end{subfigure}
\begin{subfigure}{.33\textwidth}
  \centering
  \includegraphics[width=1\linewidth]{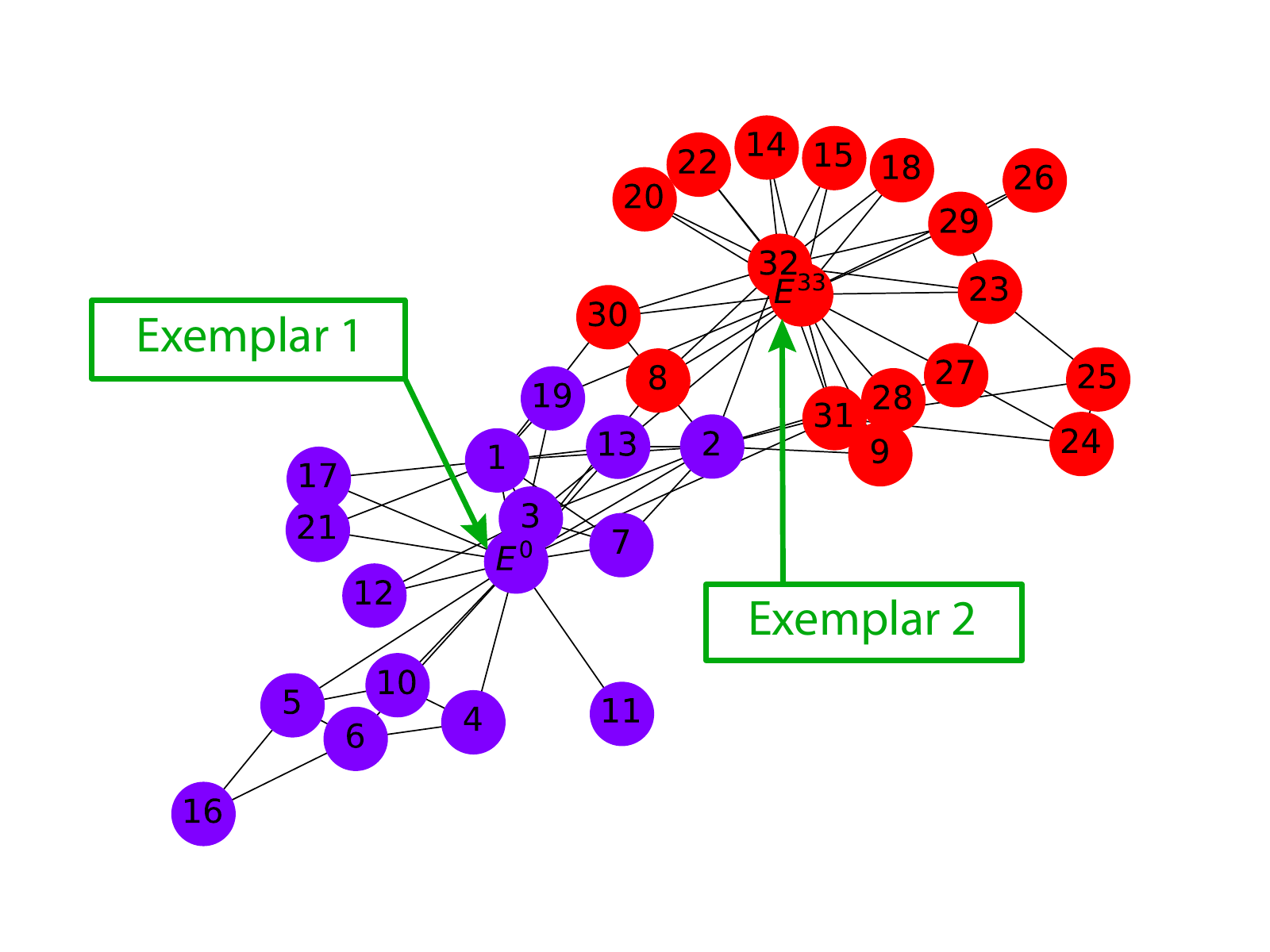}  
  \caption{Geometric-AP (NMI: 83.72\% , CR: 97.06\%, F1: 97.06\%).}
  \label{fig:karete_split_gap}
\end{subfigure}
\caption{Network visualization and clustering results of the Zachary's Karate Club with node features generated by FRFD (random seed = 13579). For visualization purposes, the node embeddings are used as plane coordinates of network nodes. Cluster exemplars are labeled by $E^{ID}$ and same cluster members are color-coded. After hyperparameter tuning, the optimal neighborhood function is found to be $N_{\mathcal{G}}^{\tau}\left ( i  \right ) = \left \{ x: \mathtt{cosine}\left( x, i\right) \leq 0.8\right \}$. (\textbf{a}) For AP, box labels highlight the identified exemplars. The green color is used when an identified exemplar matches one of the club leaders (instructor or administrator). Otherwise, the exemplar is emphasized by red box. (\textbf{b}) Color-coded network nodes based on actual class labels. Club leaders are highlighted in the network. Node with id 0 corresponds to the instructor and node with id 33 corresponds to the administrator. (\textbf{c}) geometric-AP successfully identifies both club leaders as exemplars.}
\label{fig:zachary}
\end{figure*}

\begin{figure*}[ht]
\begin{subfigure}{.33\textwidth}
  \centering
  \includegraphics[width=1\linewidth]{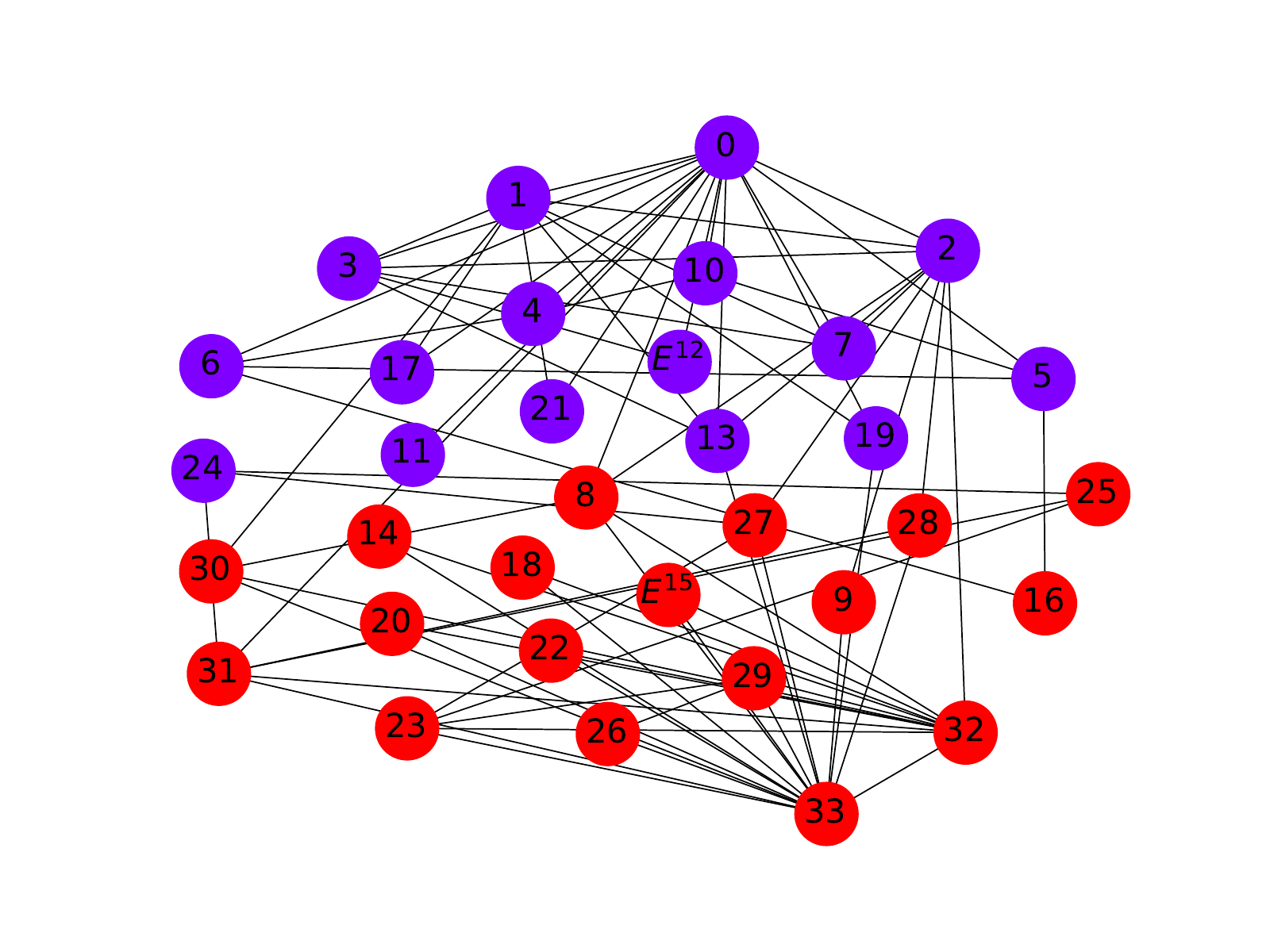}  
  \caption{AP (NMI: 57.56\% , CR: 91.18\%, F1: 91.17\%).}
  \label{fig:karete_split_ap_tsne}
\end{subfigure}
\begin{subfigure}{.33\textwidth}
  \centering
  \includegraphics[width=1\linewidth]{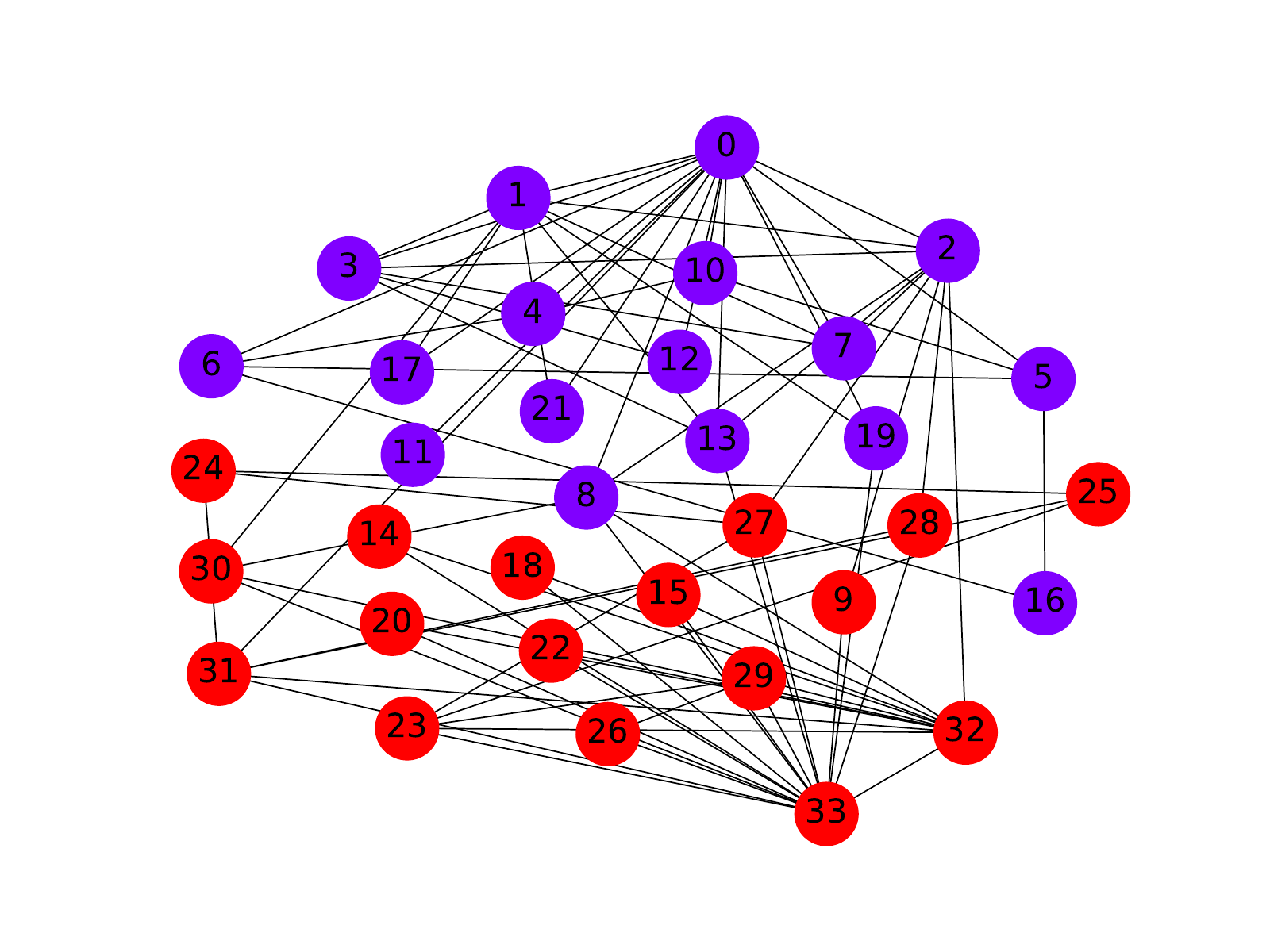}  
  \caption{Ground-truth labels}
  \label{fig:karete_split_tsne}
\end{subfigure}
\begin{subfigure}{.33\textwidth}
  \centering
  \includegraphics[width=1\linewidth]{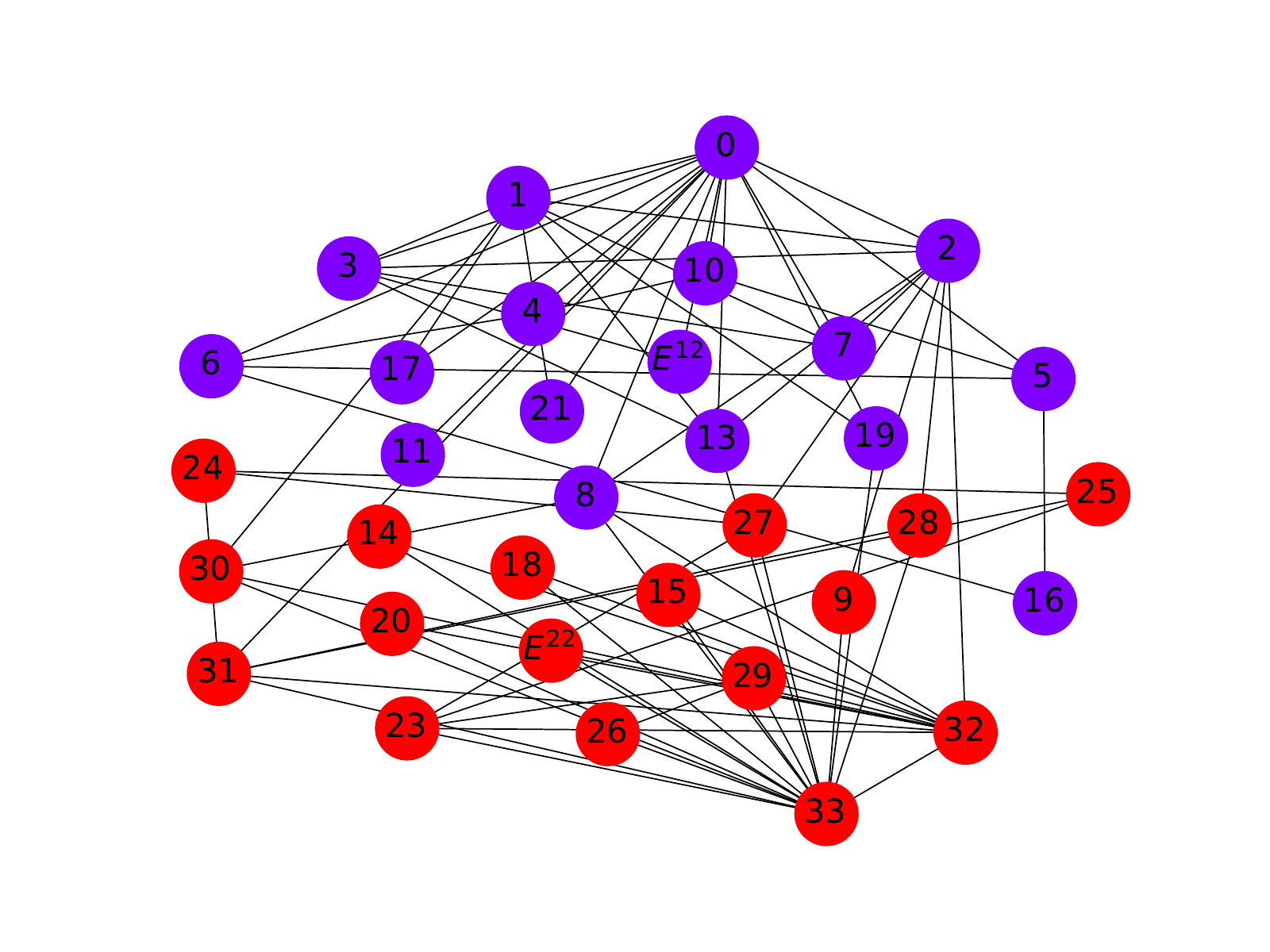}  
  \caption{Geometric-AP (NMI: 100\% , CR: 100\%, F1: 100\%).}
  \label{fig:karete_split_gap_tsne}
\end{subfigure}
\caption{Network visualization and clustering results of the Zachary's Karate Club with node features generated by t-SNE (random seed = 13579). For visualization purposes, the node embeddings are used as plane coordinates of network nodes. Cluster exemplars are labeled by $E^{ID}$ and same cluster members are color-coded. After hyperparameter tuning, the optimal neighborhood function is found to be $N_{\mathcal{G}}^{\tau}\left ( i  \right ) = \left \{ x: \mathtt{cosine}\left( x, i\right) \leq 0.7\right \}$.}
\label{fig:zachary_tsne}
\end{figure*}

As an illustration, we plot in Fig.~\ref{fig:zachary} and Fig.~\ref{fig:zachary_tsne} a two-dimensional visualization of the Zachary's karate club. For a fixed random seed (13579), we generate two sets of node embeddings using the aforementioned embedding methods (FRFD and t-SNE) and we color code the resulting clusters obtained by geometric-AP and the standard AP when predicting the club split classes.

With FRFD embeddings, geometric-AP misclassifies only one club member, that is node id 8, leading to a 97.06 percent CR. On the left side of Fig.~\ref{fig:zachary}, AP misassigns 4 club members resulting in an 88.24 percent CR. More interestingly, geometric-AP identifies both the instructor (node 0) and the administrator (node 33) as exemplars as illustrated in Fig.~\ref{fig:karete_split_gap}. However, AP detects only one true cluster leader (the administrator) among the two identified exemplars (Fig.~\ref{fig:karete_split_ap}).
Note that the club member that has been misclassified by geometric-AP in this case is exactly the same member that has been misclassified by the Ford–Fulkerson algorithm used in~\citep{Zachary1977}.

Using t-SNE embeddings, geometric-AP is able to successfully assign all club members to their actual groups. However, AP still misclassifies 3 members leading to a 91.17 percent CR. Visually, the identified exemplar by geometric-AP (node 22) within the administrator group occupies a more central location w.r.t other group members as compared to the one identified by AP (node 15).

Overall, this simple yet intuitive example illustrates how the new proposed geometric-AP model can combine the available network information with the feature representations of data points to conduct an efficient clustering.

\section{Geometric-AP with Random Networks}
\label{sec:Geometric_AP_with_random_networks}
As an ablation study, we aim in this section at evaluating the impact and statistical significance of network information on geometric-AP.
For this reason, we use \emph{\textbf{cora}} and \emph{\textbf{citeseer}} datasets to generate 100 random networks per dataset where we maintain identical structures as the original networks and we randomly relabel all nodes in the networks. This network permutation is expected to alter most of the information about the local neighborhood within each network.
For every generated network we repeatedly run geometric-AP to get the same sequence of cluster numbers obtained in section~\ref{subsec:Cora} and section~\ref{subsec:Citeseer} respectively. Likewise, we automatically adjust the shared preference $s_{ii}$ to produce the desired number of clusters and we report the average NMI, CR, and F1 score values.

\subsection{Cora Network}
We reproduce the simulations performed in section~\ref{subsec:Cora} and we designate geometric-AP that uses random networks by geometric-AP-RND. As expected, Figs.~\ref{fig:subfigurea7}-\ref{fig:subfigurec7} show that randomly permuting the node labels in the simulated networks severely misleads geometric-AP and makes it perform much worse than the original AP algorithm. For instance, NMI of the clustering results by geometric-AP-RND does not exceed 2 percent in all the experiments, which proves that the clustering with random network information is nearly random too. Similar trends are observed with CR and F1 as well.

\input{figures/cora_random_network}

These results demonstrate that the enhancement in clustering performance associated with geometric-AP does not emerge at random but instead it relies on meaningful network information that accurately defines the local neighborhood within the actual classes.

 \subsection{Citeseer Network}
 Similarly, Figs.~\ref{fig:subfigurea8}-\ref{fig:subfigurec8} show that randomizing the \emph{\textbf{citeseer}} network information acutely degrades the clustering performance of geometric-AP. Also, over 100 simulations with random networks, NMI of the clustering results by geometric-AP-RND does not surpass 3 percent in the most optimistic run. For example, Fig.~\ref{fig:subfigurea8} shows that geometric-AP-RND has very poor performance with small variability, which confirms that the previously observed improvement associated with geometric-AP is statistically significant.
 
\input{figures/citeseer_random_network}

 Ultimately, the effectiveness of geometric-AP has been established through an extensive validation on different real-world datasets that span different aspects of the proposed algorithm ranging from clustering accuracy to statistical significance. When assisted by relevant network information, geometric-AP has been able to significantly enhance the clustering accuracy and overcome many shortcomings associated with the original AP method.

\section{Conclusion}
\label{sec:Conclusion}
In this paper, we proposed a novel geometric clustering scheme, geometric-AP, by extending the original feature-based AP to effectively take advantage of network relations between the data points, often available in various scientific datasets.
geometric-AP locks its focus on the local network neighborhood during the message updates and makes potential exemplars only available within a predefined topological sphere in the network. The underlying objective is to maximize the similarity between the data points and their respective exemplars based on the node features while respecting the network topology to ensure the proximity of each data point from its  exemplar in the network. Using max-sum belief propagation over a factor graph, the new model has been optimized under the given network constraints at the level of function nodes. With an adjusted cluster assignment policy, the hybrid model further smooths the node labels throughout the network via majority voting. By initially considering all data points as potential exemplars, geometric-AP generates clusters insensitive to initialization.

Extensive validation based on two benchmark citation networks and one social network has clearly demonstrated the effectiveness of the proposed method and has confirmed the statistical significance of the obtained results. It has been shown that geometric-AP results in higher accuracy and robustness than the original AP in clustering the data points by leveraging relevant network knowledge. Furthermore, comparative performance assessment against other state-of-the-art methods have shown that the geometric-AP consistently yields favorable clustering results.

\appendix
\renewcommand{\theequation}{\thesection.\arabic{equation}}
\setcounter{equation}{0}
\section{Derivation of the Geometric-AP Message Updates}\label{A}

Messages sent from function nodes to variable nodes are derived by summing received messages and then maximizing over all variables other than those to which the message is being sent. The message sent from function $\gamma _{k}$ to variable $c_{i}$ is given by:
\begin{align}\label{eq35}
\alpha_{i\leftarrow k}\left ( c_{i} \right )=\max_{\left ( c_{1},c_{2},...,c_{i-1},c_{i+1},...,c_{N} \right )}
    &\left [ \gamma_{k}\left (  c_{1},c_{2},...,c_{i-1},\mathbf{c_{i}},c_{i+1},...,c_{N}\right ) +\sum_{i':i'\neq i}\rho_{i'\rightarrow k}\left ( c_{i'} \right )\right ],
\end{align}
and represents the best configuration that satisfies $\gamma _{k}$ given $c_{i}$.
Depending on the values taken by $i$, $k$, $c_{i}$, and $N_{\mathcal{G}}^{\tau}\left ( i  \right )$, all possible configurations from which the message updates emerge are shown in (\ref{eq36}).

\begin{align}\label{eq36}
\alpha_{i\leftarrow k}\left ( c_{i} \right )=\begin{cases}
\sum_{i':i'\neq i}\operatorname*{max}_{j'}\rho_{i'\rightarrow k}\left ( j' \right ), & \mbox{if}~i = k~\&~c_{i} = k~\&~k \in N_{\mathcal{G}}^{\tau}\left ( i  \right ),
\\ 
 & \\
\sum_{i':i'\neq i}\operatorname*{max}_{j':j' \neq k}\rho_{i'\rightarrow k}\left ( j' \right ), & \mbox{if}~i = k~\&~c_{i} \neq  k~\&~k \in N_{\mathcal{G}}^{\tau}\left ( i  \right ),
\\ 
& \\
\rho_{k\rightarrow k}\left ( k \right )+\sum_{i':i'\notin \left ( i,k \right )}\operatorname*{max}_{j'}\rho_{i'\rightarrow k}\left ( j' \right ), & \mbox{if}~i \neq k~\&~c_{i} =  k~\&~k \in N_{\mathcal{G}}^{\tau}\left ( i  \right ),
\\ 
& \\
\operatorname*{max}_{j':j'\neq k}\rho_{k\rightarrow k}\left ( j' \right )+\sum_{i':i'\notin \left ( i,k \right )}\operatorname*{max}_{j':j'\neq k}\rho_{i'\rightarrow k}\left ( j' \right ), & \mbox{if}~i \neq k~\&~c_{i} =  k~\&~k \notin N_{\mathcal{G}}^{\tau}\left ( i  \right ),
\\ 
& \\
\operatorname*{max}\left ( \operatorname*{max}_{j':j'\neq k}\rho_{k\rightarrow k}\left ( j' \right )+\sum_{i':i'\notin \left ( i,k \right )}\operatorname*{max}_{j':j'\neq k}\rho_{i'\rightarrow k}\left ( j' \right )~, \right.\\
\left.\rho_{k\rightarrow k}\left ( k \right )+\sum_{i':i'\notin \left ( i,k \right )}\operatorname*{max}_{j'}\rho_{i'\rightarrow k}\left ( j' \right )\right ), & \mbox{if}~i \neq k~\&~c_{i} \neq  k~\&~k \in N_{\mathcal{G}}^{\tau}\left ( i  \right ),
\\ 
& \\
\operatorname*{max}\left ( \operatorname*{max}_{j':j'\neq k}\rho_{k\rightarrow k}\left ( j' \right )+\sum_{i':i'\notin \left ( i,k \right )}\operatorname*{max}_{j':j'\neq k}\rho_{i'\rightarrow k}\left ( j' \right )~, \right. \\
\left.\rho_{k\rightarrow k}\left ( k \right )+\sum_{i':i'\notin \left ( i,k \right )}\operatorname*{max}_{j'}\rho_{i'\rightarrow k}\left ( j' \right )\right ), & \mbox{if}~i \neq k~\&~c_{i} \neq  k~\&~k \notin N_{\mathcal{G}}^{\tau}\left ( i  \right ).
\end{cases}
\end{align}
By following the formulation explained in~\cite{Frey2007}, we consider the exchanged messages as the sum of constant and variable components with respect to $c_{i}$.
\begin{align}\label{eq37}
  \begin{split}
     \rho_{i \rightarrow k}\left ( c_{i} \right ) & = \tilde{\rho}_{i \rightarrow k}\left ( c_{i} \right ) + \bar{\rho}_{i \rightarrow k}, \\
     \alpha_{i \leftarrow k}\left ( c_{i} \right )  & = \tilde{\alpha}_{i \leftarrow k}\left ( c_{i} \right ) + \bar{\alpha}_{i \leftarrow k}.
  \end{split}
\end{align}
Consequently, the messages can be written as shown by (\ref{eq38}) and (\ref{eq39}):
\begin{align}\label{eq38}
  \rho_{i\rightarrow k}\left ( c_{i} \right )=s\left ( i,c_{i} \right )+\sum_{k':k'\neq k}\tilde{\alpha}_{i\leftarrow k'}\left ( c_{i} \right )+\sum_{k':k'\neq k}\bar{\alpha}_{i\leftarrow k'}.
\end{align}

\begin{align}\label{eq39}
\alpha_{i\leftarrow k}\left ( c_{i} \right )=\begin{cases}
\sum_{i':i'\neq i}\operatorname*{max}_{j'}\tilde{\rho}_{i'\rightarrow k}\left ( j' \right )+\sum_{i':i'\neq i}\bar{\rho}_{i'\rightarrow k},  & \mbox{if}~i=k~\&~c_{i}=k~\&~k \in N_{\mathcal{G}}^{\tau}\left ( i  \right ),
\\ 
& \\
\sum_{i':i'\neq i}\operatorname*{max}_{j':j' \neq k}\tilde{\rho}_{i'\rightarrow k}\left ( j' \right )+\sum_{i':i'\neq i}\bar{\rho}_{i'\rightarrow k}\left ( j' \right ), &  \mbox{if}~i=k~\&~c_{i} \neq  k~\&~k \in N_{\mathcal{G}}^{\tau}\left ( i  \right ),
\\
& \\
\tilde{\rho}_{k\rightarrow k}\left ( k \right )+\sum_{i':i'\notin \left ( i,k \right )}\operatorname*{max}_{j'}\tilde{\rho}_{i'\rightarrow k}\left ( j' \right )+\sum_{i':i'\neq i}\bar{\rho}_{i'\rightarrow k}, &  \mbox{if}~i\neq k~\&~c_{i}=k~\&~k \in N_{\mathcal{G}}^{\tau}\left ( i  \right ),
\\
& \\
\operatorname*{max}_{j':j'\neq k}\tilde{\rho}_{k\rightarrow k}\left ( j' \right )+\sum_{i':i'\notin \left ( i,k \right )}\operatorname*{max}_{j':j'\neq k}\tilde{\rho}_{i'\rightarrow k}\left ( j' \right )\\+\sum_{i':i'\neq i}\bar{\rho}_{i'\rightarrow k}, &  \mbox{if}~i\neq k~\&~c_{i}=k~\&~k \notin N_{\mathcal{G}}^{\tau}\left ( i  \right ),
\\
& \\
\max \left ( \operatorname*{max}_{j':j'\neq k}\tilde{\rho}_{k\rightarrow k}\left ( j' \right )+\sum_{i':i'\notin \left ( i,k \right )}\operatorname*{max}_{j':j'\neq k}\tilde{\rho}_{i'\rightarrow k}\left ( j' \right )\right.\\
\left.+\sum_{i':i'\neq i}\bar{\rho}_{i'\rightarrow k}~,\right.
\\\left.
\tilde{\rho}_{k\rightarrow k}\left ( k \right )+\sum_{i':i'\notin \left ( i,k \right )}\operatorname*{max}_{j'}\tilde{\rho}_{i'\rightarrow k}\left ( j' \right )+\sum_{i':i'\neq i}\bar{\rho}_{i'\rightarrow k} \right), &  \mbox{if}~i\neq k~\&~c_{i}\neq k~\&~k \in N_{\mathcal{G}}^{\tau}\left ( i  \right ),
\\
& \\
\max\left(\operatorname*{max}_{j':j'\neq k}\tilde{\rho}_{k\rightarrow k}\left ( j' \right )+\sum_{i':i'\notin \left ( i,k \right )}\operatorname*{max}_{j':j'\neq k}\tilde{\rho}_{i'\rightarrow k}\left ( j' \right )\right.\\
\left.+\sum_{i':i'\neq i}\bar{\rho}_{i'\rightarrow k}~,\right.\\
\left.\tilde{\rho}_{k\rightarrow k}\left ( k \right )+\sum_{i':i'\notin \left ( i,k \right )}\operatorname*{max}_{j'}\tilde{\rho}_{i'\rightarrow k}\left ( j' \right )+\sum_{i':i'\neq i}\bar{\rho}_{i'\rightarrow k}\right), &  \mbox{if}~i\neq k~\&~c_{i}\neq k~\&~k \notin N_{\mathcal{G}}^{\tau}\left ( i  \right ).
\end{cases}
\end{align}

Similar to~\cite{Frey2007}, we set the maximum load of $\rho_{i\rightarrow k}\left ( j\right )$ on $\bar{\rho}_{i\rightarrow k}$ ($\bar{\rho}_{i\rightarrow k}= \operatorname*{max}_{j:j\neq k} \rho_{i\rightarrow k}\left ( j\right )$).
This choice will make:
\begin{align}\label{eq40}
  \begin{split}
     \operatorname*{max}_{j:j\neq k} \tilde{\rho}_{i\rightarrow k}\left ( j\right )& =0,\\
       \operatorname*{max}_{j} \tilde{\rho}_{i\rightarrow k}\left ( j\right )& = \max \left ( 0, \tilde{\rho}_{i\rightarrow k}\left ( k\right )\right ).
  \end{split}
\end{align}
Additionally, it should be noticed that the different expressions of the availability messages do not depend directly on $c_{i}$ but instead, only the choice of the right expression depends on $c_{i}$. We look at these expressions separately for $c_{i} = k$ and $c_{i} \neq k$ and we set $\bar{\alpha}_{i \leftarrow k} = \alpha_{i \leftarrow k}\left ( c_{i}:c_{i} \neq k\right )$.
This makes $\tilde{\alpha}_{i \leftarrow k} \left ( c_{i}\right ) = 0$ for all $c_{i} \neq k$.
Also, we get:
\begin{align}\label{eq41}
  \sum_{k': k' \neq k} \tilde{\alpha}_{i \leftarrow k'}\left ( c_{i} \right )=
  \begin{cases}
\tilde{\alpha}_{i \leftarrow c_{i}}\left ( c_{i} \right ), & \mbox{if}~c_{i} \neq k,
\\
 0, & \mbox{if} ~c_{i} = k.
\end{cases}
\end{align}
This leads to the simplifications given in (\ref{eq42}) and (\ref{eq43}):

\begin{align}\label{eq42}
\rho_{i \rightarrow k}\left ( c_{i} \right )=\begin{cases}
s\left ( i,k \right )+\sum_{k': k' \neq k}\bar{\alpha}_{i \leftarrow k'}, & \mbox{if}~c_{i}=k ,
\\
s\left ( i,c_{i} \right )+\tilde{\alpha}_{i \leftarrow c_{i}}\left ( c_{i} \right )+\sum_{k': k' \neq k}\bar{\alpha}_{i \leftarrow k'}, & \mbox{if} ~c_{i} \neq k.
\end{cases}
\end{align}
\begin{align}\label{eq43}
\alpha_{i \leftarrow k}\left ( c_{i} \right )=\begin{cases}
\sum_{i':i' \neq i}\max\left ( 0,\tilde{\rho}_{i' \rightarrow k}\left ( k \right ) \right )+ \sum_{i':i' \neq i}\bar{\rho}_{i' \rightarrow k}, & \mbox{if} \ c_{i}=k=i,
\\
& \\
\sum_{i':i' \neq i}\bar{\rho}_{i' \rightarrow k}, & \mbox{if} \ c_{i}\neq k=i,
\\
& \\
\tilde{\rho}_{k \rightarrow k}\left ( k \right )+\sum_{i':i' \notin \left \{ i,k \right \}}\max\left ( 0, \tilde{\rho}_{i' \rightarrow k}\left ( k \right )\right )+\sum_{i':i' \neq i}\bar{\rho}_{i' \rightarrow k}, & \mbox{if} \ c_{i}=k \neq i\ \& \ k \in N_{\mathcal{G}}^{\tau}\left ( i \right ),
\\
& \\
\max\left ( 0, \tilde{\rho}_{k \rightarrow k}\left ( k \right )+\sum_{i':i' \notin \left \{ i,k \right \}}\max\left ( 0, \tilde{\rho}_{i' \rightarrow k}\left ( k \right )\right )\right )+\sum_{i':i' \neq i}\bar{\rho}_{i' \rightarrow k}, & \mbox{if}\ c_{i} \neq k \neq i\ \&\ k \in N_{\mathcal{G}}^{r}\left ( i \right ),
\\
& \\
\sum_{i':i' \neq i}\bar{\rho}_{i' \rightarrow k}, & \mbox{if}\ c_{i}= k\neq i\ \& \ k \notin N_{\mathcal{G}}^{r}\left ( i \right ),
\\
& \\
\max\left ( 0, \tilde{\rho}_{k \rightarrow k}\left ( k \right )+\sum_{i':i' \notin \left \{ i,k \right \}}max\left ( 0, \tilde{\rho}_{i' \rightarrow k}\left ( k \right )\right )\right )+\sum_{i':i' \neq i}\bar{\rho}_{i' \rightarrow k}, & \mbox{if}\ c_{i} \neq k \neq i\ \& \ k \notin N_{\mathcal{G}}^{r}\left ( i \right ).
\end{cases}
\end{align}

Afterwards, we solve for:
\begin{align}\label{eq44}
  \begin{split}
     \tilde{\rho}_{i \rightarrow k}\left ( c_{i}=k \right ) & =\rho_{i \rightarrow k}\left ( c_{i}=k \right )-\bar{\rho}_{i \rightarrow k},\\
       \tilde{\alpha}_{i \rightarrow k}\left ( c_{i}=k \right ) & =\alpha_{i \rightarrow k}\left ( c_{i}=k \right )-\bar{\alpha}_{i \rightarrow k},
  \end{split}
\end{align}
where $\bar{\rho}$ and $\bar{\alpha}$ will cancel out and hence simpler equations are derived in (\ref{eq45}) and (\ref{eq46}):
\begin{align}\label{eq45}
    \begin{split}
        \tilde{\rho}_{i \rightarrow k}\left ( c_{i}=k \right ) &=\rho_{i \rightarrow k}\left ( c_{i}=k \right )-\bar{\rho}_{i \rightarrow k} \\
         &= \rho_{i \rightarrow k}\left ( k \right )-\operatorname*{max}_{j:j \neq k}\rho_{i \rightarrow k}\left ( j \right )\\
         &= s\left ( i,k \right )+\sum_{k':k' \neq k}\bar{\alpha}_{i \leftarrow k'}-\operatorname*{max}_{j:j \neq k}\left [ s\left ( i,j \right )+\tilde{\alpha}_{i \leftarrow j}\left ( j \right )+\sum_{k':k' \neq k}\bar{\alpha}_{i \leftarrow k'} \right ].
     \end{split}
\end{align}
\begin{align}\label{eq46}
\nonumber &\tilde{\alpha}_{i \rightarrow k}\left ( c_{i}=k \right ) =\alpha_{i \rightarrow k}\left ( c_{i}=k \right )-\bar{\alpha}_{i \rightarrow k} 
        = \alpha_{i \rightarrow k}\left ( k \right )-\alpha_{i \rightarrow k}\left ( j:j \neq k \right )\\
&= \begin{cases} 
\sum_{i':i' \neq k}\max\left ( 0,\tilde{\rho}_{i' \rightarrow k} \right )+\sum_{i':i' \neq i}\bar{\rho}_{i' \rightarrow k}-\sum_{i':i' \neq i}\bar{\rho}_{i' \rightarrow k}, & \mbox{if}\ k=i,
\\
& \\
 \tilde{\rho}_{k \rightarrow k}\left ( k \right )+\sum_{i':i' \notin \left \{ i,k \right \}}\max\left ( 0, \tilde{\rho}_{i' \rightarrow k}\left ( k \right )\right )+\sum_{i':i' \neq i}\bar{\rho}_{i' \rightarrow k}\\
-\max\left ( 0, \tilde{\rho}_{k \rightarrow k}\left ( k \right )+\sum_{i':i' \notin \left \{ i,k \right \}}\max\left ( 0, \tilde{\rho}_{i' \rightarrow k}\left ( k \right )\right )\right )-\sum_{i':i' \neq i}\bar{\rho}_{i' \rightarrow k}, & \mbox{if}\ k \neq i\ \&\ k \in N_{\mathcal{G}}^{\tau}\left ( i \right ),
 \\
 & \\
\sum_{i':i' \neq i}\bar{\rho}_{i' \rightarrow k}- \max\left ( 0, \tilde{\rho}_{k \rightarrow k}\left ( k \right )+\sum_{i':i' \notin \left \{ i,k \right \}}\max\left ( 0, \tilde{\rho}_{i' \rightarrow k}\left ( k \right )\right )\right )-\sum_{i':i' \neq i}\bar{\rho}_{i' \rightarrow k},& \mbox{if}\ k \neq i\ \&\ k \notin N_{\mathcal{G}}^{\tau}\left ( i \right ).
\end{cases}
\end{align}
The availability and responsibility messages are then defined as follows:
\begin{align}\label{eq47}
    \begin{split}
     r\left ( i,k \right ) &= \tilde{\rho}_{i \rightarrow k}\left ( k \right ) ,\\
     a\left ( i,k \right )  &= \tilde{\alpha}_{i \leftarrow k}\left ( k \right ).
  \end{split}
\end{align}
Finally, we get the expression in (\ref{eq48}) and the desired result given by (\ref{Eq19}).
\begin{align}\label{eq48}
    \begin{split}
         r\left ( i,k \right ) & =\tilde{\rho}_{i \rightarrow k}\left ( c_{i}=k \right ) \\
           & = s\left ( i,k \right )-\operatorname*{max}_{j:j \neq k}\left [ s\left ( i,j \right )+a\left ( i,j \right ) \right ].
    \end{split}
\end{align}


\end{document}

%% file: figures/cora.tex
\begin{figure*}[ht!]
\centering
\begin{minipage}{0.9775\textwidth}
\begin{subfigure}[b]{0.33\textwidth}
\begin{tikzpicture}
    \begin{axis}[ width=\linewidth,font=\footnotesize,
    xlabel= $\tau$, ylabel= Percentage (\%),
  xmin=1,xmax=5,ymin=0,ymax=50,ytick={0,10,20,30,40,50},
    legend style={nodes=right},legend pos= south west,legend style={nodes={scale=0.7, transform shape}},
      xmajorgrids,
    grid style={dotted},
    ymajorgrids,
   ]
\addplot[smooth,blue!80!black,mark=diamond*,mark options={solid},semithick] plot coordinates {
(1,43.28)(2,46.79)(3,47.27)(4,46.94)(5,42.51)
};
     \addplot[smooth,red!80!black,mark=otimes*,mark options={solid},semithick] plot coordinates {
(1,25.64)(2,31.97)(3,33.27)(4,32.89)(5,28.96)
};
     \addplot[smooth,green!60!black,mark=triangle*,mark options={solid},semithick] plot coordinates {
(1,22.52)(2,24.15)(3,27.21)(4,24.98)(5,19.86)
};
\addplot [red, thick, dashed] coordinates {(3,0) (3,47.27) } ;
\legend{CR,F1,NMI};
\end{axis}
\end{tikzpicture}
\caption{Geometric-AP (Shortest Path)}
\label{fig:subfigurea1}
\end{subfigure}
\begin{subfigure}[b]{0.33\textwidth}
\begin{tikzpicture}
    \begin{axis}[ width=\linewidth, font=\footnotesize,
    xlabel= $\tau$, ylabel= Percentage (\%),
  xmin=0.49,xmax=0.9,ymin=0,ymax=50,ytick={0,10,20,30,40,50},xtick={0.5,0.6,0.7,0.8,0.9},
    legend style={nodes=right},legend pos= south east,legend style={nodes={scale=0.7, transform shape}},
      xmajorgrids,
    grid style={dotted},
    ymajorgrids,
   ]
     \addplot[smooth,blue!80!black,mark=diamond*,mark options={solid},semithick] plot coordinates {
(0.5,44.43)(0.6,43.06)(0.7,43.76)(0.8,44.32)(0.9,44.58)
};
     \addplot[smooth,red!80!black,mark=otimes*,mark options={solid},semithick] plot coordinates {
(0.5,26.51)(0.6,25.86)(0.7,26.37)(0.8,26.75)(0.9,28.92)
};
     \addplot[smooth,green!60!black,mark=triangle*,mark options={solid},semithick] plot coordinates {
(0.5,22.92)(0.6,19.09)(0.7,20.16)(0.8,21.30)(0.9,21.59)
};
\addplot [red, thick, dashed] coordinates {(0.5,0) (0.5,44.43) } ;
\legend{CR,F1,NMI};
\end{axis}
\end{tikzpicture}
\caption{Geometric AP (Jaccard)}
\label{fig:subfigureb1}
\end{subfigure}
\begin{subfigure}[b]{0.33\textwidth}
\begin{tikzpicture}
    \begin{axis}[ width=\linewidth,font=\footnotesize,
    xlabel= $\tau$, ylabel= Percentage (\%),
  xmin=0.5,xmax=0.91,ymin=0,ymax=50,ytick={0,10,20,30,40,50},xtick={0.5,0.6,0.7,0.8,0.9},
    legend style={nodes=right},legend pos= south west,legend style={nodes={scale=0.7, transform shape}},
      xmajorgrids,
    grid style={dotted},
    ymajorgrids,
   ]
     \addplot[smooth,blue!80!black,mark=diamond*,mark options={solid},semithick] plot coordinates {
(0.5,44.17)(0.6,44.43)(0.7,44.43)(0.8,45.43)(0.9,46.46)
};
     \addplot[smooth,red!80!black,mark=otimes*,mark options={solid},semithick] plot coordinates {
(0.5,26.65)(0.6,26.83)(0.7,26.73)(0.8,29.68)(0.9,31.45)
};
     \addplot[smooth,green!60!black,mark=triangle*,mark options={solid},semithick] plot coordinates {
(0.5,20.82)(0.6,21.34)(0.7,21.32)(0.8,22.90)(0.9,24.40)
};
\addplot [red, thick, dashed] coordinates {(0.9,0) (0.9,46.46) } ;
\legend{CR,F1,NMI};
\end{axis}
\end{tikzpicture}
\caption{Geometric-AP (Cosine)}
\label{fig:subfigurec1}
\end{subfigure}
\par\bigskip 
\begin{subfigure}[b]{0.33\textwidth}
\begin{tikzpicture}
\begin{axis}[width=\linewidth,font=\footnotesize,
 ybar,bar shift=0,ymin=0,ymax=30,bar width=8pt,
ylabel={Percentage (\%)},
symbolic x coords={Geometric-AP,AP,kmedoids,zero1,kmeans,Spectral-g,HAC,GMM,DPGMM},
x tick label style={font=\scriptsize},
x tick label style={rotate=90,anchor=east,nodes near coords align={horizontal}},
xtick={Geometric-AP,AP,kmedoids,kmeans,Spectral-g,HAC,GMM,DPGMM},
ytick={0,5,10,15,20,25,30},
after end axis/.code={
\node(L1) at (axis cs:AP,1) {};
\node(L2) at (axis cs:HAC,1) {};
\node(XTL) at (xticklabel cs:0)  {};
\node[anchor=east](A1) at (axis cs:Geometric-AP,-14) {};
\node[anchor=west](B1) at (axis cs:kmedoids,-14) {};
\draw (A1|- XTL) -- (B1|- XTL);
\node[anchor=center,below] at (L1 |- XTL) {\scriptsize{Exemplar-Based}};
\node[anchor=east](A2) at (axis cs:kmeans,-14) {};
\node[anchor=west](B2) at (axis cs:DPGMM,-14) {};
\draw (A2|- XTL) -- (B2|- XTL);
\node[anchor=center,below] at (L2 |- XTL) {\scriptsize{Non Exemplar-Based}};
}
]
\addplot[red!80!black,fill=red!30!white,mark=none] coordinates {(Geometric-AP,27.21)};

\addplot[blue!80!black,fill=blue!30!white,mark=none] coordinates {(AP,13.25)};

\addplot[green!60!black,fill=green!30!white,mark=none] coordinates {(kmedoids,12.25	)};

\addplot[xtick=\empty, draw=none] coordinates {(zero1,0)};

\addplot[orange!80!black,fill=orange!30!white,mark=none] coordinates {(kmeans,19.56)};

\addplot[cyan!80!black,fill=cyan!30!white,mark=none] coordinates {(Spectral-g,7.63)};

\addplot[violet!80!black,fill=violet!30!white,mark=none] coordinates {(HAC,4.7)};

\addplot[yellow!80!black,fill=yellow!50!white,mark=none] coordinates {(GMM,15.35)};

\addplot[black,fill=lightgray,mark=none] coordinates {(DPGMM,17.82)};
\end{axis}

\end{tikzpicture}
\caption{NMI}
\label{fig:subfigured1}
\end{subfigure}
\begin{subfigure}[b]{0.33\textwidth}
\begin{tikzpicture}
\begin{axis}[width=\linewidth,font=\footnotesize,
 ybar,bar shift=0,ymin=0,ymax=50,bar width=8pt,
ylabel={Percentage (\%)},
symbolic x coords={Geometric-AP,AP,kmedoids,zero1,kmeans,Spectral-g,HAC,GMM,DPGMM},
x tick label style={font=\scriptsize},
x tick label style={rotate=90,anchor=east,nodes near coords align={horizontal}},
xtick={Geometric-AP,AP,kmedoids,kmeans,Spectral-g,HAC,GMM,DPGMM},
ytick={0,10,20,30,40,50},
after end axis/.code={
\node(L1) at (axis cs:AP,1) {};
\node(L2) at (axis cs:HAC,1) {};
\node(XTL) at (xticklabel cs:0)  {};
\node[anchor=east](A1) at (axis cs:Geometric-AP,-14) {};
\node[anchor=west](B1) at (axis cs:kmedoids,-14) {};
\draw (A1|- XTL) -- (B1|- XTL);
\node[anchor=center,below] at (L1 |- XTL) {\scriptsize{Exemplar-Based}};
\node[anchor=east](A2) at (axis cs:kmeans,-14) {};
\node[anchor=west](B2) at (axis cs:DPGMM,-14) {};
\draw (A2|- XTL) -- (B2|- XTL);
\node[anchor=center,below] at (L2 |- XTL) {\scriptsize{Non Exemplar-Based}};
}
]
\addplot[red!80!black,fill=red!30!white,mark=none] coordinates {(Geometric-AP,47.27	)};

\addplot[blue!80!black,fill=blue!30!white,mark=none] coordinates {(AP,38.37)};

\addplot[green!60!black,fill=green!30!white,mark=none] coordinates {(kmedoids,38.74	)};

\addplot[xtick=\empty, draw=none] coordinates {(zero1,0)};

\addplot[orange!80!black,fill=orange!30!white,mark=none] coordinates {(kmeans,42.1)};

\addplot[cyan!80!black,fill=cyan!30!white,mark=none] coordinates {(Spectral-g,33.42)};

\addplot[violet!80!black,fill=violet!30!white,mark=none] coordinates {(HAC,30.54)};

\addplot[yellow!80!black,fill=yellow!50!white,mark=none] coordinates {(GMM,38.26)};

\addplot[black,fill=lightgray,mark=none] coordinates {(DPGMM,41.18)};
\end{axis}
\end{tikzpicture}
\caption{CR}
\label{fig:subfiguree1}
\end{subfigure}
\begin{subfigure}[b]{0.33\textwidth}
\begin{tikzpicture}
\begin{axis}[width=\linewidth,font=\footnotesize,
 ybar,bar shift=0,ymin=0,ymax=40,bar width=8pt,
ylabel={Percentage (\%)},
symbolic x coords={Geometric-AP,AP,kmedoids,zero1,kmeans,Spectral-g,HAC,GMM,DPGMM},
x tick label style={font=\scriptsize},
x tick label style={rotate=90,anchor=east,nodes near coords align={horizontal}},
xtick={Geometric-AP,AP,kmedoids,kmeans,Spectral-g,HAC,GMM,DPGMM},
ytick={0,10,20,30,40},
after end axis/.code={
\node(L1) at (axis cs:AP,1) {};
\node(L2) at (axis cs:HAC,1) {};
\node(XTL) at (xticklabel cs:0)  {};
\node[anchor=east](A1) at (axis cs:Geometric-AP,-14) {};
\node[anchor=west](B1) at (axis cs:kmedoids,-14) {};
\draw (A1|- XTL) -- (B1|- XTL);
\node[anchor=center,below] at (L1 |- XTL) {\scriptsize{Exemplar-Based}};
\node[anchor=east](A2) at (axis cs:kmeans,-14) {};
\node[anchor=west](B2) at (axis cs:DPGMM,-14) {};
\draw (A2|- XTL) -- (B2|- XTL);
\node[anchor=center,below] at (L2 |- XTL) {\scriptsize{Non Exemplar-Based}};
}
]
\addplot[red!80!black,fill=red!30!white,mark=none] coordinates {(Geometric-AP,33.27)};

\addplot[blue!80!black,fill=blue!30!white,mark=none] coordinates {(AP,22.26)};

\addplot[green!60!black,fill=green!30!white,mark=none] coordinates {(kmedoids,23.06)};

\addplot[xtick=\empty, draw=none] coordinates {(zero1,0)};

\addplot[orange!80!black,fill=orange!30!white,mark=none] coordinates {(kmeans,31.59)};

\addplot[cyan!80!black,fill=cyan!30!white,mark=none] coordinates {(Spectral-g,11.84)};

\addplot[violet!80!black,fill=violet!30!white,mark=none] coordinates {(HAC,9.06)};

\addplot[yellow!80!black,fill=yellow!50!white,mark=none] coordinates {(GMM,28.37)};

\addplot[black,fill=lightgray,mark=none] coordinates {(DPGMM,29.44)};
\end{axis}

\end{tikzpicture}
\caption{F1}
\label{fig:subfiguref1}
\end{subfigure}
\par\bigskip 
\begin{subfigure}[b]{0.33\textwidth}
\begin{tikzpicture}
    \begin{axis}[width=\linewidth,font=\footnotesize,
    xlabel= $K$, ylabel= Percentage (\%),
  ymin=0,ymax=60,xmin=5,xmax=19,legend columns=2,
    legend style={nodes=right,/tikz/column 2/.style={
                column sep=2pt,},},legend pos= north west,legend style={nodes={scale=0.6, transform shape}},
      xmajorgrids,
    grid style={dotted},
    ymajorgrids,
    xtick={5,7,9,11,13,15,17,19},
    ytick={0,10,20,30,40,50,60},
    ]
\addplot[magenta!80!white,mark=pentagon*,mark options={solid},semithick] plot coordinates {
(5,26.38)(7,27.21)(9,28.03)(11,29.35)(13,29.63)(15,29.03)(17,29.75)(19,30.60)
};
     \addplot[blue!80!white,mark=triangle*,mark options={solid},semithick] plot coordinates {
(5,14.53)(7,13.25)(9,13.35)(11,13.74)(13,14.32)(15,14.95)(17,14.79)(19,15.23)
};
     \addplot[green!80!white,mark=triangle,mark options={solid},semithick] plot coordinates {
(5,7.86)(7,12.25)(9,14.58)(11,15.04)(13,14.12)(15,14.24)(17,14.32)(19,14.59)};

    \addplot[orange!80!white,mark=diamond,mark options={solid},semithick] plot coordinates {
(5,11.58)(7,19.56)(9,22.02)(11,21.59)(13,19.87)(15,23.23)(17,22.00)(19,24.53)};

     \addplot[cyan!80!white,mark=otimes*,mark options={solid},semithick] plot coordinates {
(5,1.58)(7,7.63)(9,3.08)(11,8.07)(13,8.87)(15,9.13)(17,10.76)(19,11.72)};

      \addplot[violet!80!white,mark=diamond*,mark options={solid},semithick] plot coordinates {
    (5,4.44)(7,4.70)(9,6.26)(11,7.71)(13,7.87)(15,8.01)(17,8.73)(19,9.19)};

       \addplot[yellow!90!black,mark=pentagon,mark options={solid},semithick] plot coordinates {
   (5,18.22)(7,15.35)(9,23.41)(11,20.67)(13,23.91)(15,19.79)(17,18.55)(19,22.39)};

      \addplot[gray,mark=otimes,mark options={solid},semithick] plot coordinates {
(5,16.26)(7,17.82)(9,22.22)(11,18.63)(13,17.51)(15,17.19)(17,21.70)(19,28.72)};

\legend{Geometric-AP,AP,kmedoids,kmeans,Spectral-g,HAC,GMM,DPGMM};
\end{axis}
\end{tikzpicture}
\caption{NMI}
\label{fig:subfigureg1}
\end{subfigure}
\begin{subfigure}[b]{0.33\textwidth}
\begin{tikzpicture}
    \begin{axis}[width=\linewidth,font=\footnotesize,
    xlabel= $K$, ylabel= Percentage (\%),
    ymin=30,ymax=80,xmin=5,xmax=19,legend columns=2,
    legend style={nodes=right,/tikz/column 2/.style={
                column sep=2pt,},},legend pos= north west,legend style={nodes={scale=0.6}},
      xmajorgrids,
    grid style={dotted},
    ymajorgrids,
    xtick={5,7,9,11,13,15,17,19},
    ytick={30,40,50,60,70,80},
   ]
\addplot[magenta!80!white,mark=pentagon*,mark options={solid},semithick] plot coordinates {
(5,47.79)(7,47.27)(9,47.72)(11,52.55)(13,52.77)(15,52.44)(17,55.06)(19,56.69)};

     \addplot[blue!80!white,mark=triangle*,mark options={solid},semithick] plot coordinates {
(5,39.30)(7,38.37)(9,38.34)(11,40.15)(13,41.55)(15,42.62)(17,42.73)(19,43.39)};

      \addplot[green!80!white,mark=triangle,mark options={solid},semithick] plot coordinates {
(5,31.17)(7,38.74)(9,39.78)(11,41.66)(13,41.40)(15,41.59)(17,41.92)(19,42.21)};

   \addplot[orange!80!white,mark=diamond,mark options={solid},semithick] plot coordinates {
(5,34.31)(7,42.10)(9,45.83)(11,46.64)(13,46.79)(15,48.23)(17,49.23)(19,52.22)};

     \addplot[cyan!80!white,mark=otimes*,mark options={solid},semithick] plot coordinates {
(5,30.43)(7,33.42)(9,30.73)(11,33.57)(13,33.83)(15,34.05)(17,34.98)(19,35.20)};

      \addplot[violet!80!white,mark=diamond*,mark options={solid},semithick] plot coordinates {
    (5,30.54)(7,30.54)(9,32.06)(11,34.46)(13,34.68)(15,35.49)(17,36.82)(19,37.63)};

     \addplot[yellow!90!black,mark=pentagon,mark options={solid},semithick] plot coordinates {
  (5,44.06)(7,38.26)(9,45.13)(11,46.05)(13,47.05)(15,41.62)(17,41.62)(19,45.76)};

     \addplot[gray,mark=otimes,mark options={solid},semithick] plot coordinates {
(5,38.41)(7,41.18)(9,46.46)(11,39.78)(13,40.59)(15,43.84)(17,48.42)(19,53.51)};

\legend{Geometric-AP,AP,kmedoids,kmeans,Spectral-g,HAC,GMM,DPGMM};

\end{axis}
\end{tikzpicture}
\caption{CR}
\label{fig:subfigureh1}
\end{subfigure}
\begin{subfigure}[b]{0.33\textwidth}
\begin{tikzpicture}
    \begin{axis}[width=\linewidth,font=\footnotesize,
    xlabel= $K$, ylabel= Percentage (\%),
    ymin=0,ymax=100,xmin=5,xmax=19,legend columns=2,
    legend style={nodes=right,/tikz/column 2/.style={
                column sep=2pt,},},legend pos= north west,legend style={nodes={scale=0.6, transform shape}},
      xmajorgrids,
    grid style={dotted},
    ymajorgrids,
    xtick={5,7,9,11,13,15,17,19},
   ]

\addplot[magenta!80!white,mark=pentagon*,mark options={solid},semithick] plot coordinates {
(5,33.67)(7,33.27)(9,40.89)(11,40.77)(13,40.87)(15,40.59)(17,47.11)(19,48.55)};

     \addplot[blue!80!white,mark=triangle*,mark options={solid},semithick] plot coordinates {
(5,24.05)(7,22.26)(9,22.15)(11,23.80)(13,27.58)(15,29.21)(17,30.39)(19,36.21)};
     \addplot[green!80!white,mark=triangle,mark options={solid},semithick] plot coordinates {
(5,12.40)(7,23.06)(9,24.98)(11,28.18)(13,29.16)(15,29.29)(17,29.76)(19,29.92)};
   \addplot[orange!80!white,mark=diamond,mark options={solid},semithick] plot coordinates {
(5,20.49)(7,31.59)(9,35.58)(11,38.17)(13,38.05)(15,38.68)(17,41.63)(19,47.63)};
     \addplot[cyan!80!white,mark=otimes*,mark options={solid},semithick] plot coordinates {
(5,7.32)(7,11.84)(9,8.59)(11,12.51)(13,13.75)(15,14.32)(17,15.20)(19,15.49)};
     \addplot[violet!80!white,mark=diamond*,mark options={solid},semithick] plot coordinates {
(5,9.06)(7,9.06)(9,14.11)(11,17.52)(13,22.52)(15,25.95)(17,28.00)(19,29.23)};
     \addplot[yellow!90!black,mark=pentagon,mark options={solid},semithick] plot coordinates {
(5,28.09)(7,28.37)(9,27.48)(11,36.47)(13,31.66)(15,27.09)(17,32.42)(19,34.48)};
     \addplot[gray,mark=otimes,mark options={solid},semithick] plot coordinates {
(5,24.48)(7,29.44)(9,37.95)(11,20.96)(13,21.60)(15,31.96)(17,36.65)(19,47.28)};

\legend{Geometric-AP,AP,kmedoids,kmeans,Spectral-g,HAC,GMM,DPGMM};

\end{axis}
\end{tikzpicture}
\caption{F1}
\label{fig:subfigurei1}
 \end{subfigure}
\caption{Hyperparameter tuning and evaluation results on the \emph{\textbf{cora}} dataset. (\textbf{a-c}) The three distance metrics with variable threshold values are tested to predict the ground-truth categories, i.e. 7. By referring to the NMI metric, the optimal clustering results are obtained using the $\mathtt{shortest\_path}$ distance and $\tau=3$.
(\textbf{d-f}) NMI, CR, and F1 evaluation metrics are reported for Geometric-AP and the rest of benchmark algorithms. (\textbf{g-i}) Plots of evaluation metrics as function of $K\left( \mbox{number of identified clusters}\right )$.}
\label{fig:Figure1}
\end{minipage}
\end{figure*}

%% file: figures/citeseer.tex
\begin{figure*}[ht!]
\centering
\begin{minipage}{0.9775\textwidth}
\begin{subfigure}[b]{0.33\textwidth}
\begin{tikzpicture}
    \begin{axis}[ width=\linewidth,font=\footnotesize,
    xlabel= $\tau$, ylabel= Percentage (\%),
  xmin=1,xmax=7,ymin=0,ymax=80,
    legend style={nodes=right},legend pos= north west,legend style={nodes={scale=0.7, transform shape}},
      xmajorgrids,
    grid style={dotted},
    ymajorgrids,
    xtick={1,2,3,4,5,6,7},
   ]
\addplot[smooth,blue!80!black,mark=diamond*,mark options={solid},semithick] plot coordinates {
(1,40.52)(2,45.45)(3,42.55)(4,40.83)(5,45.57)(6,44.60)(7,39.35)
};
     \addplot[smooth,red!80!black,mark=otimes*,mark options={solid},semithick] plot coordinates {
(1,30.96)(2,36.44)(3,31.92)(4,28.29)(5,33.96)(6,32.85)(7,26.81)
};
     \addplot[smooth,green!60!black,mark=triangle*,mark options={solid},semithick] plot coordinates {
(1,10.78)(2,16.18)(3,16.37)(4,14.53)(5,18.57)(6,18.53)(7,13.28)
};
\addplot [red, thick, dashed] coordinates {(5,0) (5,45.57) } ;

\legend{CR,F1,NMI};
\end{axis}
\end{tikzpicture}
\caption{Geometric-AP (Shortest Path)}
\label{fig:subfigurea2}
\end{subfigure}
\begin{subfigure}[b]{0.33\textwidth}
\begin{tikzpicture}
    \begin{axis}[ width=\linewidth, font=\footnotesize,
    xlabel= $\tau$, ylabel= Percentage (\%),
  xmin=0.5,xmax=0.9,ymin=0,ymax=80,xtick={0.5,0.6,0.7,0.8,0.9},
    legend style={nodes=right},legend pos= north west,legend style={nodes={scale=0.7, transform shape}},
      xmajorgrids,
    grid style={dotted},
    ymajorgrids,
   ]
     \addplot[smooth,blue!80!black,mark=diamond*,mark options={solid},semithick] plot coordinates {
(0.5,35.72)(0.6,40.92)(0.7,42.40)(0.8,45.87)(0.9,43.69)
};
     \addplot[smooth,red!80!black,mark=otimes*,mark options={solid},semithick] plot coordinates {
(0.5,26.21)(0.6,28.73)(0.7,35.48)(0.8,34.32)(0.9,34.68)
};
     \addplot[smooth,green!60!black,mark=triangle*,mark options={solid},semithick] plot coordinates {
(0.5,9.14)(0.6,13.95)(0.7,12.79)(0.8,17.53)(0.9,14.22)
};
\addplot [red, thick, dashed] coordinates {(0.8,0) (0.8,45.87) } ;

\legend{CR,F1,NMI};
\end{axis}
\end{tikzpicture}
\caption{Geometric AP (Jaccard Distance)}
\label{fig:subfigureb2}
\end{subfigure}
\begin{subfigure}[b]{0.33\textwidth}
\begin{tikzpicture}
    \begin{axis}[ width=\linewidth,font=\footnotesize,
    xlabel= $\tau$, ylabel= Percentage (\%),
    xmin=0.5,xmax=0.91,ymin=0,ymax=80,xtick={0.5,0.6,0.7,0.8,0.9},
    legend style={nodes=right},legend pos= north west,legend style={nodes={scale=0.7, transform shape}},
    xmajorgrids,
    grid style={dotted},
    ymajorgrids,
   ]
     \addplot[smooth,blue!80!black,mark=diamond*,mark options={solid},semithick] plot coordinates {
(0.4,35.72)(0.5,42.79)(0.6,37.84)(0.7,41.22)(0.8,41.28)(0.9,41.49)
};
     \addplot[smooth,red!80!black,mark=otimes*,mark options={solid},semithick] plot coordinates {
(0.4,26.21)(0.5,35.74)(0.6,25.92)(0.7,30.52)(0.8,30.54)(0.9,30.70)
};
     \addplot[smooth,green!60!black,mark=triangle*,mark options={solid},semithick] plot coordinates {
(0.4,9.14)(0.5,13.01)(0.6,11.89)(0.7,14.05)(0.8,14.01)(0.9,14.19)
};
\addplot [red, thick, dashed] coordinates {(0.9,0) (0.9,41.49) } ;

\legend{CR,F1,NMI};
\end{axis}
\end{tikzpicture}
\caption{Geometric-AP (Cosine Distance)}
\label{fig:subfigurec2}
\end{subfigure}
\par\bigskip 
\begin{subfigure}[b]{0.33\textwidth}
\begin{tikzpicture}
\begin{axis}[width=\linewidth,font=\footnotesize,
 ybar,bar shift=0,ymin=0,ymax=25,bar width=8pt,
ylabel={Percentage (\%)},
symbolic x coords={Geometric-AP,AP,kmedoids,zero1,kmeans,Spectral-g,HAC,GMM,DPGMM},
x tick label style={font=\scriptsize},
x tick label style={rotate=90,anchor=east,nodes near coords align={horizontal}},
xtick={Geometric-AP,AP,kmedoids,kmeans,Spectral-g,HAC,GMM,DPGMM},
ytick={0,5,10,15,20,25},
after end axis/.code={
\node(L1) at (axis cs:AP,1) {};
\node(L2) at (axis cs:HAC,1) {};
\node(XTL) at (xticklabel cs:0)  {};
\node[anchor=east](A1) at (axis cs:Geometric-AP,-14) {};
\node[anchor=west](B1) at (axis cs:kmedoids,-14) {};
\draw (A1|- XTL) -- (B1|- XTL);
\node[anchor=center,below] at (L1 |- XTL) {\scriptsize{Exemplar-Based}};
\node[anchor=east](A2) at (axis cs:kmeans,-14) {};
\node[anchor=west](B2) at (axis cs:DPGMM,-14) {};
\draw (A2|- XTL) -- (B2|- XTL);
\node[anchor=center,below] at (L2 |- XTL) {\scriptsize{Non Exemplar-Based}};
}
]
\addplot[red!80!black,fill=red!30!white,mark=none] coordinates {(Geometric-AP,18.57)};

\addplot[blue!80!black,fill=blue!30!white,mark=none] coordinates {(AP,8.02)};

\addplot[green!60!black,fill=green!30!white,mark=none] coordinates {(kmedoids,7.92	)};

\addplot[xtick=\empty, draw=none] coordinates {(zero1,0)};

\addplot[orange!80!black,fill=orange!30!white,mark=none] coordinates {(kmeans,19.24)};

\addplot[cyan!80!black,fill=cyan!30!white,mark=none] coordinates {(Spectral-g,2.3)};

\addplot[violet!80!black,fill=violet!30!white,mark=none] coordinates {(HAC,3.05)};

\addplot[yellow!80!black,fill=yellow!50!white,mark=none] coordinates {(GMM,20.66)};

\addplot[black,fill=lightgray,mark=none] coordinates {(DPGMM,15.78)};
\end{axis}

\end{tikzpicture}
\caption{NMI}
\label{fig:subfigured2}
\end{subfigure}
\begin{subfigure}[b]{0.33\textwidth}
\begin{tikzpicture}
\begin{axis}[width=\linewidth,font=\footnotesize,
 ybar,bar shift=0,ymin=0,ymax=50,bar width=8pt,
ylabel={Percentage (\%)},
symbolic x coords={Geometric-AP,AP,kmedoids,zero1,kmeans,Spectral-g,HAC,GMM,DPGMM},
x tick label style={font=\scriptsize},
x tick label style={rotate=90,anchor=east,nodes near coords align={horizontal}},
xtick={Geometric-AP,AP,kmedoids,kmeans,Spectral-g,HAC,GMM,DPGMM},
ytick={0,10,20,30,40,50},
after end axis/.code={
\node(L1) at (axis cs:AP,1) {};
\node(L2) at (axis cs:HAC,1) {};
\node(XTL) at (xticklabel cs:0)  {};
\node[anchor=east](A1) at (axis cs:Geometric-AP,-14) {};
\node[anchor=west](B1) at (axis cs:kmedoids,-14) {};
\draw (A1|- XTL) -- (B1|- XTL);
\node[anchor=center,below] at (L1 |- XTL) {\scriptsize{Exemplar-Based}};
\node[anchor=east](A2) at (axis cs:kmeans,-14) {};
\node[anchor=west](B2) at (axis cs:DPGMM,-14) {};
\draw (A2|- XTL) -- (B2|- XTL);
\node[anchor=center,below] at (L2 |- XTL) {\scriptsize{Non Exemplar-Based}};
}
]
\addplot[red!80!black,fill=red!30!white,mark=none] coordinates {(Geometric-AP,45.57	)};

\addplot[blue!80!black,fill=blue!30!white,mark=none] coordinates {(AP,37.05)};

\addplot[green!60!black,fill=green!30!white,mark=none] coordinates {(kmedoids,38.14	)};

\addplot[xtick=\empty, draw=none] coordinates {(zero1,0)};

\addplot[orange!80!black,fill=orange!30!white,mark=none] coordinates {(kmeans,43.82)};

\addplot[cyan!80!black,fill=cyan!30!white,mark=none] coordinates {(Spectral-g,22.77)};

\addplot[violet!80!black,fill=violet!30!white,mark=none] coordinates {(HAC,23.77)};

\addplot[yellow!80!black,fill=yellow!50!white,mark=none] coordinates {(GMM,42.06)};

\addplot[black,fill=lightgray,mark=none] coordinates {(DPGMM,42.34)};
\end{axis}
\end{tikzpicture}
\caption{CR}
\label{fig:subfiguree2}
\end{subfigure}
\begin{subfigure}[b]{0.33\textwidth}
\begin{tikzpicture}
\begin{axis}[width=\linewidth,font=\footnotesize,
 ybar,bar shift=0,ymin=0,ymax=40,bar width=8pt,
ylabel={Percentage (\%)},
symbolic x coords={Geometric-AP,AP,kmedoids,zero1,kmeans,Spectral-g,HAC,GMM,DPGMM},
x tick label style={font=\scriptsize},
x tick label style={rotate=90,anchor=east,nodes near coords align={horizontal}},
xtick={Geometric-AP,AP,kmedoids,kmeans,Spectral-g,HAC,GMM,DPGMM},
ytick={0,10,20,30,40},
after end axis/.code={
\node(L1) at (axis cs:AP,1) {};
\node(L2) at (axis cs:HAC,1) {};
\node(XTL) at (xticklabel cs:0)  {};
\node[anchor=east](A1) at (axis cs:Geometric-AP,-14) {};
\node[anchor=west](B1) at (axis cs:kmedoids,-14) {};
\draw (A1|- XTL) -- (B1|- XTL);
\node[anchor=center,below] at (L1 |- XTL) {\scriptsize{Exemplar-Based}};
\node[anchor=east](A2) at (axis cs:kmeans,-14) {};
\node[anchor=west](B2) at (axis cs:DPGMM,-14) {};
\draw (A2|- XTL) -- (B2|- XTL);
\node[anchor=center,below] at (L2 |- XTL) {\scriptsize{Non Exemplar-Based}};
}
]
\addplot[red!80!black,fill=red!30!white,mark=none] coordinates {(Geometric-AP,33.96)};

\addplot[blue!80!black,fill=blue!30!white,mark=none] coordinates {(AP,27.45)};

\addplot[green!60!black,fill=green!30!white,mark=none] coordinates {(kmedoids,28.71)};

\addplot[xtick=\empty, draw=none] coordinates {(zero1,0)};

\addplot[orange!80!black,fill=orange!30!white,mark=none] coordinates {(kmeans,34.82)};

\addplot[cyan!80!black,fill=cyan!30!white,mark=none] coordinates {(Spectral-g,11.03)};

\addplot[violet!80!black,fill=violet!30!white,mark=none] coordinates {(HAC,12.08)};

\addplot[yellow!80!black,fill=yellow!50!white,mark=none] coordinates {(GMM,27.43)};

\addplot[black,fill=lightgray,mark=none] coordinates {(DPGMM,27.66)};
\end{axis}

\end{tikzpicture}
\caption{F1}
\label{fig:subfiguref2}
\end{subfigure}
\par\bigskip 
\begin{subfigure}[b]{0.33\textwidth}
\begin{tikzpicture}
    \begin{axis}[width=\linewidth,font=\footnotesize,
    xlabel= $K$, ylabel= Percentage (\%),
  ymin=0,ymax=50,xmin=4,xmax=18,legend columns=2,
    legend style={nodes=right,/tikz/column 2/.style={
                column sep=2pt,},},legend pos= north west,legend style={nodes={scale=0.6, transform shape}},
      xmajorgrids,
    grid style={dotted},
    ymajorgrids,
    xtick={4,6,8,10,12,14,16,18},
    ytick={0,10,20,30,40,50},
    ]
\addplot[magenta!80!white,mark=pentagon*,mark options={solid},semithick] plot coordinates {
(4,15.28)(6,18.57)(8,16.49)(10,17.73)(12,17.36)(14,16.80)(16,17.13)(18,17.87)
};
     \addplot[blue!80!white,mark=triangle*,mark options={solid},semithick] plot coordinates {
(4,6.18)(6,8.02)(8,11.62)(10,9.72)(12,11.74)(14,11.09)(16,13.19)(18,14.04)
};
     \addplot[green!80!white,mark=triangle,mark options={solid},semithick] plot coordinates {
(4,6.18)(6,7.92)(8,8.86)(10,9.84)(12,10.11)(14,9.81)(16,9.72)(18,9.79)
};
     \addplot[orange!80!white,mark=diamond,mark options={solid},semithick] plot coordinates {
(4,18.44)(6,19.24)(8,21.87)(10,23.45)(12,20.84)(14,21.45)(16,19.17)(18,20.41)
};
     \addplot[cyan!80!white,mark=otimes*,mark options={solid},semithick] plot coordinates {
(4,1.14)(6,2.30)(8,1.92)(10,6.38)(12,3.81)(14,5.45)(16,14.93)(18,15.35)
};
     \addplot[violet!80!white,mark=diamond*,mark options={solid},semithick] plot coordinates {
(4,2.36)(6,3.05)(8,4.07)(10,4.47)(12,5.55)(14,5.96)(16,6.56)(18,7.06)
};
     \addplot[yellow!90!black,mark=pentagon,mark options={solid},semithick] plot coordinates {
(4,19.25)(6,20.66)(8,21.68)(10,23.15)(12,19.07)(14,20.64)(16,19.87)(18,18.24)
};
     \addplot[gray,mark=otimes,mark options={solid},semithick] plot coordinates {
(4,19.21)(6,21.54)(8,18.20)(10,17.03)(12,22.11)(14,18.56)(16,20.20)(18,22.06)
};
\legend{Geometric-AP,AP,kmedoids,kmeans,Spectral-g,HAC,GMM,DPGMM};
\end{axis}
\end{tikzpicture}
\caption{NMI}
\label{fig:subfigureg2}
\end{subfigure}
\begin{subfigure}[b]{0.33\textwidth}
\begin{tikzpicture}
    \begin{axis}[width=\linewidth,font=\footnotesize,
    xlabel= $K$, ylabel= Percentage (\%),
    ymin=20,ymax=100,xmin=4,xmax=18,legend columns=2,
    legend style={nodes=right,/tikz/column 2/.style={
                column sep=2pt,},},legend pos= north west,legend style={nodes={scale=0.6}},
      xmajorgrids,
    grid style={dotted},
    ymajorgrids,
      xtick={4,6,8,10,12,14,16,18},
   ]
\addplot[magenta!80!white,mark=pentagon*,mark options={solid},semithick] plot coordinates {
(4,42.00)(6,45.57)(8,44.30)(10,46.71)(12,47.17)(14,46.26)(16,47.05)(18,48.47)
};
     \addplot[blue!80!white,mark=triangle*,mark options={solid},semithick] plot coordinates {
(4,35.51)(6,37.05)(8,43.27)(10,41.58)(12,44.03)(14,44.18)(16,46.65)(18,47.59)
};
     \addplot[green!80!white,mark=triangle,mark options={solid},semithick] plot coordinates {
(4,34.52)(6,38.14)(8,39.92)(10,41.13)(12,41.85)(14,41.82)(16,42.12)(18,42.25)
};
     \addplot[orange!80!white,mark=diamond,mark options={solid},semithick] plot coordinates {
(4,40.65)(6,43.82)(8,50.49)(10,53.96)(12,50.52)(14,52.06)(16,50.58)(18,51.09)
};
     \addplot[cyan!80!white,mark=otimes*,mark options={solid},semithick] plot coordinates {
(4,21.90)(6,22.77)(8,22.88)(10,26.76)(12,23.62)(14,25.22)(16,39.98)(18,40.10)
};
     \addplot[violet!80!white,mark=diamond*,mark options={solid},semithick] plot coordinates {
(4,23.07)(6,23.77)(8,24.28)(10,25.82)(12,27.12)(14,27.63)(16,28.02)(18,28.36)
};
     \addplot[yellow!90!black,mark=pentagon,mark options={solid},semithick] plot coordinates {
(4,40.74)(6,42.06)(8,36.81)(10,51.76)(12,45.96)(14,43.12)(16,46.44)(18,42.46)
};
     \addplot[gray,mark=otimes,mark options={solid},semithick] plot coordinates {
(4,45.02)(6,46.44)(8,44.21)(10,43.18)(12,51.94)(14,44.66)(16,50.37)(18,53.27)
};
\legend{Geometric-AP,AP,kmedoids,kmeans,Spectral-g,HAC,GMM,DPGMM};

\end{axis}
\end{tikzpicture}
\caption{CR}
\label{fig:subfigureh2}
\end{subfigure}
\begin{subfigure}[b]{0.33\textwidth}
\begin{tikzpicture}
    \begin{axis}[width=\linewidth,font=\footnotesize,
    xlabel= $K$, ylabel= Percentage (\%),
    ymin=0,ymax=100,xmin=4,xmax=18,legend columns=2,
    legend style={nodes=right,/tikz/column 2/.style={
                column sep=2pt,},},legend pos= north west,legend style={nodes={scale=0.6, transform shape}},
      xmajorgrids,
    grid style={dotted},
    ymajorgrids,
    xtick={4,6,8,10,12,14,16,18},
   ]
\addplot[magenta!80!white,mark=pentagon*,mark options={solid},semithick] plot coordinates {
(4,29.82)(6,33.96)(8,32.55)(10,34.48)(12,34.81)(14,36.64)(16,35.97)(18,37.21)
};
     \addplot[blue!80!white,mark=triangle*,mark options={solid},semithick] plot coordinates {
(4,26.76)(6,27.45)(8,36.71)(10,35.06)(12,37.72)(14,37.57)(16,39.93)(18,40.45)
};
     \addplot[green!80!white,mark=triangle,mark options={solid},semithick] plot coordinates {
(4,21.70)(6,28.71)(8,33.72)(10,34.27)(12,35.38)(14,35.44)(16,35.96)(18,35.66)
};
     \addplot[orange!80!white,mark=diamond,mark options={solid},semithick] plot coordinates {
(4,28.17)(6,34.82)(8,44.16)(10,46.68)(12,47.87)(14,44.91)(16,43.78)(18,44.42)
};
     \addplot[cyan!80!white,mark=otimes*,mark options={solid},semithick] plot coordinates {
(4,7.83)(6,11.03)(8,11.47)(10,15.80)(12,14.74)(14,14.91)(16,29.42)(18,29.13)
};
     \addplot[violet!80!white,mark=diamond*,mark options={solid},semithick] plot coordinates {
(4,11.34)(6,12.08)(8,13.04)(10,16.82)(12,18.65)(14,19.17)(16,21.34)(18,21.92)
};
     \addplot[yellow!90!black,mark=pentagon,mark options={solid},semithick] plot coordinates {
(4,28.35)(6,27.43)(8,18.67)(10,46.68)(12,38.47)(14,33.25)(16,38.95)(18,32.87)
};
     \addplot[gray,mark=otimes,mark options={solid},semithick] plot coordinates {
(4,35.17)(6,36.21)(8,34.77)(10,37.60)(12,45.70)(14,34.64)(16,43.95)(18,46.56)
};
\legend{Geometric-AP,AP,kmedoids,kmeans,Spectral-g,HAC,GMM,DPGMM};

\end{axis}
\end{tikzpicture}
\caption{F1}
\label{fig:subfigurei2}
 \end{subfigure}
\caption{Hyperparameter optimization and evaluation results on the \emph{\textbf{citeseer}} dataset. (\textbf{a-c}) Three topological distances are tested over a range of threshold values to predict the ground-truth categories, i.e. 6. By consulting the NMI measure, the optimal clustering results are obtained using $\mathtt{shortest\_path}$ distance and $\tau=5$.
(\textbf{d-f}) NMI, CR, and F1 evaluation metrics are reported for the set of tested algorithms. (\textbf{g-i}) Plots of evaluation metrics as functions of $K\left( \mbox{number of identified clusters}\right )$.}
\label{fig:Figure2}
\end{minipage}
\end{figure*}

%% file: figures/karate_split.tex
\begin{figure*}[ht!]
\centering
\begin{subfigure}[b]{0.33\textwidth}
\begin{tikzpicture}
    \begin{axis}[ width=\linewidth,font=\footnotesize,
    xlabel= Shortest Path Distance, ylabel= Percentage (\%),
  xmin=0.9,xmax=5,ymin=0,ymax=100,
    legend style={nodes=right},legend pos= south east,legend style={nodes={scale=0.8, transform shape}},
      xmajorgrids,
    grid style={dotted},
    ymajorgrids,
    xtick={1,2,3,4,5},
   ]
\addplot[smooth,blue!80!black,mark=diamond*,mark options={solid},semithick] plot coordinates {
(1,91.18)(2,82.36)(3,82.36)(4,82.36)(5,82.36)
};
     \addplot[smooth,red!80!black,mark=otimes*,mark options={solid},semithick] plot coordinates {
(1,91.17)(2,81.79)(3,81.79)(4,81.79)(5,81.79)
};
     \addplot[smooth,green!60!black,mark=triangle*,mark options={solid},semithick] plot coordinates {
(1,57.56)(2,46.11)(3,46.11)(4,46.11)(5,46.11)
};
\addplot [red, thick, dashed] coordinates {(1,0) (1,91.18) } ;

\legend{CR,F1,NMI};
\end{axis}
\end{tikzpicture}
\caption{Geometric-AP (Shortest Path)}
\label{fig:subfigurea3}
\end{subfigure}
\begin{subfigure}[b]{0.33\textwidth}
\begin{tikzpicture}
    \begin{axis}[ width=\linewidth, font=\footnotesize,
    xlabel= Jaccard Distance, ylabel= Percentage (\%),
  xmin=0.49,xmax=0.9,ymin=0,ymax=100,xtick={0.5,0.6,0.7,0.8,0.9},
    legend style={nodes=right},legend pos= south east,legend style={nodes={scale=0.8, transform shape}},
      xmajorgrids,
    grid style={dotted},
    ymajorgrids,
   ]
     \addplot[smooth,blue!80!black,mark=diamond*,mark options={solid},semithick] plot coordinates {
(0.5,97.06)(0.6,97.06)(0.7,82.36)(0.8,79.42)(0.9,82.36)
};
     \addplot[smooth,red!80!black,mark=otimes*,mark options={solid},semithick] plot coordinates {
(0.5,97.06)(0.6,97.06)(0.7,81.79)(0.8,78.51)(0.9,81.79)
};
     \addplot[smooth,green!60!black,mark=triangle*,mark options={solid},semithick] plot coordinates {
(0.5,83.72)(0.6,83.72)(0.7,46.11)(0.8,41.12)(0.9,46.11)
};
\addplot [red, thick, dashed] coordinates {(0.5,0) (0.5,97.06) } ;

\legend{CR,F1,NMI};
\end{axis}
\end{tikzpicture}
\caption{Geometric AP (Jaccard Distance)}
\label{fig:subfigureb3}
\end{subfigure}
\begin{subfigure}[b]{0.33\textwidth}
\begin{tikzpicture}
    \begin{axis}[ width=\linewidth,font=\footnotesize,
    xlabel= Cosine Distance, ylabel= Percentage (\%),
    xmin=0.5,xmax=0.9,ymin=0,ymax=100,xtick={0.5,0.6,0.7,0.8,0.9},
    legend style={nodes=right},legend pos= south east,legend style={nodes={scale=0.8, transform shape}},
    xmajorgrids,
    grid style={dotted},
    ymajorgrids,
   ]
     \addplot[smooth,blue!80!black,mark=diamond*,mark options={solid},semithick] plot coordinates {
(0.5,79.42)(0.6,82.36)(0.7,82.36)(0.8,82.36)(0.9,82.36)
};
     \addplot[smooth,red!80!black,mark=otimes*,mark options={solid},semithick] plot coordinates {
(0.5,78.51)(0.6,81.79)(0.7,81.79)(0.8,81.79)(0.9,81.79)
};
     \addplot[smooth,green!60!black,mark=triangle*,mark options={solid},semithick] plot coordinates {
(0.5,41.12)(0.6,46.11)(0.7,46.11)(0.8,46.11)(0.9,46.11)
};
\addplot [red, thick, dashed] coordinates {(0.6,0) (0.6,82.36) } ;

\legend{CR,F1,NMI};
\end{axis}
\end{tikzpicture}
\caption{Geometric-AP (Cosine Distance)}
\label{fig:subfigurec3}
\end{subfigure}
\par\bigskip 

\begin{subfigure}[b]{0.33\textwidth}
\begin{tikzpicture}
\begin{axis}[width=\linewidth,font=\footnotesize,
 ybar,bar shift=0,ymin=0,ymax=110,ytick={0,20,40,60,80,100},bar width=8pt,
ylabel={Percentage (\%)},
symbolic x coords={Geometric-AP,AP,kmedoids,zero1,kmeans,Spectral-g,HAC,GMM,DPGMM},
x tick label style={font=\scriptsize},
x tick label style={rotate=90,anchor=east,nodes near coords align={horizontal}},
xtick={Geometric-AP,AP,kmedoids,kmeans,Spectral-g,HAC,GMM,DPGMM},
after end axis/.code={
\node(L1) at (axis cs:AP,1) {};
\node(L2) at (axis cs:HAC,1) {};
\node(XTL) at (xticklabel cs:0)  {};
\node[anchor=east](A1) at (axis cs:Geometric-AP,-14) {};
\node[anchor=west](B1) at (axis cs:kmedoids,-14) {};
\draw (A1|- XTL) -- (B1|- XTL);
\node[anchor=center,below] at (L1 |- XTL) {\scriptsize{Exemplar-Based}};
\node[anchor=east](A2) at (axis cs:kmeans,-14) {};
\node[anchor=west](B2) at (axis cs:DPGMM,-14) {};
\draw (A2|- XTL) -- (B2|- XTL);
\node[anchor=center,below] at (L2 |- XTL) {\scriptsize{Non Exemplar-Based}};
}
]
\addplot[red!80!black,fill=red!30!white,mark=none] coordinates {(Geometric-AP,83.72)};

\addplot[blue!80!black,fill=blue!30!white,mark=none] coordinates {(AP,31.7)};

\addplot[green!60!black,fill=green!30!white,mark=none] coordinates {(kmedoids,73.24)};

\addplot[xtick=\empty, draw=none] coordinates {(zero1,0)};

\addplot[orange!80!black,fill=orange!30!white,mark=none] coordinates {(kmeans,83.72)};

\addplot[cyan!80!black,fill=cyan!30!white,mark=none] coordinates {(Spectral-g,73.24)};

\addplot[violet!80!black,fill=violet!30!white,mark=none] coordinates {(HAC,9.3)};

\addplot[yellow!80!black,fill=yellow!50!white,mark=none] coordinates {(GMM,83.72)};

\addplot[black,fill=lightgray,mark=none] coordinates {(DPGMM,83.72)};
\end{axis}

\end{tikzpicture}
\caption{NMI}
\label{fig:subfigured3}
\end{subfigure}
\begin{subfigure}[b]{0.33\textwidth}
\begin{tikzpicture}
\begin{axis}[width=\linewidth,font=\footnotesize,
 ybar,bar shift=0,ymin=0,ymax=110,ytick={0,20,40,60,80,100},bar width=8pt,
ylabel={Percentage (\%)},
symbolic x coords={Geometric-AP,AP,kmedoids,zero1,kmeans,Spectral-g,HAC,GMM,DPGMM},
x tick label style={font=\scriptsize},
x tick label style={rotate=90,anchor=east,nodes near coords align={horizontal}},
xtick={Geometric-AP,AP,kmedoids,kmeans,Spectral-g,HAC,GMM,DPGMM},
after end axis/.code={
\node(L1) at (axis cs:AP,1) {};
\node(L2) at (axis cs:HAC,1) {};
\node(XTL) at (xticklabel cs:0)  {};
\node[anchor=east](A1) at (axis cs:Geometric-AP,-14) {};
\node[anchor=west](B1) at (axis cs:kmedoids,-14) {};
\draw (A1|- XTL) -- (B1|- XTL);
\node[anchor=center,below] at (L1 |- XTL) {\scriptsize{Exemplar-Based}};
\node[anchor=east](A2) at (axis cs:kmeans,-14) {};
\node[anchor=west](B2) at (axis cs:DPGMM,-14) {};
\draw (A2|- XTL) -- (B2|- XTL);
\node[anchor=center,below] at (L2 |- XTL) {\scriptsize{Non Exemplar-Based}};
}
]
\addplot[red!80!black,fill=red!30!white,mark=none] coordinates {(Geometric-AP,97.06	)};

\addplot[blue!80!black,fill=blue!30!white,mark=none] coordinates {(AP,79.42)};

\addplot[green!60!black,fill=green!30!white,mark=none] coordinates {(kmedoids,94.12)};

\addplot[xtick=\empty, draw=none] coordinates {(zero1,0)};

\addplot[orange!80!black,fill=orange!30!white,mark=none] coordinates {(kmeans,97.06)};

\addplot[cyan!80!black,fill=cyan!30!white,mark=none] coordinates {(Spectral-g,94.12)};

\addplot[violet!80!black,fill=violet!30!white,mark=none] coordinates {(HAC,55.89)};

\addplot[yellow!80!black,fill=yellow!50!white,mark=none] coordinates {(GMM,97.06)};

\addplot[black,fill=lightgray,mark=none] coordinates {(DPGMM,97.06)};
\end{axis}
\end{tikzpicture}
\caption{CR}
\label{fig:subfiguree3}
\end{subfigure}
\begin{subfigure}[b]{0.33\textwidth}
\begin{tikzpicture}
\begin{axis}[width=\linewidth,font=\footnotesize,
 ybar,bar shift=0,ymin=0,ymax=110,ytick={0,20,40,60,80,100},bar width=8pt,
ylabel={Percentage (\%)},
symbolic x coords={Geometric-AP,AP,kmedoids,zero1,kmeans,Spectral-g,HAC,GMM,DPGMM},
x tick label style={font=\scriptsize},
x tick label style={rotate=90,anchor=east,nodes near coords align={horizontal}},
xtick={Geometric-AP,AP,kmedoids,kmeans,Spectral-g,HAC,GMM,DPGMM},
after end axis/.code={
\node(L1) at (axis cs:AP,1) {};
\node(L2) at (axis cs:HAC,1) {};
\node(XTL) at (xticklabel cs:0)  {};
\node[anchor=east](A1) at (axis cs:Geometric-AP,-14) {};
\node[anchor=west](B1) at (axis cs:kmedoids,-14) {};
\draw (A1|- XTL) -- (B1|- XTL);
\node[anchor=center,below] at (L1 |- XTL) {\scriptsize{Exemplar-Based}};
\node[anchor=east](A2) at (axis cs:kmeans,-14) {};
\node[anchor=west](B2) at (axis cs:DPGMM,-14) {};
\draw (A2|- XTL) -- (B2|- XTL);
\node[anchor=center,below] at (L2 |- XTL) {\scriptsize{Non Exemplar-Based}};
}
]
\addplot[red!80!black,fill=red!30!white,mark=none] coordinates {(Geometric-AP,97.06)};

\addplot[blue!80!black,fill=blue!30!white,mark=none] coordinates {(AP,78.95)};

\addplot[green!60!black,fill=green!30!white,mark=none] coordinates {(kmedoids,94.1)};

\addplot[xtick=\empty, draw=none] coordinates {(zero1,0)};

\addplot[orange!80!black,fill=orange!30!white,mark=none] coordinates {(kmeans,97.06)};

\addplot[cyan!80!black,fill=cyan!30!white,mark=none] coordinates {(Spectral-g,94.1)};

\addplot[violet!80!black,fill=violet!30!white,mark=none] coordinates {(HAC,45.23)};

\addplot[yellow!80!black,fill=yellow!50!white,mark=none] coordinates {(GMM,97.06)};

\addplot[black,fill=lightgray,mark=none] coordinates {(DPGMM,97.06)};
\end{axis}

\end{tikzpicture}
\caption{F1}
\label{fig:subfiguref3}
\end{subfigure}

\caption{Hyperparameter tuning and evaluation results on the \emph{\textbf{Zachary's Karate Club}} network using the club split as ground-truth classes. (\textbf{a-c}) Selected topological distances are optimized over a range of threshold values to predict the ground-truth categories, i.e. 2. By referring to the NMI measure, the optimal clustering results are produced via $\mathtt{Jaccard}$ distance and $\tau=0.5$.
(\textbf{d-f}) NMI, CR, and F1 evaluation metrics are reported for the different tested algorithms.}
\label{fig:Figure3}
\end{figure*}


%% file: figures/karate_modularity.tex
\begin{figure*}[ht!]
\centering
\begin{subfigure}[b]{0.33\textwidth}
\begin{tikzpicture}
    \begin{axis}[ width=\linewidth,font=\footnotesize,
    xlabel= Shortest Path Distance, ylabel= Percentage (\%),
  xmin=1,xmax=5,ymin=40,ymax=100,
    legend style={nodes=right},legend pos= south east,legend style={nodes={scale=0.8, transform shape}},
      xmajorgrids,
    grid style={dotted},
    ymajorgrids,
    xtick={1,2,3,4,5},
   ]
\addplot[smooth,blue!80!black,mark=diamond*,mark options={solid},semithick] plot coordinates {
(1,73.53)(2,76.48)(3,76.48)(4,76.48)(5,76.48)
};
     \addplot[smooth,red!80!black,mark=otimes*,mark options={solid},semithick] plot coordinates {
(1,64.36)(2,64.85)(3,64.85)(4,64.85)(5,64.85)
};
     \addplot[smooth,green!60!black,mark=triangle*,mark options={solid},semithick] plot coordinates {
(1,60.27)(2,65.71)(3,65.71)(4,65.71)(5,65.71)
};
\addplot [red, thick, dashed] coordinates {(2,0) (2,76.48) } ;

\legend{CR,F1,NMI};
\end{axis}
\end{tikzpicture}
\caption{Geometric-AP (Shortest Path)}
\label{fig:subfigurea4}
\end{subfigure}
\begin{subfigure}[b]{0.33\textwidth}
\begin{tikzpicture}
    \begin{axis}[ width=\linewidth, font=\footnotesize,
    xlabel= Jaccard Distance, ylabel= Percentage (\%),
  xmin=0.5,xmax=0.9,ymin=40,ymax=100,xtick={0.5,0.6,0.7,0.8,0.9},
    legend style={nodes=right},legend pos= south west,legend style={nodes={scale=0.8, transform shape}},
      xmajorgrids,
    grid style={dotted},
    ymajorgrids,
   ]
     \addplot[smooth,blue!80!black,mark=diamond*,mark options={solid},semithick] plot coordinates {
(0.5,79.42)(0.6,76.48)(0.7,82.36)(0.8,82.36)(0.9,76.48)
};
     \addplot[smooth,red!80!black,mark=otimes*,mark options={solid},semithick] plot coordinates {
(0.5,72.17)(0.6,71.51)(0.7,78.36)(0.8,75.75)(0.9,67.27)
};
     \addplot[smooth,green!60!black,mark=triangle*,mark options={solid},semithick] plot coordinates {
(0.5,68.19)(0.6,64.62)(0.7,66.93)(0.8,76.76)(0.9,62.57)
};
\addplot [red, thick, dashed] coordinates {(0.8,0) (0.8,82.36) } ;

\legend{CR,F1,NMI};
\end{axis}
\end{tikzpicture}
\caption{Geometric AP (Jaccard Distance)}
\label{fig:subfigureb4}
\end{subfigure}
\begin{subfigure}[b]{0.33\textwidth}
\begin{tikzpicture}
    \begin{axis}[ width=\linewidth,font=\footnotesize,
    xlabel= Cosine Distance, ylabel= Percentage (\%),
    xmin=0.5,xmax=0.9,ymin=40,ymax=100,xtick={0.5,0.6,0.7,0.8,0.9},
    legend style={nodes=right},legend pos= south east,legend style={nodes={scale=0.8, transform shape}},
    xmajorgrids,
    grid style={dotted},
    ymajorgrids,
   ]
     \addplot[smooth,blue!80!black,mark=diamond*,mark options={solid},semithick] plot coordinates {
(0.5,79.42)(0.6,82.36)(0.7,76.48)(0.8,79.42)(0.9,79.42)
};
     \addplot[smooth,red!80!black,mark=otimes*,mark options={solid},semithick] plot coordinates {
(0.5,68.81)(0.6,75.75)(0.7,67.27)(0.8,71.84)(0.9,71.84)
};
     \addplot[smooth,green!60!black,mark=triangle*,mark options={solid},semithick] plot coordinates {
(0.5,70.34)(0.6,76.76)(0.7,62.57)(0.8,67.95)(0.9,67.95)
};
\addplot [red, thick, dashed] coordinates {(0.6,0) (0.6,82.36) } ;

\legend{CR,F1,NMI};
\end{axis}
\end{tikzpicture}
\caption{Geometric-AP (Cosine Distance)}
\label{fig:subfigurec4}
\end{subfigure}
\par\bigskip 
\begin{subfigure}[b]{0.33\textwidth}
\begin{tikzpicture}
\begin{axis}[width=\linewidth,font=\footnotesize,
 ybar,bar shift=0,ymin=0,ymax=100,ytick={0,20,40,60,80,100},bar width=8pt,
ylabel={Percentage (\%)},
symbolic x coords={Geometric-AP,AP,kmedoids,zero1,kmeans,Spectral-g,HAC,GMM,DPGMM},
x tick label style={font=\scriptsize},
x tick label style={rotate=90,anchor=east,nodes near coords align={horizontal}},
xtick={Geometric-AP,AP,kmedoids,kmeans,Spectral-g,HAC,GMM,DPGMM},
after end axis/.code={
\node(L1) at (axis cs:AP,1) {};
\node(L2) at (axis cs:HAC,1) {};
\node(XTL) at (xticklabel cs:0)  {};
\node[anchor=east](A1) at (axis cs:Geometric-AP,-14) {};
\node[anchor=west](B1) at (axis cs:kmedoids,-14) {};
\draw (A1|- XTL) -- (B1|- XTL);
\node[anchor=center,below] at (L1 |- XTL) {\scriptsize{Exemplar-Based}};
\node[anchor=east](A2) at (axis cs:kmeans,-14) {};
\node[anchor=west](B2) at (axis cs:DPGMM,-14) {};
\draw (A2|- XTL) -- (B2|- XTL);
\node[anchor=center,below] at (L2 |- XTL) {\scriptsize{Non Exemplar-Based}};
}
]
\addplot[red!80!black,fill=red!30!white,mark=none] coordinates {(Geometric-AP,76.76)};

\addplot[blue!80!black,fill=blue!30!white,mark=none] coordinates {(AP,64.72)};

\addplot[green!60!black,fill=green!30!white,mark=none] coordinates {(kmedoids,65.08)};

\addplot[xtick=\empty, draw=none] coordinates {(zero1,0)};

\addplot[orange!80!black,fill=orange!30!white,mark=none] coordinates {(kmeans,57.98)};

\addplot[cyan!80!black,fill=cyan!30!white,mark=none] coordinates {(Spectral-g,76.01)};

\addplot[violet!80!black,fill=violet!30!white,mark=none] coordinates {(HAC,16.7)};

\addplot[yellow!80!black,fill=yellow!50!white,mark=none] coordinates {(GMM,67.28)};

\addplot[black,fill=lightgray,mark=none] coordinates {(DPGMM,55.9)};
\end{axis}

\end{tikzpicture}
\caption{NMI}
\label{fig:subfigured4}
\end{subfigure}
\begin{subfigure}[b]{0.33\textwidth}
\begin{tikzpicture}
\begin{axis}[width=\linewidth,font=\footnotesize,
 ybar,bar shift=0,ymin=0,ymax=100,ytick={0,20,40,60,80,100},bar width=8pt,
ylabel={Percentage (\%)},
symbolic x coords={Geometric-AP,AP,kmedoids,zero1,kmeans,Spectral-g,HAC,GMM,DPGMM},
x tick label style={font=\scriptsize},
x tick label style={rotate=90,anchor=east,nodes near coords align={horizontal}},
xtick={Geometric-AP,AP,kmedoids,kmeans,Spectral-g,HAC,GMM,DPGMM},
after end axis/.code={
\node(L1) at (axis cs:AP,1) {};
\node(L2) at (axis cs:HAC,1) {};
\node(XTL) at (xticklabel cs:0)  {};
\node[anchor=east](A1) at (axis cs:Geometric-AP,-14) {};
\node[anchor=west](B1) at (axis cs:kmedoids,-14) {};
\draw (A1|- XTL) -- (B1|- XTL);
\node[anchor=center,below] at (L1 |- XTL) {\scriptsize{Exemplar-Based}};
\node[anchor=east](A2) at (axis cs:kmeans,-14) {};
\node[anchor=west](B2) at (axis cs:DPGMM,-14) {};
\draw (A2|- XTL) -- (B2|- XTL);
\node[anchor=center,below] at (L2 |- XTL) {\scriptsize{Non Exemplar-Based}};
}
]
\addplot[red!80!black,fill=red!30!white,mark=none] coordinates {(Geometric-AP,82.36)};

\addplot[blue!80!black,fill=blue!30!white,mark=none] coordinates {(AP,79.42)};

\addplot[green!60!black,fill=green!30!white,mark=none] coordinates {(kmedoids,79.42)};

\addplot[xtick=\empty, draw=none] coordinates {(zero1,0)};

\addplot[orange!80!black,fill=orange!30!white,mark=none] coordinates {(kmeans,64.71)};

\addplot[cyan!80!black,fill=cyan!30!white,mark=none] coordinates {(Spectral-g,88.24)};

\addplot[violet!80!black,fill=violet!30!white,mark=none] coordinates {(HAC,44.12)};

\addplot[yellow!80!black,fill=yellow!50!white,mark=none] coordinates {(GMM,76.48)};

\addplot[black,fill=lightgray,mark=none] coordinates {(DPGMM,73.53)};
\end{axis}
\end{tikzpicture}
\caption{CR}
\label{fig:subfiguree4}
\end{subfigure}
\begin{subfigure}[b]{0.33\textwidth}
\begin{tikzpicture}
\begin{axis}[width=\linewidth,font=\footnotesize,
 ybar,bar shift=0,ymin=0,ymax=100,ytick={0,20,40,60,80,100},bar width=8pt,
ylabel={Percentage (\%)},
symbolic x coords={Geometric-AP,AP,kmedoids,zero1,kmeans,Spectral-g,HAC,GMM,DPGMM},
x tick label style={font=\scriptsize},
x tick label style={rotate=90,anchor=east,nodes near coords align={horizontal}},
xtick={Geometric-AP,AP,kmedoids,kmeans,Spectral-g,HAC,GMM,DPGMM},
after end axis/.code={
\node(L1) at (axis cs:AP,1) {};
\node(L2) at (axis cs:HAC,1) {};
\node(XTL) at (xticklabel cs:0)  {};
\node[anchor=east](A1) at (axis cs:Geometric-AP,-14) {};
\node[anchor=west](B1) at (axis cs:kmedoids,-14) {};
\draw (A1|- XTL) -- (B1|- XTL);
\node[anchor=center,below] at (L1 |- XTL) {\scriptsize{Exemplar-Based}};
\node[anchor=east](A2) at (axis cs:kmeans,-14) {};
\node[anchor=west](B2) at (axis cs:DPGMM,-14) {};
\draw (A2|- XTL) -- (B2|- XTL);
\node[anchor=center,below] at (L2 |- XTL) {\scriptsize{Non Exemplar-Based}};
}
]
\addplot[red!80!black,fill=red!30!white,mark=none] coordinates {(Geometric-AP,75.75)};

\addplot[blue!80!black,fill=blue!30!white,mark=none] coordinates {(AP,77.43)};

\addplot[green!60!black,fill=green!30!white,mark=none] coordinates {(kmedoids,75.32)};

\addplot[xtick=\empty, draw=none] coordinates {(zero1,0)};

\addplot[orange!80!black,fill=orange!30!white,mark=none] coordinates {(kmeans,39.24)};

\addplot[cyan!80!black,fill=cyan!30!white,mark=none] coordinates {(Spectral-g,87.32)};

\addplot[violet!80!black,fill=violet!30!white,mark=none] coordinates {(HAC,25)};

\addplot[yellow!80!black,fill=yellow!50!white,mark=none] coordinates {(GMM,65.23)};

\addplot[black,fill=lightgray,mark=none] coordinates {(DPGMM,58.78)};
\end{axis}

\end{tikzpicture}
\caption{F1}
\label{fig:subfiguref4}
\end{subfigure}

\caption{Hyperparameter optimization and evaluation results on the \emph{\textbf{Zachary's Karate Club}} dataset using the modularity-based classes as ground-truth labels. (\textbf{a-c}) Distance metrics are evaluated with variable threshold values to predict the ground-truth categories, i.e. 4. By taking the NMI measure as reference, the optimal clustering results are obtained with the $\mathtt{Jaccard}$ distance and $\tau=0.8$.
(\textbf{d-f}) NMI, CR, and F1 evaluation metrics are reported for the different tested algorithms.}
\label{fig:Figure4}
\end{figure*}

%% file: figures/karate_FRFD.tex
\begin{figure*}[ht!]
\centering
\begin{subfigure}[b]{0.33\textwidth}
\begin{tikzpicture}
\begin{axis}[width=\linewidth,font=\footnotesize,
 ybar,bar shift=0,ymin=0,ymax=110,ytick={0,20,40,60,80,100},bar width=8pt,
ylabel={Percentage (\%)},
symbolic x coords={Geometric-AP,AP,kmedoids,zero1,kmeans,Spectral-g,HAC,GMM,DPGMM},
x tick label style={font=\scriptsize},
x tick label style={rotate=90,anchor=east,nodes near coords align={horizontal}},
xtick={Geometric-AP,AP,kmedoids,kmeans,Spectral-g,HAC,GMM,DPGMM},
after end axis/.code={
\node(L1) at (axis cs:AP,1) {};
\node(L2) at (axis cs:HAC,1) {};
\node(XTL) at (xticklabel cs:0)  {};
\node[anchor=east](A1) at (axis cs:Geometric-AP,-14) {};
\node[anchor=west](B1) at (axis cs:kmedoids,-14) {};
\draw (A1|- XTL) -- (B1|- XTL);
\node[anchor=center,below] at (L1 |- XTL) {\scriptsize{Exemplar-Based}};
\node[anchor=east](A2) at (axis cs:kmeans,-14) {};
\node[anchor=west](B2) at (axis cs:DPGMM,-14) {};
\draw (A2|- XTL) -- (B2|- XTL);
\node[anchor=center,below] at (L2 |- XTL) {\scriptsize{Non Exemplar-Based}};
}
]
\addplot[red!80!black,fill=red!30!white,mark=none,error bars/.cd, y dir=both, y explicit] coordinates {(Geometric-AP,81.88)+- (0,5.2)};

\addplot[blue!80!black,fill=blue!30!white,mark=none,error bars/.cd, y dir=both, y explicit] coordinates {(AP,65.78)+-(0,9.25)};

\addplot[green!60!black,fill=green!30!white,mark=none,error bars/.cd, y dir=both, y explicit] coordinates {(kmedoids,76.19)+-(0,8.62)};

\addplot[xtick=\empty, draw=none] coordinates {(zero1,0)};

\addplot[orange!80!black,fill=orange!30!white,mark=none,error bars/.cd, y dir=both, y explicit] coordinates {(kmeans,62.13)+-(0,6.29)};

\addplot[cyan!80!black,fill=cyan!30!white,mark=none,error bars/.cd, y dir=both, y explicit] coordinates {(Spectral-g,73.24)+-(0,0)};

\addplot[violet!80!black,fill=violet!30!white,mark=none,error bars/.cd, y dir=both, y explicit] coordinates {(HAC,65.23)+-(0,19.68)};

\addplot[yellow!80!black,fill=yellow!50!white,mark=none,error bars/.cd, y dir=both, y explicit] coordinates {(GMM,72.72)+-(0,13.43)};

\addplot[black,fill=lightgray,mark=none,error bars/.cd, y dir=both, y explicit] coordinates {(DPGMM,60.96)+-(0,22.4)};
\end{axis}

\end{tikzpicture}
\caption{NMI}
\label{fig:subfigurea5}
\end{subfigure}
\begin{subfigure}[b]{0.33\textwidth}
\begin{tikzpicture}
\begin{axis}[width=\linewidth,font=\footnotesize,
 ybar,bar shift=0,ymin=0,ymax=110,ytick={0,20,40,60,80,100},bar width=8pt,
ylabel={Percentage (\%)},
symbolic x coords={Geometric-AP,AP,kmedoids,zero1,kmeans,Spectral-g,HAC,GMM,DPGMM},
x tick label style={font=\scriptsize},
x tick label style={rotate=90,anchor=east,nodes near coords align={horizontal}},
xtick={Geometric-AP,AP,kmedoids,kmeans,Spectral-g,HAC,GMM,DPGMM},
after end axis/.code={
\node(L1) at (axis cs:AP,1) {};
\node(L2) at (axis cs:HAC,1) {};
\node(XTL) at (xticklabel cs:0)  {};
\node[anchor=east](A1) at (axis cs:Geometric-AP,-14) {};
\node[anchor=west](B1) at (axis cs:kmedoids,-14) {};
\draw (A1|- XTL) -- (B1|- XTL);
\node[anchor=center,below] at (L1 |- XTL) {\scriptsize{Exemplar-Based}};
\node[anchor=east](A2) at (axis cs:kmeans,-14) {};
\node[anchor=west](B2) at (axis cs:DPGMM,-14) {};
\draw (A2|- XTL) -- (B2|- XTL);
\node[anchor=center,below] at (L2 |- XTL) {\scriptsize{Non Exemplar-Base}};
}
]
\addplot[red!80!black,fill=red!30!white,mark=none,error bars/.cd, y dir=both, y explicit] coordinates {(Geometric-AP,96.59	)+-(0,1.29)};

\addplot[blue!80!black,fill=blue!30!white,mark=none,error bars/.cd, y dir=both, y explicit] coordinates {(AP,65.78)+-(0,9.25)};

\addplot[green!60!black,fill=green!30!white,mark=none,error bars/.cd, y dir=both, y explicit] coordinates {(kmedoids,94.75)+-(0,2.81)};

\addplot[xtick=\empty, draw=none] coordinates {(zero1,0)};

\addplot[orange!80!black,fill=orange!30!white,mark=none,error bars/.cd, y dir=both, y explicit] coordinates {(kmeans,89.93)+-(0,2.41)};

\addplot[cyan!80!black,fill=cyan!30!white,mark=none,error bars/.cd, y dir=both, y explicit] coordinates {(Spectral-g,94.12)+-(0,0)};

\addplot[violet!80!black,fill=violet!30!white,mark=none,error bars/.cd, y dir=both, y explicit] coordinates {(HAC,89.69)+-(0,7.75)};

\addplot[yellow!80!black,fill=yellow!50!white,mark=none,error bars/.cd, y dir=both, y explicit] coordinates {(GMM,93.26)+-(0,4.2)};

\addplot[black,fill=lightgray,mark=none,error bars/.cd, y dir=both, y explicit] coordinates {(DPGMM,88.48)+-(0,10.61)};
\end{axis}
\end{tikzpicture}
\caption{CR}
\label{fig:subfigureb5}
\end{subfigure}
\begin{subfigure}[b]{0.33\textwidth}
\begin{tikzpicture}
\begin{axis}[width=\linewidth,font=\footnotesize,
 ybar,bar shift=0,ymin=0,ymax=110,ytick={0,20,40,60,80,100},bar width=8pt,
ylabel={Percentage (\%)},
symbolic x coords={Geometric-AP,AP,kmedoids,zero1,kmeans,Spectral-g,HAC,GMM,DPGMM},
x tick label style={font=\scriptsize},
x tick label style={rotate=90,anchor=east,nodes near coords align={horizontal}},
xtick={Geometric-AP,AP,kmedoids,kmeans,Spectral-g,HAC,GMM,DPGMM},
after end axis/.code={
\node(L1) at (axis cs:AP,1) {};
\node(L2) at (axis cs:HAC,1) {};
\node(XTL) at (xticklabel cs:0)  {};
\node[anchor=east](A1) at (axis cs:Geometric-AP,-14) {};
\node[anchor=west](B1) at (axis cs:kmedoids,-14) {};
\draw (A1|- XTL) -- (B1|- XTL);
\node[anchor=center,below] at (L1 |- XTL) {\scriptsize{Exemplar-Based}};
\node[anchor=east](A2) at (axis cs:kmeans,-14) {};
\node[anchor=west](B2) at (axis cs:DPGMM,-14) {};
\draw (A2|- XTL) -- (B2|- XTL);
\node[anchor=center,below] at (L2 |- XTL) {\scriptsize{Non Exemplar-Based}};
}
]
\addplot[red!80!black,fill=red!30!white,mark=none,error bars/.cd, y dir=both, y explicit] coordinates {(Geometric-AP,96.58)+-(0,1.29)};

\addplot[blue!80!black,fill=blue!30!white,mark=none,error bars/.cd, y dir=both, y explicit] coordinates {(AP,91.09)+-(0,3.31)};

\addplot[green!60!black,fill=green!30!white,mark=none,error bars/.cd, y dir=both, y explicit] coordinates {(kmedoids,94.72)+-(0,2.86)};

\addplot[xtick=\empty, draw=none] coordinates {(zero1,0)};

\addplot[orange!80!black,fill=orange!30!white,mark=none,error bars/.cd, y dir=both, y explicit] coordinates {(kmeans,89.81)+-(0,2.48)};

\addplot[cyan!80!black,fill=cyan!30!white,mark=none,error bars/.cd, y dir=both, y explicit] coordinates {(Spectral-g,94.1)+-(0,0)};

\addplot[violet!80!black,fill=violet!30!white,mark=none,error bars/.cd, y dir=both, y explicit] coordinates {(HAC,89.34)+-(0,8.39)};

\addplot[yellow!80!black,fill=yellow!50!white,mark=none,error bars/.cd, y dir=both, y explicit] coordinates {(GMM,93.19)+-(0,4.28)};

\addplot[black,fill=lightgray,mark=none,error bars/.cd, y dir=both, y explicit] coordinates {(DPGMM,87.59)+-(0,13.31)};
\end{axis}

\end{tikzpicture}
\caption{F1}
\label{fig:subfigurec5}
\end{subfigure}
\par\bigskip 

\begin{subfigure}[b]{0.33\textwidth}
\begin{tikzpicture}
\begin{axis}[width=\linewidth,font=\footnotesize,
 ybar,bar shift=0,ymin=0,ymax=100,ytick={0,20,40,60,80,100},bar width=8pt,
ylabel={Percentage (\%)},
symbolic x coords={Geometric-AP,AP,kmedoids,zero1,kmeans,Spectral-g,HAC,GMM,DPGMM},
x tick label style={font=\scriptsize},
x tick label style={rotate=90,anchor=east,nodes near coords align={horizontal}},
xtick={Geometric-AP,AP,kmedoids,kmeans,Spectral-g,HAC,GMM,DPGMM},
after end axis/.code={
\node(L1) at (axis cs:AP,1) {};
\node(L2) at (axis cs:HAC,1) {};
\node(XTL) at (xticklabel cs:0)  {};
\node[anchor=east](A1) at (axis cs:Geometric-AP,-14) {};
\node[anchor=west](B1) at (axis cs:kmedoids,-14) {};
\draw (A1|- XTL) -- (B1|- XTL);
\node[anchor=center,below] at (L1 |- XTL) {\scriptsize{Exemplar-Based}};
\node[anchor=east](A2) at (axis cs:kmeans,-14) {};
\node[anchor=west](B2) at (axis cs:DPGMM,-14) {};
\draw (A2|- XTL) -- (B2|- XTL);
\node[anchor=center,below] at (L2 |- XTL) {\scriptsize{Non Exemplar-Based}};
}
]
\addplot[red!80!black,fill=red!30!white,mark=none,error bars/.cd, y dir=both, y explicit] coordinates {(Geometric-AP,73.71)+-(0,4.83)};

\addplot[blue!80!black,fill=blue!30!white,mark=none,error bars/.cd, y dir=both, y explicit] coordinates {(AP,73.63)+-(0,7.22)};

\addplot[green!60!black,fill=green!30!white,mark=none,error bars/.cd, y dir=both, y explicit] coordinates {(kmedoids,60.89)+-(0,9.7)};

\addplot[xtick=\empty, draw=none] coordinates {(zero1,0)};

\addplot[orange!80!black,fill=orange!30!white,mark=none,error bars/.cd, y dir=both, y explicit] coordinates {(kmeans,63.89)+-(0,6)};

\addplot[cyan!80!black,fill=cyan!30!white,mark=none,error bars/.cd, y dir=both, y explicit] coordinates {(Spectral-g,76.01)+-(0,0)};

\addplot[violet!80!black,fill=violet!30!white,mark=none,error bars/.cd, y dir=both, y explicit] coordinates {(HAC,70.1)+-(0,7.67)};

\addplot[yellow!80!black,fill=yellow!50!white,mark=none,error bars/.cd, y dir=both, y explicit] coordinates {(GMM,64.35)+-(0,9.73)};

\addplot[black,fill=lightgray,mark=none,error bars/.cd, y dir=both, y explicit] coordinates {(DPGMM,30.64)+-(0,10.49)};
\end{axis}

\end{tikzpicture}
\caption{NMI}
\label{fig:subfigured5}
\end{subfigure}
\begin{subfigure}[b]{0.33\textwidth}
\begin{tikzpicture}
\begin{axis}[width=\linewidth,font=\footnotesize,
 ybar,bar shift=0,ymin=0,ymax=100,ytick={0,20,40,60,80,100},bar width=8pt,
ylabel={Percentage (\%)},
symbolic x coords={Geometric-AP,AP,kmedoids,zero1,kmeans,Spectral-g,HAC,GMM,DPGMM},
x tick label style={font=\scriptsize},
x tick label style={rotate=90,anchor=east,nodes near coords align={horizontal}},
xtick={Geometric-AP,AP,kmedoids,kmeans,Spectral-g,HAC,GMM,DPGMM},
after end axis/.code={
\node(L1) at (axis cs:AP,1) {};
\node(L2) at (axis cs:HAC,1) {};
\node(XTL) at (xticklabel cs:0)  {};
\node[anchor=east](A1) at (axis cs:Geometric-AP,-14) {};
\node[anchor=west](B1) at (axis cs:kmedoids,-14) {};
\draw (A1|- XTL) -- (B1|- XTL);
\node[anchor=center,below] at (L1 |- XTL) {\scriptsize{Exemplar-Based}};
\node[anchor=east](A2) at (axis cs:kmeans,-14) {};
\node[anchor=west](B2) at (axis cs:DPGMM,-14) {};
\draw (A2|- XTL) -- (B2|- XTL);
\node[anchor=center,below] at (L2 |- XTL) {\scriptsize{Non Exemplar-Based}};
}
]
\addplot[red!80!black,fill=red!30!white,mark=none,error bars/.cd, y dir=both, y explicit] coordinates {(Geometric-AP,85.66	)+-(0,3.97)};

\addplot[blue!80!black,fill=blue!30!white,mark=none,error bars/.cd, y dir=both, y explicit] coordinates {(AP,73.63)+-(0,7.22)};

\addplot[green!60!black,fill=green!30!white,mark=none,error bars/.cd, y dir=both, y explicit] coordinates {(kmedoids,74.17)+-(0,8.76)};

\addplot[xtick=\empty, draw=none] coordinates {(zero1,0)};

\addplot[orange!80!black,fill=orange!30!white,mark=none,error bars/.cd, y dir=both, y explicit] coordinates {(kmeans,80.43)+-(0,5.12)};

\addplot[cyan!80!black,fill=cyan!30!white,mark=none,error bars/.cd, y dir=both, y explicit] coordinates {(Spectral-g,88.24)+-(0,0)};

\addplot[violet!80!black,fill=violet!30!white,mark=none,error bars/.cd, y dir=both, y explicit] coordinates {(HAC,82.57)+-(0,6.46)};

\addplot[yellow!80!black,fill=yellow!50!white,mark=none,error bars/.cd, y dir=both, y explicit] coordinates {(GMM,78.05)+-(0,8.27)};

\addplot[black,fill=lightgray,mark=none,error bars/.cd, y dir=both, y explicit] coordinates {(DPGMM,49.93)+-(0,14.06)};
\end{axis}
\end{tikzpicture}
\caption{CR}
\label{fig:subfiguree5}
\end{subfigure}
\begin{subfigure}[b]{0.33\textwidth}
\begin{tikzpicture}
\begin{axis}[width=\linewidth,font=\footnotesize,
 ybar,bar shift=0,ymin=0,ymax=100,ytick={0,20,40,60,80,100},bar width=8pt,
ylabel={Percentage (\%)},
symbolic x coords={Geometric-AP,AP,kmedoids,zero1,kmeans,Spectral-g,HAC,GMM,DPGMM},
x tick label style={font=\scriptsize},
x tick label style={rotate=90,anchor=east,nodes near coords align={horizontal}},
xtick={Geometric-AP,AP,kmedoids,kmeans,Spectral-g,HAC,GMM,DPGMM},
after end axis/.code={
\node(L1) at (axis cs:AP,1) {};
\node(L2) at (axis cs:HAC,1) {};
\node(XTL) at (xticklabel cs:0)  {};
\node[anchor=east](A1) at (axis cs:Geometric-AP,-14) {};
\node[anchor=west](B1) at (axis cs:kmedoids,-14) {};
\draw (A1|- XTL) -- (B1|- XTL);
\node[anchor=center,below] at (L1 |- XTL) {\scriptsize{Exemplar-Based}};
\node[anchor=east](A2) at (axis cs:kmeans,-14) {};
\node[anchor=west](B2) at (axis cs:DPGMM,-14) {};
\draw (A2|- XTL) -- (B2|- XTL);
\node[anchor=center,below] at (L2 |- XTL) {\scriptsize{Non Exemplar-Based}};
}
]
\addplot[red!80!black,fill=red!30!white,mark=none,error bars/.cd, y dir=both, y explicit] coordinates {(Geometric-AP,82.89)+-(0,6.79)};

\addplot[blue!80!black,fill=blue!30!white,mark=none,error bars/.cd, y dir=both, y explicit] coordinates {(AP,85.51)+-(0,4.37)};

\addplot[green!60!black,fill=green!30!white,mark=none,error bars/.cd, y dir=both, y explicit] coordinates {(kmedoids,68.83)+-(0,14.1)};

\addplot[xtick=\empty, draw=none] coordinates {(zero1,0)};

\addplot[orange!80!black,fill=orange!30!white,mark=none,error bars/.cd, y dir=both, y explicit] coordinates {(kmeans,80.5)+-(0,5.27)};

\addplot[cyan!80!black,fill=cyan!30!white,mark=none,error bars/.cd, y dir=both, y explicit] coordinates {(Spectral-g,87.32)+-(0,0)};

\addplot[violet!80!black,fill=violet!30!white,mark=none,error bars/.cd, y dir=both, y explicit] coordinates {(HAC,82.39)+-(0,8.5)};

\addplot[yellow!80!black,fill=yellow!50!white,mark=none,error bars/.cd, y dir=both, y explicit] coordinates {(GMM,75.58)+-(0,11.91)};

\addplot[black,fill=lightgray,mark=none,error bars/.cd, y dir=both, y explicit] coordinates {(DPGMM,27.27)+-(0,13.79)};
\end{axis}

\end{tikzpicture}
\caption{F1}
\label{fig:subfiguref5}
\end{subfigure}

\caption{Average clustering results using 1000 sets of FRFD node embeddings. Error bars show the standard deviations. (\textbf{a-c}) Results when using the club split classes. (\textbf{d-f}) Results when using the modularity-based classes.}
\label{fig:Figure5}
\end{figure*}

%% file: figures/karate_tsne.tex
\begin{figure*}[ht!]
\centering
\begin{subfigure}[b]{0.33\textwidth}
\begin{tikzpicture}
\begin{axis}[width=\linewidth,font=\footnotesize,
 ybar,bar shift=0,ymin=0,ymax=110,ytick={0,20,40,60,80,100},bar width=8pt,
ylabel={Percentage (\%)},
symbolic x coords={Geometric-AP,AP,kmedoids,zero1,kmeans,Spectral-g,HAC,GMM,DPGMM},
x tick label style={font=\scriptsize},
x tick label style={rotate=90,anchor=east,nodes near coords align={horizontal}},
xtick={Geometric-AP,AP,kmedoids,kmeans,Spectral-g,HAC,GMM,DPGMM},
after end axis/.code={
\node(L1) at (axis cs:AP,1) {};
\node(L2) at (axis cs:HAC,1) {};
\node(XTL) at (xticklabel cs:0)  {};
\node[anchor=east](A1) at (axis cs:Geometric-AP,-14) {};
\node[anchor=west](B1) at (axis cs:kmedoids,-14) {};
\draw (A1|- XTL) -- (B1|- XTL);
\node[anchor=center,below] at (L1 |- XTL) {\scriptsize{Exemplar-Based}};
\node[anchor=east](A2) at (axis cs:kmeans,-14) {};
\node[anchor=west](B2) at (axis cs:DPGMM,-14) {};
\draw (A2|- XTL) -- (B2|- XTL);
\node[anchor=center,below] at (L2 |- XTL) {\scriptsize{Non Exemplar-Based}};
}
]
\addplot[red!80!black,fill=red!30!white,mark=none,error bars/.cd, y dir=both, y explicit] coordinates {(Geometric-AP,73.59)+- (0,22.47)};

\addplot[blue!80!black,fill=blue!30!white,mark=none,error bars/.cd, y dir=both, y explicit] coordinates {(AP,43.25)+-(0,17.51)};

\addplot[green!60!black,fill=green!30!white,mark=none,error bars/.cd, y dir=both, y explicit] coordinates {(kmedoids,28.23)+-(0,22.6)};

\addplot[xtick=\empty, draw=none] coordinates {(zero1,0)};

\addplot[orange!80!black,fill=orange!30!white,mark=none,error bars/.cd, y dir=both, y explicit] coordinates {(kmeans,45.64)+-(0,18.16)};

\addplot[cyan!80!black,fill=cyan!30!white,mark=none,error bars/.cd, y dir=both, y explicit] coordinates {(Spectral-g,73.24)+-(0,0)};

\addplot[violet!80!black,fill=violet!30!white,mark=none,error bars/.cd, y dir=both, y explicit] coordinates {(HAC,33.4 )+-(0,21.8)};

\addplot[yellow!80!black,fill=yellow!50!white,mark=none,error bars/.cd, y dir=both, y explicit] coordinates {(GMM,32.85)+-(0,23.87)};

\addplot[black,fill=lightgray,mark=none,error bars/.cd, y dir=both, y explicit] coordinates {(DPGMM,9.3)+-(0,0)};
\end{axis}

\end{tikzpicture}
\caption{NMI}
\label{fig:subfigurea6}
\end{subfigure}
\begin{subfigure}[b]{0.33\textwidth}
\begin{tikzpicture}
\begin{axis}[width=\linewidth,font=\footnotesize,
 ybar,bar shift=0,ymin=0,ymax=110,ytick={0,20,40,60,80,100},bar width=8pt,
ylabel={Percentage (\%)},
symbolic x coords={Geometric-AP,AP,kmedoids,zero1,kmeans,Spectral-g,HAC,GMM,DPGMM},
x tick label style={font=\scriptsize},
x tick label style={rotate=90,anchor=east,nodes near coords align={horizontal}},
xtick={Geometric-AP,AP,kmedoids,kmeans,Spectral-g,HAC,GMM,DPGMM},
after end axis/.code={
\node(L1) at (axis cs:AP,1) {};
\node(L2) at (axis cs:HAC,1) {};
\node(XTL) at (xticklabel cs:0)  {};
\node[anchor=east](A1) at (axis cs:Geometric-AP,-14) {};
\node[anchor=west](B1) at (axis cs:kmedoids,-14) {};
\draw (A1|- XTL) -- (B1|- XTL);
\node[anchor=center,below] at (L1 |- XTL) {\scriptsize{Exemplar-Based}};
\node[anchor=east](A2) at (axis cs:kmeans,-14) {};
\node[anchor=west](B2) at (axis cs:DPGMM,-14) {};
\draw (A2|- XTL) -- (B2|- XTL);
\node[anchor=center,below] at (L2 |- XTL) {\scriptsize{Non Exemplar-Based}};
}
]
\addplot[red!80!black,fill=red!30!white,mark=none,error bars/.cd, y dir=both, y explicit] coordinates {(Geometric-AP,92.94)+-(0,9.32)};

\addplot[blue!80!black,fill=blue!30!white,mark=none,error bars/.cd, y dir=both, y explicit] coordinates {(AP,43.25)+-(0,17.51)};

\addplot[green!60!black,fill=green!30!white,mark=none,error bars/.cd, y dir=both, y explicit] coordinates {(kmedoids,76.48)+-(0,12.25)};

\addplot[xtick=\empty, draw=none] coordinates {(zero1,0)};

\addplot[orange!80!black,fill=orange!30!white,mark=none,error bars/.cd, y dir=both, y explicit] coordinates {(kmeans,86.24)+-(0,6.7)};

\addplot[cyan!80!black,fill=cyan!30!white,mark=none,error bars/.cd, y dir=both, y explicit] coordinates {(Spectral-g,94.12)+-(0,0)};

\addplot[violet!80!black,fill=violet!30!white,mark=none,error bars/.cd, y dir=both, y explicit] coordinates {(HAC,77.93)+-(0,10.97)};

\addplot[yellow!80!black,fill=yellow!50!white,mark=none,error bars/.cd, y dir=both, y explicit] coordinates {(GMM,78.24)+-(0,13.2)};

\addplot[black,fill=lightgray,mark=none,error bars/.cd, y dir=both, y explicit] coordinates {(DPGMM,55.89)+-(0,0)};
\end{axis}
\end{tikzpicture}
\caption{CR}
\label{fig:subfigureb6}
\end{subfigure}
\begin{subfigure}[b]{0.33\textwidth}
\begin{tikzpicture}
\begin{axis}[width=\linewidth,font=\footnotesize,
 ybar,bar shift=0,ymin=0,ymax=110,ytick={0,20,40,60,80,100},bar width=8pt,
ylabel={Percentage (\%)},
symbolic x coords={Geometric-AP,AP,kmedoids,zero1,kmeans,Spectral-g,HAC,GMM,DPGMM},
x tick label style={font=\scriptsize},
x tick label style={rotate=90,anchor=east,nodes near coords align={horizontal}},
xtick={Geometric-AP,AP,kmedoids,kmeans,Spectral-g,HAC,GMM,DPGMM},
after end axis/.code={
\node(L1) at (axis cs:AP,1) {};
\node(L2) at (axis cs:HAC,1) {};
\node(XTL) at (xticklabel cs:0)  {};
\node[anchor=east](A1) at (axis cs:Geometric-AP,-14) {};
\node[anchor=west](B1) at (axis cs:kmedoids,-14) {};
\draw (A1|- XTL) -- (B1|- XTL);
\node[anchor=center,below] at (L1 |- XTL) {\scriptsize{Exemplar-Based}};
\node[anchor=east](A2) at (axis cs:kmeans,-14) {};
\node[anchor=west](B2) at (axis cs:DPGMM,-14) {};
\draw (A2|- XTL) -- (B2|- XTL);
\node[anchor=center,below] at (L2 |- XTL) {\scriptsize{Non Exemplar-Based}};
}
]
\addplot[red!80!black,fill=red!30!white,mark=none,error bars/.cd, y dir=both, y explicit] coordinates {(Geometric-AP,92.5 )+-(0,10.9)};

\addplot[blue!80!black,fill=blue!30!white,mark=none,error bars/.cd, y dir=both, y explicit] coordinates {(AP,85.3)+-(0,6.48)};

\addplot[green!60!black,fill=green!30!white,mark=none,error bars/.cd, y dir=both, y explicit] coordinates {(kmedoids,76.29)+-(0,12.66)};

\addplot[xtick=\empty, draw=none] coordinates {(zero1,0)};

\addplot[orange!80!black,fill=orange!30!white,mark=none,error bars/.cd, y dir=both, y explicit] coordinates {(kmeans,86.22)+-(0,6.72)};

\addplot[cyan!80!black,fill=cyan!30!white,mark=none,error bars/.cd, y dir=both, y explicit] coordinates {(Spectral-g,94.1)+-(0,0)};

\addplot[violet!80!black,fill=violet!30!white,mark=none,error bars/.cd, y dir=both, y explicit] coordinates {(HAC,77.08)+-(0,12.11)};

\addplot[yellow!80!black,fill=yellow!50!white,mark=none,error bars/.cd, y dir=both, y explicit] coordinates {(GMM,78.14)+-(0,13.29)};

\addplot[black,fill=lightgray,mark=none,error bars/.cd, y dir=both, y explicit] coordinates {(DPGMM,45.23)+-(0,0)};
\end{axis}

\end{tikzpicture}
\caption{F1}
\label{fig:subfigurec6}
\end{subfigure}
\par\bigskip 

\begin{subfigure}[b]{0.33\textwidth}
\begin{tikzpicture}
\begin{axis}[width=\linewidth,font=\footnotesize,
 ybar,bar shift=0,ymin=0,ymax=100,ytick={0,20,40,60,80,100},bar width=8pt,
ylabel={Percentage (\%)},
symbolic x coords={Geometric-AP,AP,kmedoids,zero1,kmeans,Spectral-g,HAC,GMM,DPGMM},
x tick label style={font=\scriptsize},
x tick label style={rotate=90,anchor=east,nodes near coords align={horizontal}},
xtick={Geometric-AP,AP,kmedoids,kmeans,Spectral-g,HAC,GMM,DPGMM},
after end axis/.code={
\node(L1) at (axis cs:AP,1) {};
\node(L2) at (axis cs:HAC,1) {};
\node(XTL) at (xticklabel cs:0)  {};
\node[anchor=east](A1) at (axis cs:Geometric-AP,-14) {};
\node[anchor=west](B1) at (axis cs:kmedoids,-14) {};
\draw (A1|- XTL) -- (B1|- XTL);
\node[anchor=center,below] at (L1 |- XTL) {\scriptsize{Exemplar-Based}};
\node[anchor=east](A2) at (axis cs:kmeans,-14) {};
\node[anchor=west](B2) at (axis cs:DPGMM,-14) {};
\draw (A2|- XTL) -- (B2|- XTL);
\node[anchor=center,below] at (L2 |- XTL) {\scriptsize{Non Exemplar-Based}};
}
]
\addplot[red!80!black,fill=red!30!white,mark=none,error bars/.cd, y dir=both, y explicit] coordinates {(Geometric-AP,58.31)+-(0,11.98)};

\addplot[blue!80!black,fill=blue!30!white,mark=none,error bars/.cd, y dir=both, y explicit] coordinates {(AP,28.85)+-(0,10.34)};

\addplot[green!60!black,fill=green!30!white,mark=none,error bars/.cd, y dir=both, y explicit] coordinates {(kmedoids,27.37)+-(0,10.24)};

\addplot[xtick=\empty, draw=none] coordinates {(zero1,0)};

\addplot[orange!80!black,fill=orange!30!white,mark=none,error bars/.cd, y dir=both, y explicit] coordinates {(kmeans,27.97)+-(0,10.47)};

\addplot[cyan!80!black,fill=cyan!30!white,mark=none,error bars/.cd, y dir=both, y explicit] coordinates {(Spectral-g,76.01)+-(0,0)};

\addplot[violet!80!black,fill=violet!30!white,mark=none,error bars/.cd, y dir=both, y explicit] coordinates {(HAC,29.6)+-(0,10.66)};

\addplot[yellow!80!black,fill=yellow!50!white,mark=none,error bars/.cd, y dir=both, y explicit] coordinates {(GMM,26.49)+-(0,9.54)};

\addplot[black,fill=lightgray,mark=none,error bars/.cd, y dir=both, y explicit] coordinates {(DPGMM,8.32)+-(0,1.58)};
\end{axis}

\end{tikzpicture}
\caption{NMI}
\label{fig:subfigured6}
\end{subfigure}
\begin{subfigure}[b]{0.33\textwidth}
\begin{tikzpicture}
\begin{axis}[width=\linewidth,font=\footnotesize,
 ybar,bar shift=0,ymin=0,ymax=100,ytick={0,20,40,60,80,100},bar width=8pt,
ylabel={Percentage (\%)},
symbolic x coords={Geometric-AP,AP,kmedoids,zero1,kmeans,Spectral-g,HAC,GMM,DPGMM},
x tick label style={font=\scriptsize},
x tick label style={rotate=90,anchor=east,nodes near coords align={horizontal}},
xtick={Geometric-AP,AP,kmedoids,kmeans,Spectral-g,HAC,GMM,DPGMM},
after end axis/.code={
\node(L1) at (axis cs:AP,1) {};
\node(L2) at (axis cs:HAC,1) {};
\node(XTL) at (xticklabel cs:0)  {};
\node[anchor=east](A1) at (axis cs:Geometric-AP,-14) {};
\node[anchor=west](B1) at (axis cs:kmedoids,-14) {};
\draw (A1|- XTL) -- (B1|- XTL);
\node[anchor=center,below] at (L1 |- XTL) {\scriptsize{Exemplar-Based}};
\node[anchor=east](A2) at (axis cs:kmeans,-14) {};
\node[anchor=west](B2) at (axis cs:DPGMM,-14) {};
\draw (A2|- XTL) -- (B2|- XTL);
\node[anchor=center,below] at (L2 |- XTL) {\scriptsize{Non Exemplar-Based}};
}
]
\addplot[red!80!black,fill=red!30!white,mark=none,error bars/.cd, y dir=both, y explicit] coordinates {(Geometric-AP,71.82	)+-(0,8.89)};

\addplot[blue!80!black,fill=blue!30!white,mark=none,error bars/.cd, y dir=both, y explicit] coordinates {(AP,28.85)+-(0,10.34)};

\addplot[green!60!black,fill=green!30!white,mark=none,error bars/.cd, y dir=both, y explicit] coordinates {(kmedoids,54.89)+-(0,7.14)};

\addplot[xtick=\empty, draw=none] coordinates {(zero1,0)};

\addplot[orange!80!black,fill=orange!30!white,mark=none,error bars/.cd, y dir=both, y explicit] coordinates {(kmeans,55.67)+-(0,6.89)};

\addplot[cyan!80!black,fill=cyan!30!white,mark=none,error bars/.cd, y dir=both, y explicit] coordinates {(Spectral-g,88.24)+-(0,0)};

\addplot[violet!80!black,fill=violet!30!white,mark=none,error bars/.cd, y dir=both, y explicit] coordinates {(HAC,55.95)+-(0,7.47)};

\addplot[yellow!80!black,fill=yellow!50!white,mark=none,error bars/.cd, y dir=both, y explicit] coordinates {(GMM,54.4 )+-(0,6.87)};

\addplot[black,fill=lightgray,mark=none,error bars/.cd, y dir=both, y explicit] coordinates {(DPGMM,35.3)+-(0,0)};
\end{axis}
\end{tikzpicture}
\caption{CR}
\label{fig:subfiguree6}
\end{subfigure}
\begin{subfigure}[b]{0.33\textwidth}
\begin{tikzpicture}
\begin{axis}[width=\linewidth,font=\footnotesize,
 ybar,bar shift=0,ymin=0,ymax=100,ytick={0,20,40,60,80,100},bar width=8pt,
ylabel={Percentage (\%)},
symbolic x coords={Geometric-AP,AP,kmedoids,zero1,kmeans,Spectral-g,HAC,GMM,DPGMM},
x tick label style={font=\scriptsize},
x tick label style={rotate=90,anchor=east,nodes near coords align={horizontal}},
xtick={Geometric-AP,AP,kmedoids,kmeans,Spectral-g,HAC,GMM,DPGMM},
after end axis/.code={
\node(L1) at (axis cs:AP,1) {};
\node(L2) at (axis cs:HAC,1) {};
\node(XTL) at (xticklabel cs:0)  {};
\node[anchor=east](A1) at (axis cs:Geometric-AP,-14) {};
\node[anchor=west](B1) at (axis cs:kmedoids,-14) {};
\draw (A1|- XTL) -- (B1|- XTL);
\node[anchor=center,below] at (L1 |- XTL) {\scriptsize{Exemplar-Based}};
\node[anchor=east](A2) at (axis cs:kmeans,-14) {};
\node[anchor=west](B2) at (axis cs:DPGMM,-14) {};
\draw (A2|- XTL) -- (B2|- XTL);
\node[anchor=center,below] at (L2 |- XTL) {\scriptsize{Non Exemplar-Based}};
}
]
\addplot[red!80!black,fill=red!30!white,mark=none,error bars/.cd, y dir=both, y explicit] coordinates {(Geometric-AP,61.46)+-(0,13.38)};

\addplot[blue!80!black,fill=blue!30!white,mark=none,error bars/.cd, y dir=both, y explicit] coordinates {(AP,44.21)+-(0,9.39)};

\addplot[green!60!black,fill=green!30!white,mark=none,error bars/.cd, y dir=both, y explicit] coordinates {(kmedoids,42.33)+-(0,9.26)};

\addplot[xtick=\empty, draw=none] coordinates {(zero1,0)};

\addplot[orange!80!black,fill=orange!30!white,mark=none,error bars/.cd, y dir=both, y explicit] coordinates {(kmeans,43.79)+-(0,9.46)};

\addplot[cyan!80!black,fill=cyan!30!white,mark=none,error bars/.cd, y dir=both, y explicit] coordinates {(Spectral-g,87.32)+-(0,0)};

\addplot[violet!80!black,fill=violet!30!white,mark=none,error bars/.cd, y dir=both, y explicit] coordinates {(HAC,44.51)+-(0,9.89)};

\addplot[yellow!80!black,fill=yellow!50!white,mark=none,error bars/.cd, y dir=both, y explicit] coordinates {(GMM,42.15)+-(0,8.82)};

\addplot[black,fill=lightgray,mark=none,error bars/.cd, y dir=both, y explicit] coordinates {(DPGMM,13.05)+-(0,0)};
\end{axis}

\end{tikzpicture}
\caption{F1}
\label{fig:subfiguref6}
\end{subfigure}

\caption{Average clustering results using 1000 sets of t-SNE node embeddings. Error bars show the standard deviations. (\textbf{a-c}) Results when using the club split classes. (\textbf{d-f}) Results when using the modularity-based classes.}
\label{fig:Figure6}
\end{figure*}

%% file: figures/cora_random_network.tex
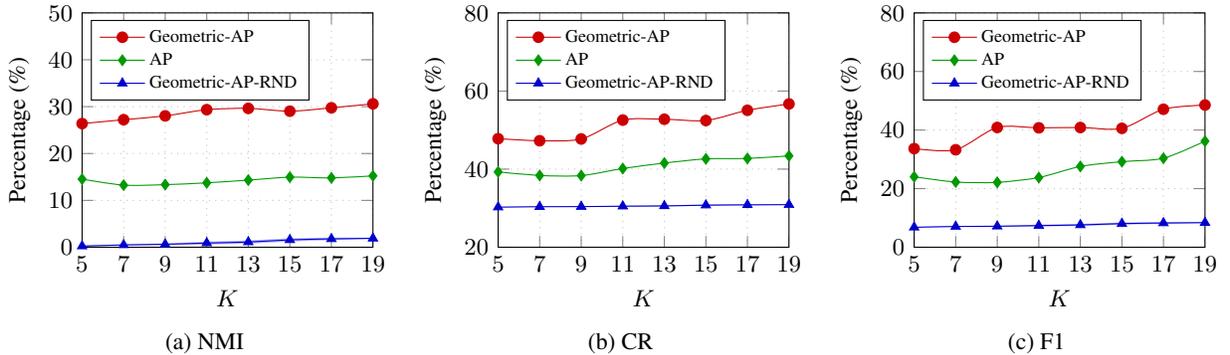
\begin{figure*}[ht!]
\begin{subfigure}[b]{0.33\textwidth}
\begin{tikzpicture}
    \begin{axis}[width=\linewidth,font=\footnotesize,
    xlabel= $K$, ylabel= Percentage (\%),
  ymin=0,ymax=50,xmin=5,xmax=19,
 legend pos= north west,legend style={nodes=right},legend style={nodes={scale=0.75, transform shape}},
      xmajorgrids,
    grid style={dotted},
    ymajorgrids,
    xtick={5,7,9,11,13,15,17,19},
    ytick={0,10,20,30,40,50},
    ]
\addplot[smooth,red!80!black,mark=*,mark options={solid}] plot coordinates {
(5,26.38)(7,27.21)(9,28.03)(11,29.35)(13,29.63)(15,29.03)(17,29.75)(19,30.60)
};
     \addplot[smooth,green!60!black,mark=diamond*,mark options={solid}] plot coordinates {
(5,14.53)(7,13.25)(9,13.35)(11,13.74)(13,14.32)(15,14.95)(17,14.79)(19,15.23)
};
     \addplot[smooth,blue!80!black,mark=triangle*,mark options={solid}] plot coordinates {
(5,0.24)(7,0.49)(9,0.63)(11,0.89)(13,1.11)(15,1.57)(17,1.78)(19,1.90)
};

    \addplot[name path=A,draw=none] plot coordinates {
(5,0.12)(7,0.37)(9,0.46)(11,0.59)(13,0.76)(15,1.23)(17,1.50)(19,1.77)
};

 \addplot[name path=B,draw=none] plot coordinates {
(5,0.36)(7,0.61)(9,0.80)(11,1.19)(13,1.46)(15,1.91)(17,2.06)(19,2.03)
};

\addplot[blue!30!white] fill between[of=A and B];

\legend{Geometric-AP,AP,Geometric-AP-RND};
\end{axis}
\end{tikzpicture}
\caption{NMI}
\label{fig:subfigurea7}
\end{subfigure}
\begin{subfigure}[b]{0.33\textwidth}
\begin{tikzpicture}
    \begin{axis}[width=\linewidth,font=\footnotesize,
    xlabel= $K$, ylabel= Percentage (\%),
  ymin=20,ymax=80,xmin=5,xmax=19,legend style={nodes=right},legend pos= north west,legend style={nodes={scale=0.75, transform shape}},
      xmajorgrids,
    grid style={dotted},
    ymajorgrids,
    xtick={5,7,9,11,13,15,17,19},
    ]
\addplot[smooth,red!80!black,mark=*,mark options={solid}] plot coordinates {
(5,47.79)(7,47.27)(9,47.72)(11,52.55)(13,52.77)(15,52.44)(17,55.06)(19,56.69)
};
     \addplot[smooth,green!60!black,mark=diamond*,mark options={solid}] plot coordinates {
(5,39.30)(7,38.37)(9,38.34)(11,40.15)(13,41.55)(15,42.62)(17,42.73)(19,43.39)
};
     \addplot[smooth,blue!80!black,mark=triangle*,mark options={solid}] plot coordinates {
(5,30.26)(7,30.37)(9,30.40)(11,30.50)(13,30.58)(15,30.77)(17,30.85)(19,30.90)
};

    \addplot[name path=A,draw=none] plot coordinates {
(5,30.21)(7,30.32)(9,30.33)(11,30.37)(13,30.43)(15,30.61)(17,30.73)(19,30.83)
};

 \addplot[name path=B,draw=none] plot coordinates {
(5,30.31)(7,30.42)(9,30.47)(11,30.63)(13,30.73)(15,30.93)(17,30.97)(19,30.97)
};

\addplot[blue!30!white] fill between[of=A and B];

\legend{Geometric-AP,AP,Geometric-AP-RND};
\end{axis}
\end{tikzpicture}
\caption{CR}
\label{fig:subfigureb7}
\end{subfigure}
\begin{subfigure}[b]{0.33\textwidth}
\begin{tikzpicture}
    \begin{axis}[width=\linewidth,font=\footnotesize,
    xlabel= $K$, ylabel= Percentage (\%),
  ymin=0,ymax=80,xmin=5,xmax=19,legend style={nodes=right},legend pos= north west,legend style={nodes={scale=0.75, transform shape}},
      xmajorgrids,
    grid style={dotted},
    ymajorgrids,
    xtick={5,7,9,11,13,15,17,19},
    ]
\addplot[smooth,red!80!black,mark=*,mark options={solid}] plot coordinates {
(5,33.67)(7,33.27)(9,40.89)(11,40.77)(13,40.87)(15,40.59)(17,47.11)(19,48.55)
};
     \addplot[smooth,green!60!black,mark=diamond*,mark options={solid}] plot coordinates {
(5,24.05)(7,22.26)(9,22.15)(11,23.80)(13,27.58)(15,29.21)(17,30.39)(19,36.21)
};
     \addplot[smooth,blue!80!black,mark=triangle*,mark options={solid}] plot coordinates {
(5,6.81)(7,7.08)(9,7.16)(11,7.39)(13,7.61)(15,8.07)(17,8.27)(19,8.39)
};

    \addplot[name path=A,draw=none] plot coordinates {
(5,6.68)(7,6.96)(9,6.98)(11,7.08)(13,7.24)(15,7.69)(17,7.96)(19,8.22)
};

 \addplot[name path=B,draw=none] plot coordinates {
(5,6.94)(7,7.20)(9,7.34)(11,7.70)(13,7.98)(15,8.45)(17,8.58)(19,8.56)
};

\addplot[blue!30!white] fill between[of=A and B];

\legend{Geometric-AP,AP,Geometric-AP-RND};
\end{axis}
\end{tikzpicture}
\caption{F1}
\label{fig:subfigurec7}
 \end{subfigure}
\caption{Clustering performance metrics as a function of the number of clusters learned on the Cora dataset. Evaluation measures for Geometric-AP-RND are averaged over 100 repetitions with randomly permuted networks. Standard deviations found to be too small and thus invisible w.r.t the scale of the plots.}
\label{fig:Figure7}
\end{figure*}

%% file: figures/citeseer_random_network.tex
\begin{figure*}[ht!]
\begin{subfigure}[b]{0.33\textwidth}
\begin{tikzpicture}
    \begin{axis}[width=\linewidth,font=\footnotesize,
    xlabel= $K$, ylabel= Percentage (\%),
  ymin=0,ymax=30,xmin=4,xmax=18,
 legend pos= north west,legend style={nodes=right},legend style={nodes={scale=0.75, transform shape}},
      xmajorgrids,
    grid style={dotted},
    ymajorgrids,
    xtick={4,6,8,10,12,14,16,18},
    ]
\addplot[smooth,red!80!black,mark=*,mark options={solid}] plot coordinates {
(4,15.28)(6,18.57)(8,16.49)(10,17.73)(12,17.36)(14,16.80)(16,17.13)(18,17.87)
};
     \addplot[smooth,green!60!black,mark=diamond*,mark options={solid}] plot coordinates {
(4,6.18)(6,8.02)(8,11.62)(10,9.72)(12,11.74)(14,11.09)(16,13.19)(18,14.04)
};
     \addplot[smooth,blue!80!black,mark=triangle*,mark options={solid}] plot coordinates {
(4,0.42)(6,0.48)(8,0.69)(10,0.89)(12,0.93)(14,1.02)(16,1.21)(18,1.54)
};

    \addplot[name path=A,draw=none] plot coordinates {
(4,0.17)(6,0.18)(8,0.41)(10,0.56)(12,0.60)(14,0.70)(16,0.83)(18,1.16)
};

 \addplot[name path=B,draw=none] plot coordinates {
(4,0.67)(6,0.78)(8,0.97)(10,1.22)(12,1.26)(14,1.34)(16,1.59)(18,1.92)
};

\addplot[blue!30!white] fill between[of=A and B];

\legend{Geometric-AP,AP,Geometric-AP-RND};
\end{axis}
\end{tikzpicture}
\caption{NMI}
\label{fig:subfigurea8}
\end{subfigure}
\begin{subfigure}[b]{0.33\textwidth}
\begin{tikzpicture}
    \begin{axis}[width=\linewidth,font=\footnotesize,
    xlabel= $K$, ylabel= Percentage (\%),
  ymin=20,ymax=80,xmin=4,xmax=18,legend style={nodes=right},legend pos= north west,legend style={nodes={scale=0.75, transform shape}},
      xmajorgrids,
    grid style={dotted},
    ymajorgrids,
 xtick={4,6,8,10,12,14,16,18},
    ]
\addplot[smooth,red!80!black,mark=*,mark options={solid}] plot coordinates {
(4,42.00)(6,45.57)(8,44.30)(10,46.71)(12,47.17)(14,46.26)(16,47.05)(18,48.47)
};
     \addplot[smooth,green!60!black,mark=diamond*,mark options={solid}] plot coordinates {
(4,35.51)(6,37.05)(8,43.27)(10,41.58)(12,44.03)(14,44.18)(16,46.65)(18,47.59)
};
     \addplot[smooth,blue!80!black,mark=triangle*,mark options={solid}] plot coordinates {
(4,21.41)(6,21.45)(8,21.51)(10,21.56)(12,21.59)(14,21.64)(16,21.69)(18,21.84)
};

    \addplot[name path=A,draw=none] plot coordinates {
(4,21.21)(6,21.21)(8,21.34)(10,21.38)(12,21.41)(14,21.46)(16,21.59)(18,21.64)
};

 \addplot[name path=B,draw=none] plot coordinates {
(4,21.61)(6,21.69)(8,21.68)(10,21.74)(12,21.77)(14,21.82)(16,21.79)(18,22.04)
};

\addplot[blue!30!white] fill between[of=A and B];

\legend{Geometric-AP,AP,Geometric-AP-RND};
\end{axis}
\end{tikzpicture}
\caption{CR}
\label{fig:subfigureb8}
\end{subfigure}
\begin{subfigure}[b]{0.33\textwidth}
\begin{tikzpicture}
    \begin{axis}[width=\linewidth,font=\footnotesize,
    xlabel= $K$, ylabel= Percentage (\%),
  ymin=0,ymax=80,xmin=4,xmax=18,legend style={nodes=right},legend pos= north west,legend style={nodes={scale=0.75, transform shape}},
      xmajorgrids,
    grid style={dotted},
    ymajorgrids,
  xtick={4,6,8,10,12,14,16,18},
    ]
\addplot[smooth,red!80!black,mark=*,mark options={solid}] plot coordinates {
(4,29.82)(6,33.96)(8,32.55)(10,34.48)(12,34.81)(14,36.64)(16,35.97)(18,37.21)
};
     \addplot[smooth,green!60!black,mark=diamond*,mark options={solid}] plot coordinates {
(4,26.76)(6,27.45)(8,36.71)(10,35.06)(12,37.72)(14,37.57)(16,39.93)(18,40.45)
};
     \addplot[smooth,blue!80!black,mark=triangle*,mark options={solid}] plot coordinates {
(4,6.59)(6,6.66)(8,6.72)(10,6.82)(12,6.88)(14,6.97)(16,7.07)(18,7.38)
};

    \addplot[name path=A,draw=none] plot coordinates {
(4,5.97)(6,5.99)(8,6.26)(10,6.34)(12,6.41)(14,6.51)(16,6.66)(18,6.89)
};

 \addplot[name path=B,draw=none] plot coordinates {
(4,7.21)(6,7.33)(8,7.18)(10,7.30)(12,7.35)(14,7.43)(16,7.48)(18,7.87)
};

\addplot[blue!30!white] fill between[of=A and B];

\legend{Geometric-AP,AP,Geometric-AP-RND};
\end{axis}
\end{tikzpicture}
\caption{F1}
\label{fig:subfigurec8}
 \end{subfigure}
\caption{Clustering performance metrics as a function of the number of clusters learned on the Citeseer dataset. Evaluation measures for Geometric-AP-RND are averaged over 100 repetitions with randomly permuted networks. Blue-shaded area in \textbf{a} denotes the standard deviation associated with NMI. For CR and F1, standard deviations found to be too small and thus invisible w.r.t the scale of the plots.}
\label{fig:Figure8}
\end{figure*}
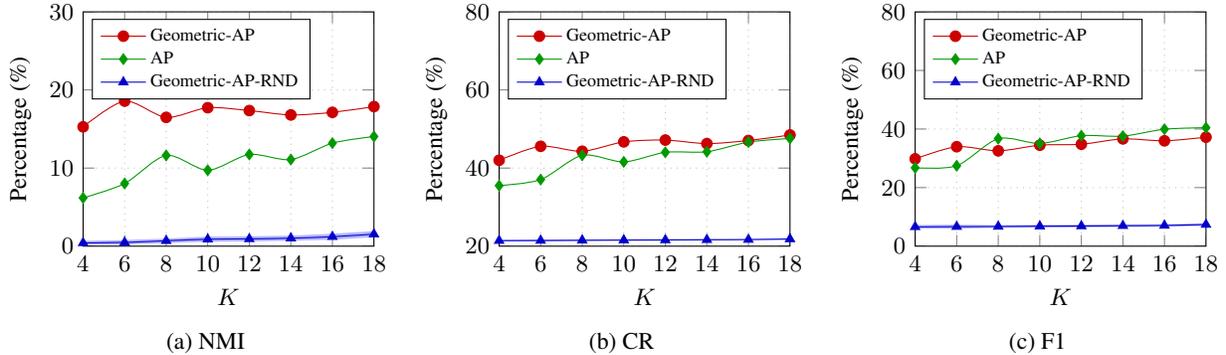